\documentclass{article}

\usepackage{microtype}
\usepackage{graphicx}
\usepackage{booktabs} 
\usepackage{caption}
\usepackage{subcaption}
\usepackage{colortbl}

\usepackage{hyperref}

\usepackage[accepted]{icml2025}

\usepackage{amsmath}
\usepackage{amssymb}
\usepackage{mathtools}
\usepackage{amsthm}

\usepackage{placeins}

\usepackage{mdframed}

\usepackage[capitalize,noabbrev]{cleveref}

\theoremstyle{plain}

\theoremstyle{definition}

\theoremstyle{remark}

\icmltitlerunning{Benchmark Shadows: Data Alignment, Parameter Footprints, and Generalization in Large Language Models}

\begin{document}

\twocolumn[
\icmltitle{Benchmark Shadows: Data Alignment, Parameter Footprints, and Generalization in Large Language Models}



\icmlsetsymbol{equal}{*}

\begin{icmlauthorlist}
\icmlauthor{Hongjian Zou}{equal,comp}
\icmlauthor{Yidan Wang}{equal,sch}
\icmlauthor{Qi Ding}{comp}
\icmlauthor{Yixuan Liao}{comp}
\icmlauthor{Xiaoxin Chen}{comp}
\end{icmlauthorlist}

\icmlaffiliation{comp}{Vivo AI Lab, Shenzhen, China}
\icmlaffiliation{sch}{Hong Kong University of Science and Technology, Hong Kong, China}

\icmlcorrespondingauthor{Hongjian Zou}{hongjianzou@gmail.com}

\icmlkeywords{Machine Learning, ICML}

\vskip 0.3in
]



\printAffiliationsAndNotice{\icmlEqualContribution} 

\begin{abstract}
Large language models often achieve strong benchmark gains without corresponding improvements in broader capability. We hypothesize that this discrepancy arises from differences in training regimes induced by data distribution.
To investigate this, we design controlled data interventions that isolate distributional effects under fixed training settings. We find that benchmark-aligned data improves narrow evaluation metrics while limiting broader representational development, whereas coverage-expanding data leads to more distributed parameter adaptation and better generalization.
We further introduce parameter-space diagnostics based on spectral and rank analyses, which reveal distinct structural signatures of these regimes. Similar patterns are observed across diverse open-source model families, including multimodal models as a key case study, suggesting that these effects extend beyond controlled settings. A case study on prompt repetition shows that not all data artifacts induce regime shifts.
These results indicate that benchmark performance alone is insufficient to characterize model capability, and highlight the importance of data distribution in shaping learning dynamics.
\end{abstract}

\section{Introduction}
\label{sec:introduction}

Recent large language models have achieved rapid progress on a wide range of benchmark evaluations. However, improvements in benchmark scores do not always translate into corresponding gains in broader capability, such as robustness, compositional reasoning, or generalization beyond evaluation distributions. This discrepancy raises a fundamental question: what aspects of training determine how model capability actually develops?

A common implicit assumption is that scaling data and optimizing for stronger task supervision will monotonically improve model performance. Yet in practice, models trained on different data mixtures---despite similar architectures and optimization settings---often exhibit qualitatively different behaviors. This suggests that training outcomes cannot be fully explained by scale alone, and that the structure of the training data plays a more central role than is typically acknowledged.

In this work, we propose a regime-centric perspective, in which differences in data distribution induce distinct training regimes that govern how models allocate capacity during learning. Rather than treating training as a uniform process, we view it as operating under different regimes characterized by how information is distributed across samples. In particular, we distinguish between coverage-expanding data, which increases semantic diversity, and benchmark-aligned data, which concentrates training signals on narrow evaluation-relevant patterns.

To study these effects, we begin with controlled experiments on a text-only decoder model, where architectural and optimization confounds can be tightly controlled, and then test whether the resulting signatures recur in independently trained multimodal systems. This allows us to isolate how data composition influences both performance and internal model structure in a simpler setting, while evaluating the broader relevance of the resulting diagnostics in more realistic model families. We complement these experiments with parameter-space diagnostics, using spectral and rank-based analyses to characterize how different regimes reshape representations across layers.

Our results reveal a consistent pattern: benchmark-aligned data can improve narrow evaluation metrics while limiting broader representational development, whereas coverage-expanding data leads to more distributed parameter adaptation and improved generalization. We further show that similar signatures appear across diverse open-source model families, including multimodal models as a key case study, suggesting that these effects extend beyond controlled settings. Finally, through a case study on prompt repetition, we demonstrate that not all data artifacts induce regime shifts, highlighting the importance of semantic structure rather than surface redundancy.

Our contributions are summarized as follows:
\begin{itemize}

\item We introduce a regime-centric framework that links data distribution to learning dynamics in large language models.
\item We design controlled data interventions that isolate the effects of coverage-reducing and coverage-expanding data under fixed training conditions.
\item We propose parameter-space diagnostics that reveal distinct structural signatures associated with different training regimes.
\item We provide empirical evidence across both controlled experiments and external model families, showing that similar patterns arise in practice.
\item We present a case study on prompt repetition, clarifying the role of superficial redundancy versus semantic concentration.

\end{itemize}

\section{Related Work}
\label{relate_work}

Understanding how data distribution shapes learning dynamics is relevant to large language models broadly, but the issue becomes especially visible in multimodal systems, where heterogeneous data sources and multi-stage curation can amplify regime effects. Prior work has studied scaling behavior, benchmark sensitivity, and internal parameter dynamics, but these strands have largely been developed in isolation. Our work connects them by treating data regime as the organizing factor linking training distribution, parameter-space structure, and downstream generalization.

\subsection{Scaling Behavior of Large Models}

Scaling laws have been a central principle in the development of Large Language Models (LLMs), where performance improves predictably as a function of model size, data volume, and compute \cite{sharma2020neural, hoffmann2022training}. These findings suggest that, under appropriate data--compute trade-offs, increasing scale leads to consistent gains in capability across a wide range of tasks. However, the mechanisms through which different data regimes influence scaling behavior remain much less understood, especially in settings where training data is heterogeneous or heavily curated.

This challenge is particularly visible in Multimodal Large Language Models (MLLMs). Recent multimodal systems, which integrate visual encoders with pretrained language models, often exhibit weaker and less predictable scaling behavior compared to their text-only counterparts. Empirical results across model families indicate that increasing training data or model size does not always yield proportional improvements in multimodal reasoning or understanding performance. In some cases, multimodal pretraining can even degrade text-only capabilities during intermediate stages of training.

Several recent works have begun to explore scaling properties in multimodal settings. Early large-scale vision-language pretraining results showed that web-scale noisy image--text data can substantially improve visual and cross-modal representations even under relatively simple training objectives \cite{jia2021scaling}. More explicit scaling-law studies later demonstrated power-law scaling behavior for contrastive language--image learning under public data and open models \cite{cherti2023reproducible}, and extended scaling analysis to generative mixed-modal language models by modeling interaction effects across modalities \cite{aghajanyan2023scaling}. These studies suggest that multimodal performance depends not only on total scale, but also on the composition and interaction of different data sources. Nevertheless, existing analyses primarily focus on external factors such as data size and model architecture, while treating multimodal data as a relatively homogeneous resource. Our work instead asks how differences in data regime shape both scaling outcomes and internal learning dynamics.

\subsection{Data Regimes and Benchmark Effects}

Recent studies suggest that observed performance in large multimodal models is highly sensitive to the structure of the training data regime. In particular, benchmark-aligned curation, task-focused supervision, and contamination can substantially affect evaluation outcomes, making it difficult to distinguish genuine capability improvement from data-induced score inflation \cite{song2024both, chen2025benchmarking}. This issue is especially important in multimodal settings, where both textual and visual overlap with evaluation benchmarks may bias reported gains \cite{song2024both}. Yet prior work has rarely treated data regimes themselves as an explicit object of analysis.

At the same time, recent data-centric training work has shown that model quality depends strongly on how multimodal corpora are filtered, composed, and curated, rather than on raw scale alone \cite{dong2025scalable}. Related work on downstream specialization further indicates that optimizing toward narrow target distributions can improve task-specific performance while weakening broader generalization \cite{huang2024learn}. 

Taken together, these studies highlight a central challenge: benchmark gains do not necessarily reflect uniform capability growth. Our work builds on this observation by studying benchmark-oriented data regimes as a distinct factor that shapes both external performance and internal model structure.

\subsection{Internal Mechanisms and Parameter Dynamics}

A complementary line of work studies how training reshapes internal model structure. Prior work has shown that the spectral properties of neural network weight matrices provide useful signals about optimization and generalization. In particular, random-matrix-theoretic analyses have found that trained models often exhibit heavy-tailed empirical spectral densities, suggesting forms of implicit self-regularization closely tied to training quality and downstream behavior \cite{mahoney2019traditional, martin2021implicit, martin2025setol}. However, it remains unclear how such diagnostics reflect differences in data regime and whether they can help explain why some training distributions generalize more broadly than others.

Building on this perspective, recent diagnostic tools such as WeightWatcher \cite{weightwatcher} enable data-free analysis of layer-wise weight matrices through spectral statistics, providing a practical way to characterize learned parameter structure without relying solely on benchmark outputs. Related studies further suggest that spectral signatures can reflect differences in training dynamics, problem difficulty, and generalization behavior \cite{xiao2023heavy, meng2023impact}. 

Despite this progress, such analyses have rarely been used to study how different training data regimes influence internal organization across model families, especially in multimodal settings. Our work uses spectral and rank-based diagnostics not only as descriptive tools, but as a bridge linking training distribution, parameter-space structure, and downstream generalization.

In contrast to prior work that studies scaling behavior, data curation, or internal diagnostics in isolation, this work connects three levels of analysis: data regime as the training factor, parameter-space diagnostics as the measurement lens, and generalization behavior as the outcome.

\section{Benchmark Shadows}

\subsection{Data Regimes in Multimodal Training}

We use the term \emph{data regime} to refer to the higher-level distributional structure of a training corpus, including how samples are selected, repeated, filtered, and distributed across concepts, tasks, and benchmark-relevant patterns. Under this view, two training sets with similar scale may nevertheless induce substantially different learning dynamics if they differ in concentration, repetition, or alignment with downstream evaluations. This perspective is particularly important in multimodal settings, where training corpora are assembled from heterogeneous sources and shaped by multiple stages of curation.

\subsection{Benchmark Shadows as a Phenomenon}

We define \emph{benchmark shadows} as apparent capability gains that are amplified on targeted evaluations but do not correspond to proportional improvement in broader generalization. A benchmark shadow does not necessarily imply direct contamination or explicit benchmark memorization. Instead, it may arise whenever the training distribution is distorted toward patterns, concepts, or supervision signals that disproportionately benefit a particular evaluation set.

For example, a model trained predominantly on document-oriented or template-regularized data may achieve strong scores on OCR- or document-heavy benchmarks while still exhibiting clear deficits on broader reasoning tasks. Under this view, benchmark improvement alone is not sufficient evidence of uniform capability growth.

Different forms of benchmark-oriented bias may induce superficially similar benchmark gains while reflecting fundamentally different changes in the learned model. We hypothesize that such shadows arise when concentrated supervision favors shortcut adaptation over broader representation learning, producing improvements that are locally effective but weakly transferable.

\subsection{Hypothesis and Diagnostic Approach}

Based on this formulation, we study benchmark shadows along two complementary dimensions: external behavior and internal parameter structure. Externally, we examine whether performance gains remain localized to benchmark-like evaluations or transfer more broadly. Internally, we analyze whether different data regimes induce distinct parameter-space signatures that help explain why some benchmark-oriented gains are more recoverable than others.

To characterize internal structure, we use three diagnostic metrics. The heavy-tailed exponent $\alpha$ summarizes the spectral shape of layer-wise weight matrices --- values in the range [2, 6] are commonly associated with well-conditioned representations, while deviations are often consistent with degraded structure --- and serves as a proxy for learned structure and implicit regularization. Effective rank --- where higher values are generally associated with more distributed representations, and sharp drops suggest rank collapse --- measures how broadly information is distributed across parameter subspaces, providing a coarse indicator of representational spread. Change variance measures how strongly parameter updates are concentrated across layers or modules, helping distinguish broad adaptation from narrow, localized adjustment. Together, these metrics provide a compact diagnostic view of whether benchmark-oriented gains are associated with broad structural learning or more limited shortcut-like adaptation.

This formulation provides the conceptual basis for the controlled experiments in the following sections, where we instantiate different benchmark-oriented data regimes and compare their effects on generalization and parameter dynamics.

\section{Controlled Data Intervention}
\label{controlled_data}

To examine how benchmark-shadow-like effects can arise from training data, we construct a controlled set of interventions around a shared base training pipeline. Our goal is not to reproduce any specific benchmark-aligned corpus directly, but to isolate two simplified mechanisms through which benchmark-oriented data construction can distort learning. In practice, benchmark-focused curation often has two coupled effects: it repeatedly reinforces a narrow family of target-like patterns, either through exact duplication or weak rephrasing, and it shifts training toward canonical high-frequency forms while suppressing broader capability coverage. We model these effects separately through two controlled data conditions.

Across the main data conditions, the model architecture, tokenizer, optimizer, and total training budget are held fixed. In addition, we include a separate optimization-control baseline that changes only the learning-rate schedule, allowing us to distinguish data-induced effects from optimization-induced ones. This design lets us ask three questions: whether concentrated supervision produces distinct parameter-space signatures, whether different forms of concentration are equally recoverable, and which diagnostics are primarily sensitive to data regime rather than optimization.

\subsection{Experimental Setup}

\paragraph{Model and training configuration.}
All experiments use the same decoder-only Transformer language model trained from scratch, with approximately 0.6B parameters. The model has 32 layers, hidden size 1024, feed-forward dimension 4096, and 8 attention heads, with grouped-query attention using 4 query groups. It adopts RMSNorm normalization, RoPE positional encoding, and SwiGLU activations, with a maximum sequence length of 4096. Dropout is disabled for both attention and hidden states. The model is initialized using Xavier uniform initialization with standard deviation 0.02, and all linear layers are bias-free.

Training uses Adam with $(\beta_1=0.9,\beta_2=0.95,\epsilon=10^{-8})$, a peak learning rate of $3\times10^{-4}$, and cosine decay to a minimum learning rate of $3\times10^{-5}$ unless otherwise stated. We use a global batch size of 4096 with micro-batch size 4, gradient clipping of 1.0, and weight decay of 0.1. Models are trained in bf16 precision for a total of 300M training samples, approximately 1.2T tokens. A linear warmup is applied over the first 1.64M samples, followed by decay over the remaining training steps. Unless noted, architecture, tokenizer, optimizer, and total training budget are identical across conditions.

\paragraph{Condition A: Coverage-expanding baseline.}
The reference model is trained on a coverage-expanding dataset designed to maximize diversity across domains, knowledge types, task families, and instruction formats. The dataset aims to provide broad support over capability space and is deduplicated to avoid unintended concentration from repeated or near-duplicate samples. This condition serves as the main baseline for evaluating the effects of concentrated supervision.

\paragraph{Condition B: Optimization-control baseline.}
To separate data effects from optimization artifacts, we introduce an additional baseline that uses the same training data as Condition A but a different learning-rate schedule. Specifically, instead of the default fast-decay cosine schedule, this condition maintains a relatively high learning rate throughout training without fast decay. All other factors remain unchanged. This condition is not a data intervention; it is included to test whether observed structural differences can be explained by optimization alone.

\paragraph{Condition C: Repetition-concentrated regime.}
Condition C models \emph{redundant reinforcement}: a simplified proxy for benchmark-oriented curation that repeatedly exposes the model to a narrow family of target-like patterns. In real pipelines, this effect may arise through exact duplication, weak paraphrasing, repeated template reuse, or repeated inclusion of benchmark-adjacent samples. To isolate the core effect, we implement the simplest version. Let $\mathcal{D}_1$ denote the first half of the training data. We randomly sample 10\% of $\mathcal{D}_1$, repeat each selected sample 10 times, and discard the remaining 90\%. We adopt a simple 10$\times$ repetition setting as a deliberately strong intervention: prior work suggests that modest repetition may be beneficial or largely neutral in some settings, whereas heavier repetition can induce degradation, so our goal here is not to optimize the repetition rate but to make the repeated-reinforcement effect clearly observable under controlled conditions \cite{marion2023whenless, hernandez2022repeated, xue2023repeat}. This preserves the overall sample count while sharply reducing support diversity and increasing local redundancy. The second half of training remains identical to the baseline dataset. Condition C therefore represents concentration by \emph{repeated reinforcement}: the model sees a narrow subset of patterns disproportionately often, but the original sample identities themselves are preserved.

\paragraph{Condition D: Frequency-concentrated regime.}
Condition D models \emph{support collapse}: a benchmark-oriented selection pressure that shifts training toward canonical high-frequency forms while suppressing broader diversity. This effect is intended to capture what happens when data construction over-prioritizes benchmark-relevant formats and discards long-tail variation that supports general capability. Let $\mathcal{D}_1$ again denote the first half of the training data. For each sample in $\mathcal{D}_1$, we apply a teacher-based rewriting procedure using Qwen3-30B-A3B-Instruct-2507 that simplifies lexical and structural variation by favoring more frequent tokens and more canonical forms. In code data, for example, programs written in different languages are converted into a single target language with normalized naming and formatting; in natural-language data, paraphrases are simplified toward more common lexical choices and syntactic patterns. This intervention is motivated by the hypothesis that heavily benchmark-aligned or frequency-concentrated data may erode the natural long-tail coverage present in broader pretraining corpora, consistent with recent work showing that rare-word generalization is fragile and that tail behavior reveals capability differences not captured by head-focused evaluation alone \cite{algayres2025longtail}. The second half of training remains identical to the baseline dataset. This preserves the overall token budget while collapsing lexical and structural diversity in early training, thereby concentrating supervision on canonical high-frequency patterns. Although the rewriting distribution may partly reflect the teacher model's own preferences, the intervention is intended as a controlled approximation of support concentration rather than a model-specific data recipe.

\paragraph{Phase-shift design.}
Conditions C and D share the same phase-shift curriculum: the first half of training (approximately steps 0--36k) uses a concentrated regime, while the second half (steps 36k--72k) reverts to the coverage-expanding baseline. This design allows us to separate immediate structural disruption from later recovery. If the model fully returns to the baseline trajectory after exposure to diverse data, the concentrated early regime is largely reversible; if not, the intervention leaves persistent, path-dependent effects on learned structure. Checkpoints are analyzed at multiple stages to track both disruption and recovery.

\paragraph{Interpretive scope.}
We emphasize that Conditions C and D are not intended to reconstruct any single real benchmark-aligned corpus. Rather, they serve as controlled abstractions of two common consequences of benchmark-oriented data construction: concentration by redundant reinforcement (Condition C) and concentration by diversity collapse (Condition D). This allows us to study benchmark-shadow-like behavior mechanistically, rather than treating ``benchmark alignment'' as a single undifferentiated label.

\subsection{Results}

We report three main findings. First, different concentrated data regimes produce distinct parameter-space trajectories even under matched training budgets. Second, the two concentration mechanisms differ in recoverability: repetition-concentrated training is largely reversible after exposure to diverse data, whereas frequency-concentrated training leaves more persistent structural footprints. Third, the optimization-control baseline shows that some diagnostics are primarily sensitive to learning-rate schedule, whereas others more clearly track data regime. Table~\ref{tab:ctrl_data_alpha_range} summarizes these temporal trends, and additional checkpoint-wise model-level histograms (see Appendix) make the recovery asymmetry between Conditions C and D visually explicit.

\begin{figure}[t]
\centering

\begin{subfigure}[b]{0.45\linewidth}
    \centering
    \includegraphics[width=\linewidth]{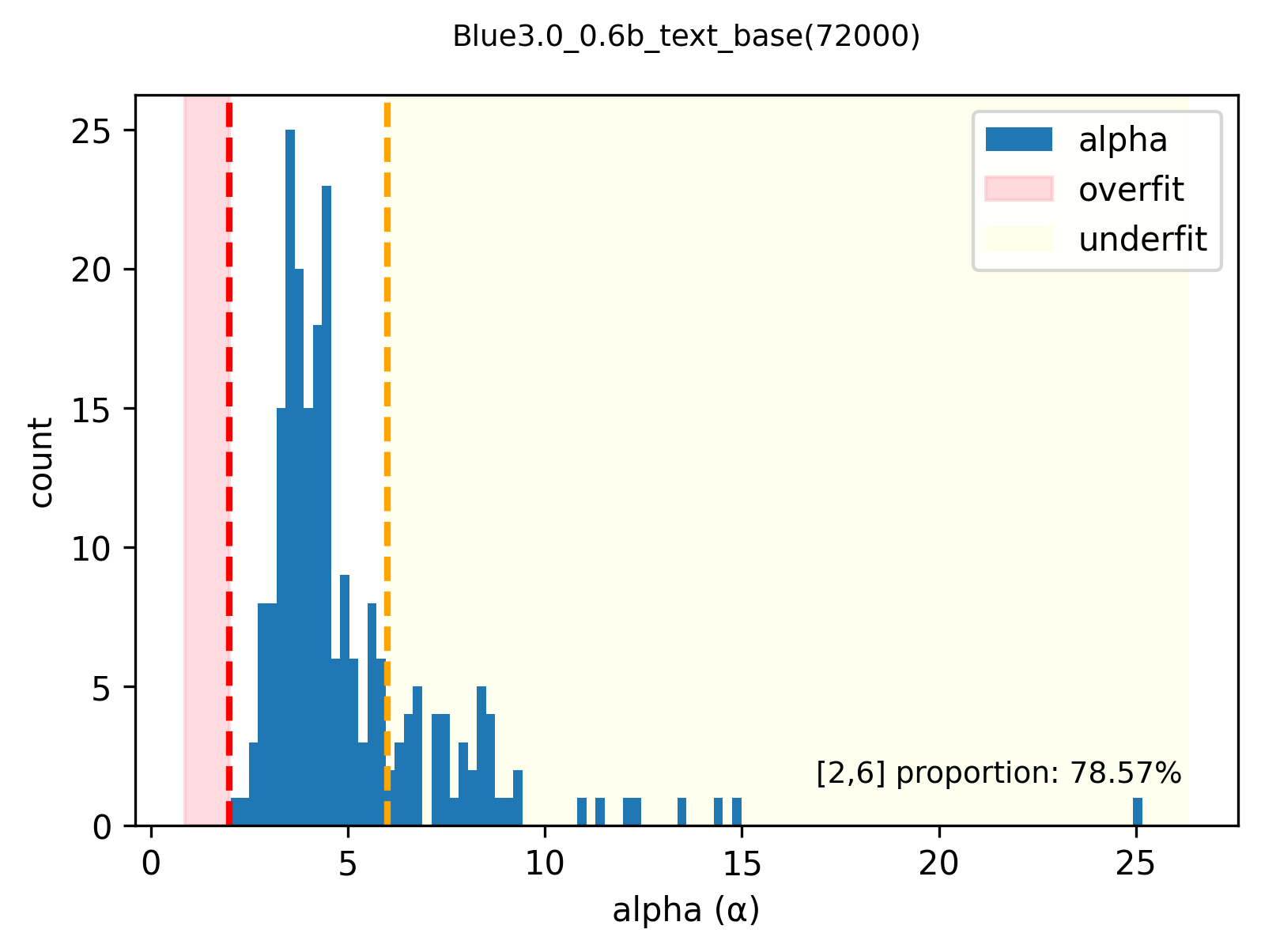}
    \caption{Condition A}
\end{subfigure}
\hfill
\begin{subfigure}[b]{0.45\linewidth}
    \centering
    \includegraphics[width=\linewidth]{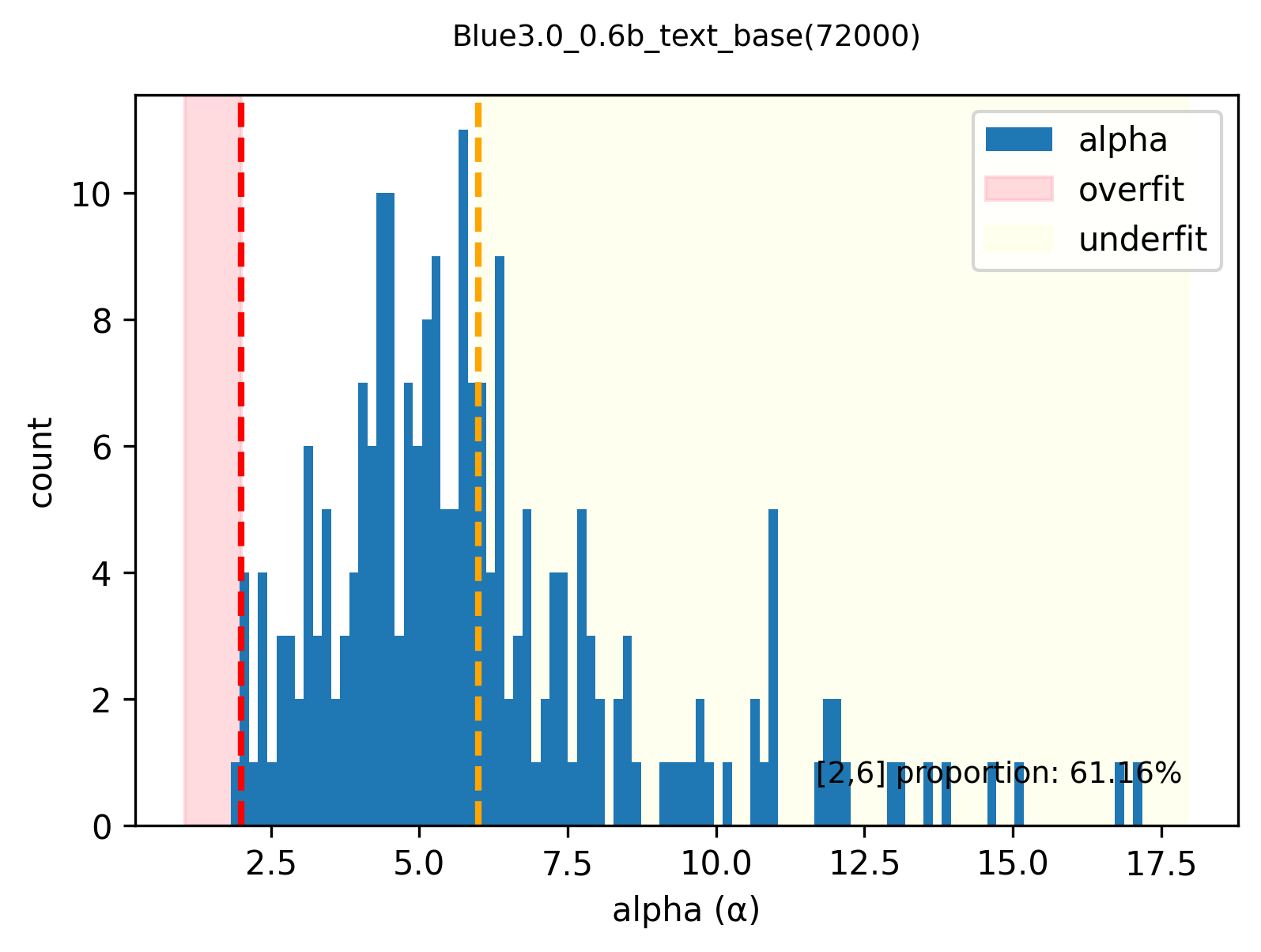}
    \caption{Condition B}
\end{subfigure}

\vspace{0.3cm}

\begin{subfigure}[b]{0.45\linewidth}
    \centering
    \includegraphics[width=\linewidth]{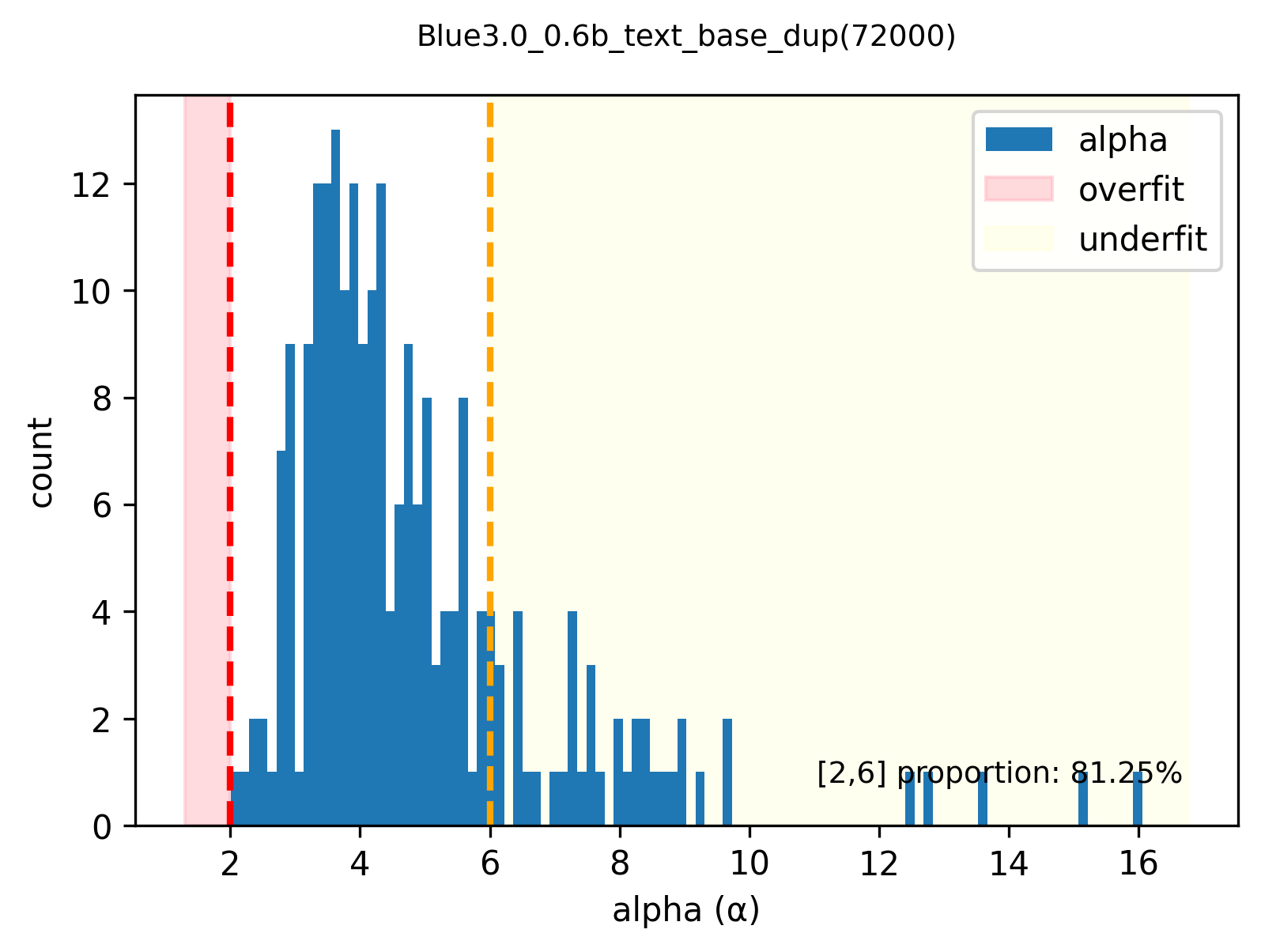}
    \caption{Condition C}
\end{subfigure}
\hfill
\begin{subfigure}[b]{0.45\linewidth}
    \centering
    \includegraphics[width=\linewidth]{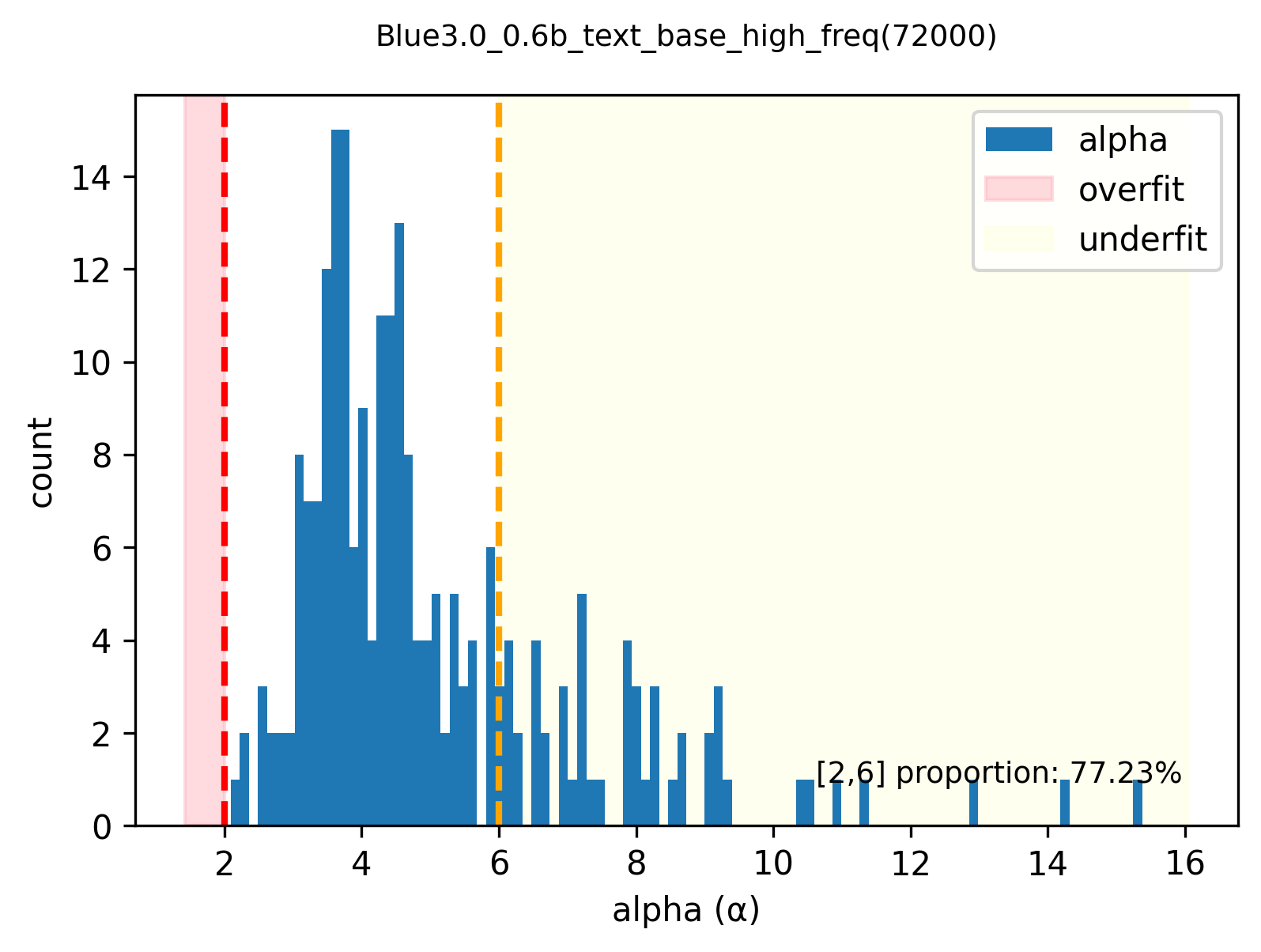}
    \caption{Condition D}
\end{subfigure}

\caption{Comparison of four training conditions. Condition A: baseline under the coverage-expanding regime. Condition B: baseline with an alternative learning-rate schedule (no fast decay). Condition C: repetition-concentrated regime. Condition D: frequency-concentrated regime.}
\label{fig:ctrl_data_mdl_lvl_cmp}
\end{figure}

\begin{table}[t]
  \centering
  \caption{Proportion of weight matrices with $\alpha\!\in\![2,6]$ (\%) across training steps.
  \colorbox{green!15}{Green}: $\geq$78\%; \colorbox{yellow!20}{yellow}: 70--78\%; \colorbox{red!12}{red}: $<$70\%.}
  \label{tab:ctrl_data_alpha_range}
  \scriptsize
  \setlength{\tabcolsep}{2.5pt}
  \begin{tabular}{@{}lcccccc@{}}
    \toprule
    & \textbf{2k} & \textbf{20k} & \textbf{40k} & \textbf{60k} & \textbf{72k} & \textbf{Trend} \\
    \midrule
    A: overage-expanding baseline
      & 60.7 & \cellcolor{green!15}81.7 & \cellcolor{green!15}79.0 & \cellcolor{green!15}79.5 & \cellcolor{green!15}78.6
      & Stable \\
    B: optimization-control baseline
      & 62.1 & \cellcolor{red!12}57.6 & \cellcolor{red!12}57.6 & \cellcolor{red!12}61.2 & \cellcolor{red!12}61.2
      & Stuck \\
    C: repetition-concentrated regime
      & 58.9 & \cellcolor{green!15}79.9 & \cellcolor{green!15}81.7 & \cellcolor{green!15}80.4 & \cellcolor{green!15}81.3
      & Recover \\
    D: requency-concentrated regime
      & 61.2 & \cellcolor{yellow!20}77.2 & \cellcolor{yellow!20}78.1 & \cellcolor{yellow!20}77.7 & \cellcolor{yellow!20}77.2
      & Gap \\
    \bottomrule
  \end{tabular}
  \vspace{-0.5em}
\end{table}

\subsubsection{Model-Level Parameter Distribution}

Figure~\ref{fig:ctrl_data_mdl_lvl_cmp} shows the distribution of $\alpha$ values across layers at the final training stage (72k) for different conditions. The coverage-expanding baseline remains concentrated within the well-conditioned range, whereas the concentrated regimes shift toward heavier-tailed distributions. This shift is especially informative when viewed together with Table~\ref{tab:ctrl_data_alpha_range}, which reports the proportion of weight matrices with $\alpha \in [2,6]$ over training. The baseline remains stable at a high level, while Condition D retains a persistent gap even after later exposure to diverse data. In contrast, Condition C shows early degradation but recovers to baseline-level conditioning at convergence.

These results indicate that not all concentration mechanisms behave similarly. Repetition-concentrated training mainly distorts exposure frequency and is substantially reversible once the model returns to broader coverage. Frequency-concentrated training, by contrast, removes lexical and structural support from the long tail during early training and leaves a more persistent geometric footprint. In other words, the more damaging effect is not repetition alone, but narrowing the support of training data itself.

To further examine temporal behavior, we compare the evolution of $\alpha$ distributions across checkpoints. Both concentrated regimes show early movement toward less favorable spectral structure, followed by later recovery once training returns to the coverage-expanding distribution. However, the recovery is asymmetric: Condition C returns to a baseline-like endpoint, whereas Condition D remains offset. This asymmetry supports the view that support collapse is harder to undo than redundant reinforcement.

This addresses the first two questions posed above: concentrated supervision does produce distinct parameter-space signatures, and the two concentration mechanisms differ in recoverability.

\subsubsection{Layer-wise Spectral Comparison}

\begin{figure}[t]
\centering

\begin{subfigure}[b]{0.45\linewidth}
    \centering
    \includegraphics[width=\linewidth]{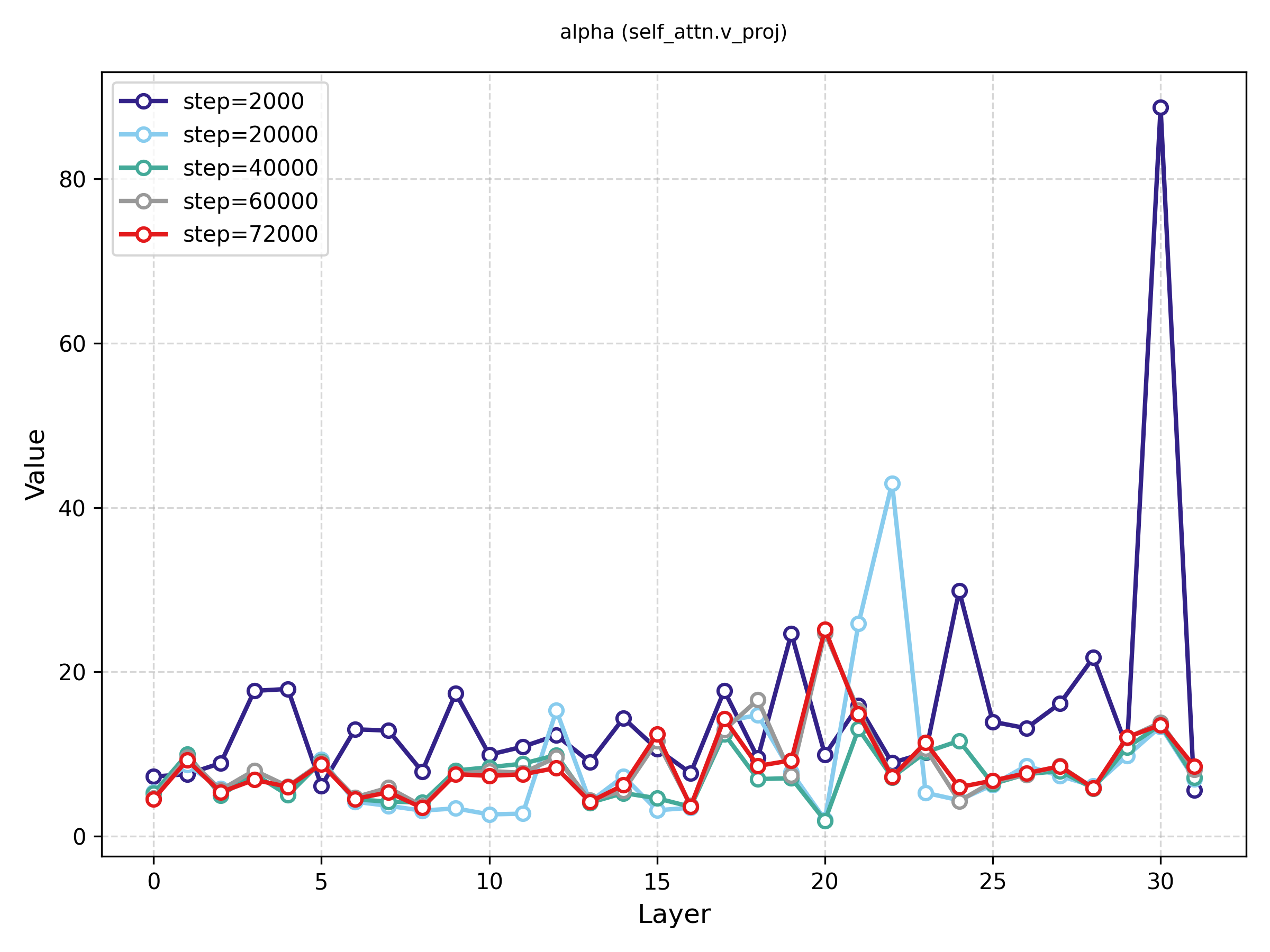}
    \caption{Condition A}
\end{subfigure}
\hfill
\begin{subfigure}[b]{0.45\linewidth}
    \centering
    \includegraphics[width=\linewidth]{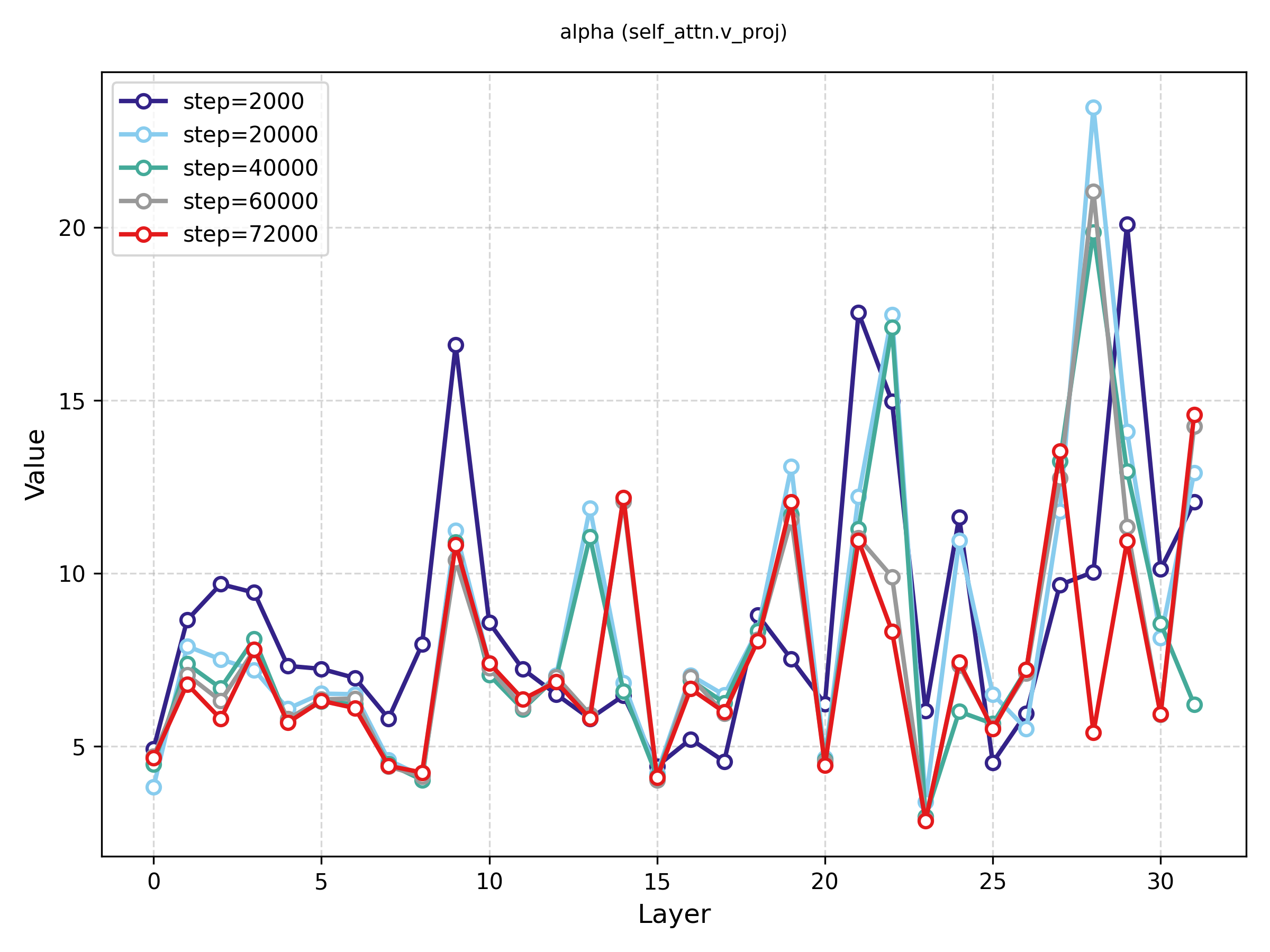}
    \caption{Condition B}
\end{subfigure}

\vspace{0.3cm}

\begin{subfigure}[b]{0.45\linewidth}
    \centering
    \includegraphics[width=\linewidth]{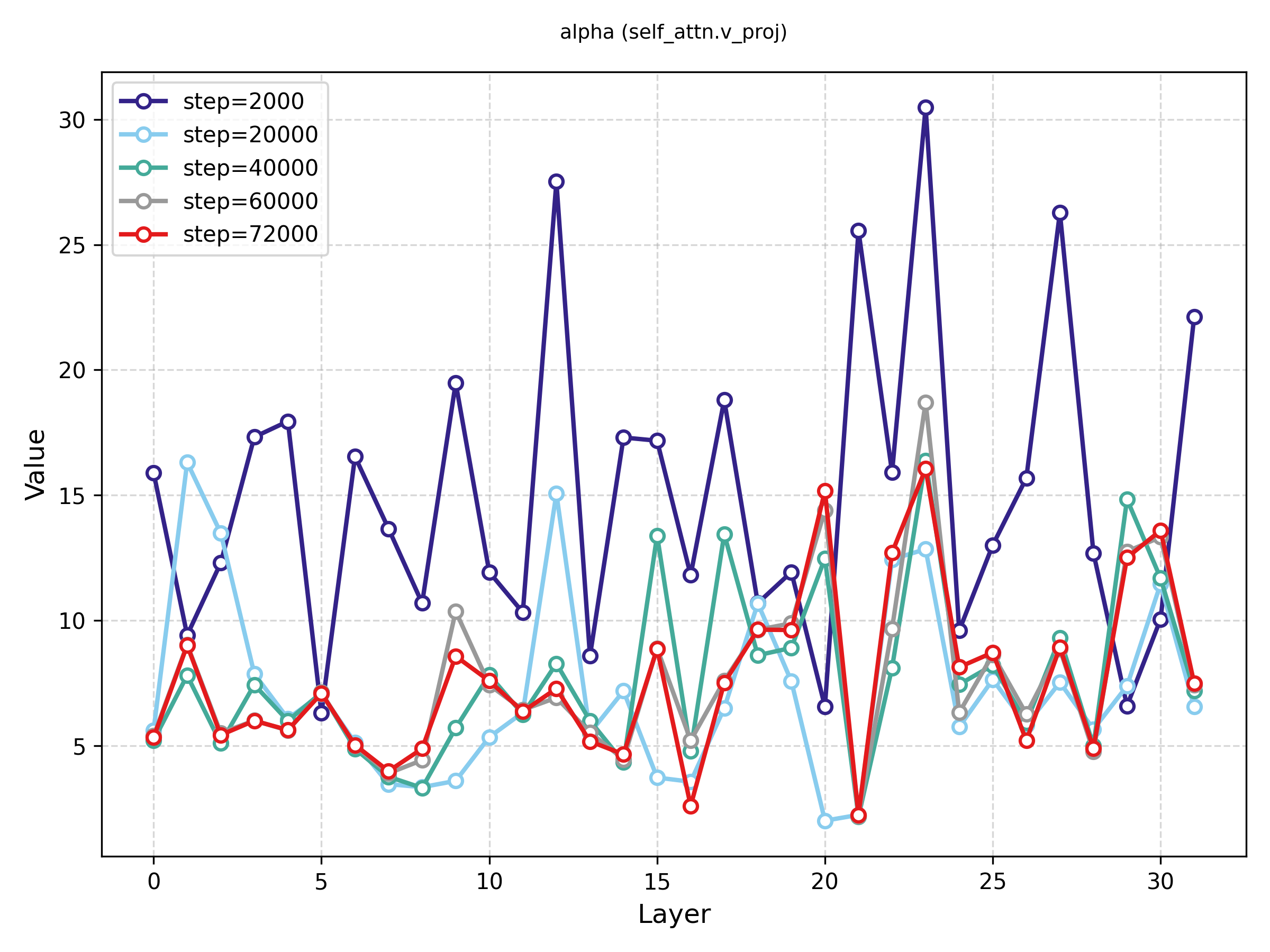}
    \caption{Condition C}
\end{subfigure}
\hfill
\begin{subfigure}[b]{0.45\linewidth}
    \centering
    \includegraphics[width=\linewidth]{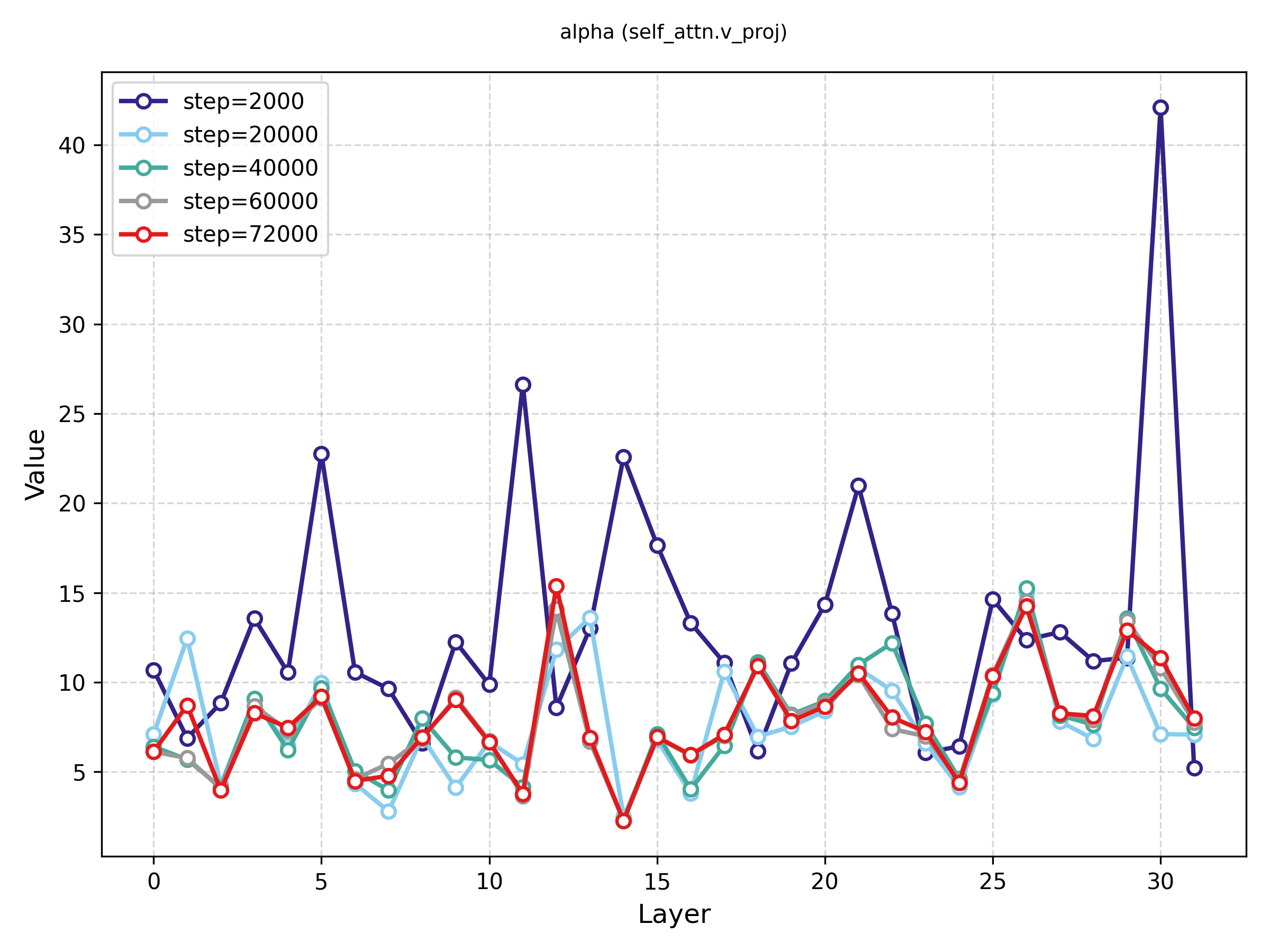}
    \caption{Condition D}
\end{subfigure}

\caption{
Layer-wise $\alpha$ values in \texttt{self\_attn.v\_proj} at the final training stage (72k steps).
Conditions C and D exhibit stronger layer-wise heterogeneity than the baseline settings, with persistent deviations concentrated in deeper layers.
}
\label{fig:ctrl_data_spec_cmp_attn_alpha}
\end{figure}

To understand where regime effects accumulate, we next analyze layer-wise spectral statistics across network components. In contrast to the model-level view, which summarizes aggregate conditioning, this analysis reveals how concentrated supervision manifests heterogeneously across depth.

\paragraph{Attention projections as sensitive indicators.}
Among all components, attention projection layers---particularly the value projection (\texttt{v\_proj})---show the strongest sensitivity to training regime. Figure~\ref{fig:ctrl_data_spec_cmp_attn_alpha} presents the layer-wise $\alpha$ profiles at the final training stage. Under the coverage-expanding baseline (Condition A), $\alpha$ values remain relatively smooth across layers, consistent with stable and distributed representation updates. The optimization-control baseline (Condition B) produces a broadly similar overall profile, suggesting that learning-rate variation alone does not reproduce the same regime-specific distortions. In contrast, both concentrated data conditions introduce stronger layer-wise heterogeneity, especially in deeper layers. Here again the difference between C and D is important: Condition C largely returns to a baseline-like structure, whereas Condition D retains persistent elevation in upper layers, indicating incomplete structural recovery.

Notably, this regime sensitivity is not uniform across pathways. Across spectral diagnostics, condition-specific differences are most pronounced in attention projections, whereas MLP layers are comparatively stable and show much weaker separation across conditions (see Appendix). This attention--MLP asymmetry will also reappear in the external validation results, where cross-model divergences concentrate most clearly in \texttt{v\_proj}.

\paragraph{Distinguishing data and optimization effects.}
Comparing Conditions A and B helps separate optimizer-driven behavior from data-driven behavior. While their layer-wise $\alpha$ profiles are broadly consistent, Condition B exhibits a distinct monotonic trend in $\lambda_{\min}$ across layers (see Appendix), consistent with an optimization-driven effect rather than a data-induced regime shift. By contrast, Conditions C and D show irregular, layer-specific fluctuations rather than a smooth global deformation. This difference suggests that concentrated data leaves structural fingerprints that are qualitatively different from those induced by learning-rate schedule alone.

\paragraph{Inhomogeneous representational change.}
Related metrics such as effective feature number (see Appendix) show a consistent pattern. In particular, the frequency-concentrated regime (Condition D) exhibits the strongest inhomogeneity in later training stages, with sharper variations concentrated in upper layers, whereas the baseline remains more uniform. This is consistent with uneven allocation of representational capacity under support collapse and with the broader interpretation that diversity-reducing concentration distorts representation formation more persistently than repetition-concentrated training.

\paragraph{Summary.}
Overall, the layer-wise analysis shows that regime effects are not uniform across depth. The baseline remains comparatively smooth, the optimization-control baseline alters trajectory without producing the same irregular distortions, Condition C partially normalizes by convergence, and Condition D retains persistent upper-layer deviations. These differences indicate that data concentration leaves distinct structural footprints beyond those induced by learning-rate variation alone.

\subsubsection{Parameter Dynamics}

\begin{figure}[t]
\centering

\begin{subfigure}[b]{0.45\linewidth}
    \centering
    \includegraphics[width=\linewidth]{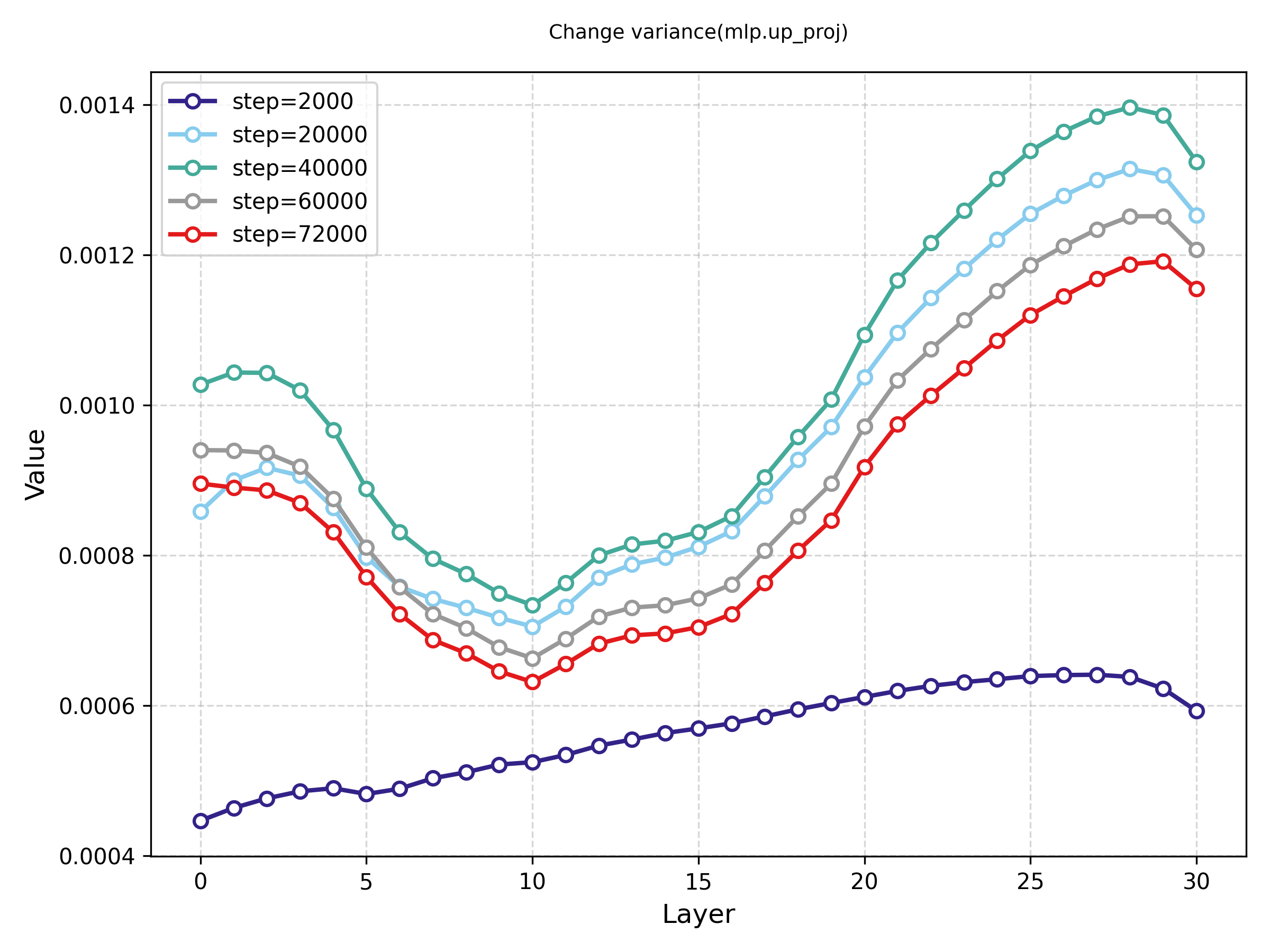}
    \caption{Condition A}
\end{subfigure}
\hfill
\begin{subfigure}[b]{0.45\linewidth}
    \centering
    \includegraphics[width=\linewidth]{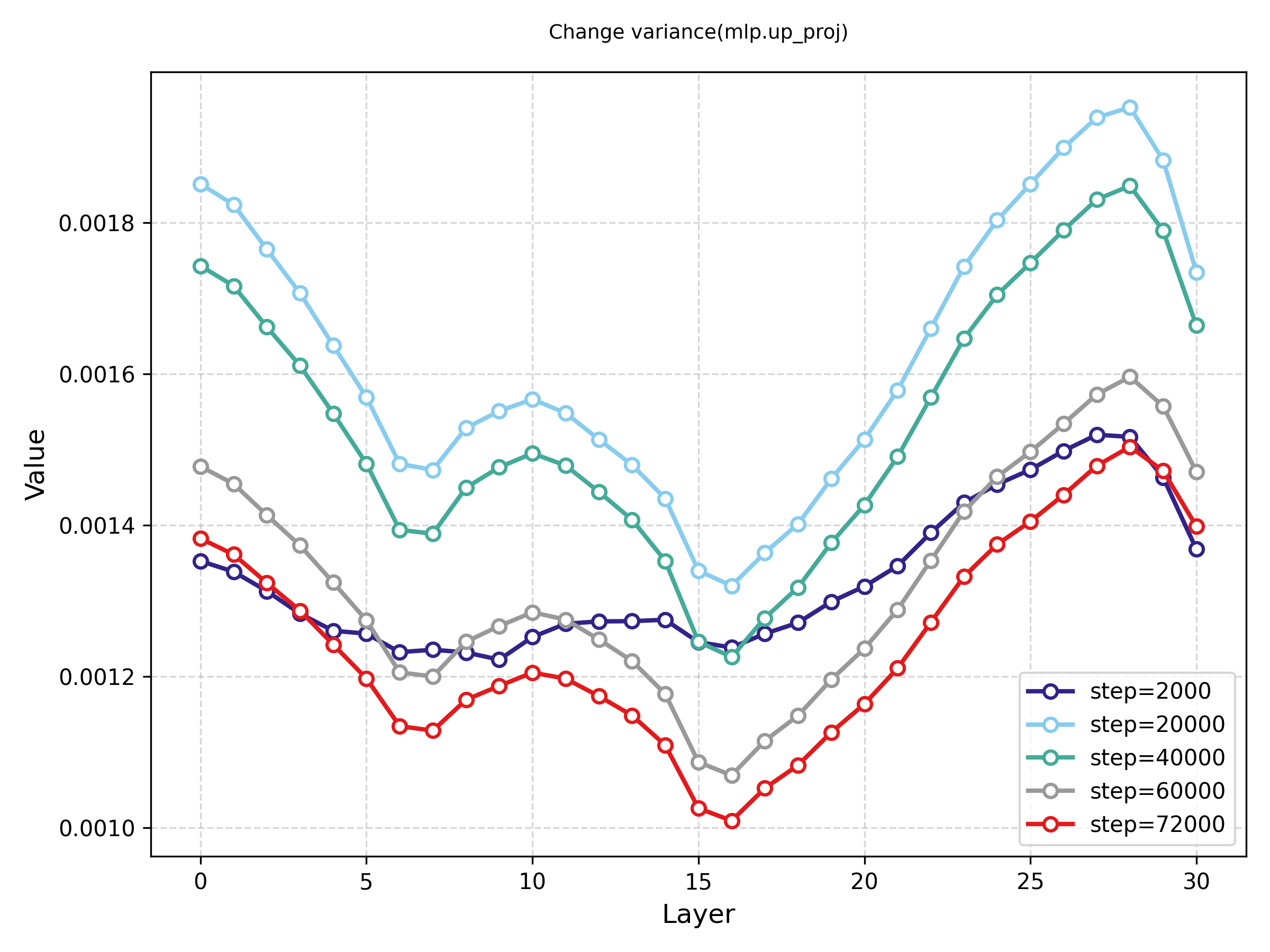}
    \caption{Condition B}
\end{subfigure}

\caption{Change variance in \texttt{mlp.up\_proj}. (a)~Baseline-cosine-lr and (b)~large-LR-no-decay share the same U-shaped depth profile, but condition~B shows $1.3\times$ amplification and non-monotonic step ordering. Conditions~C and~D (not shown; see Appendix) are nearly identical to~(a).}
\label{fig:ctrl_data_chg_var}
\end{figure}

\begin{figure}[t]
\centering

\begin{subfigure}[b]{0.45\linewidth}
    \centering
    \includegraphics[width=\linewidth]{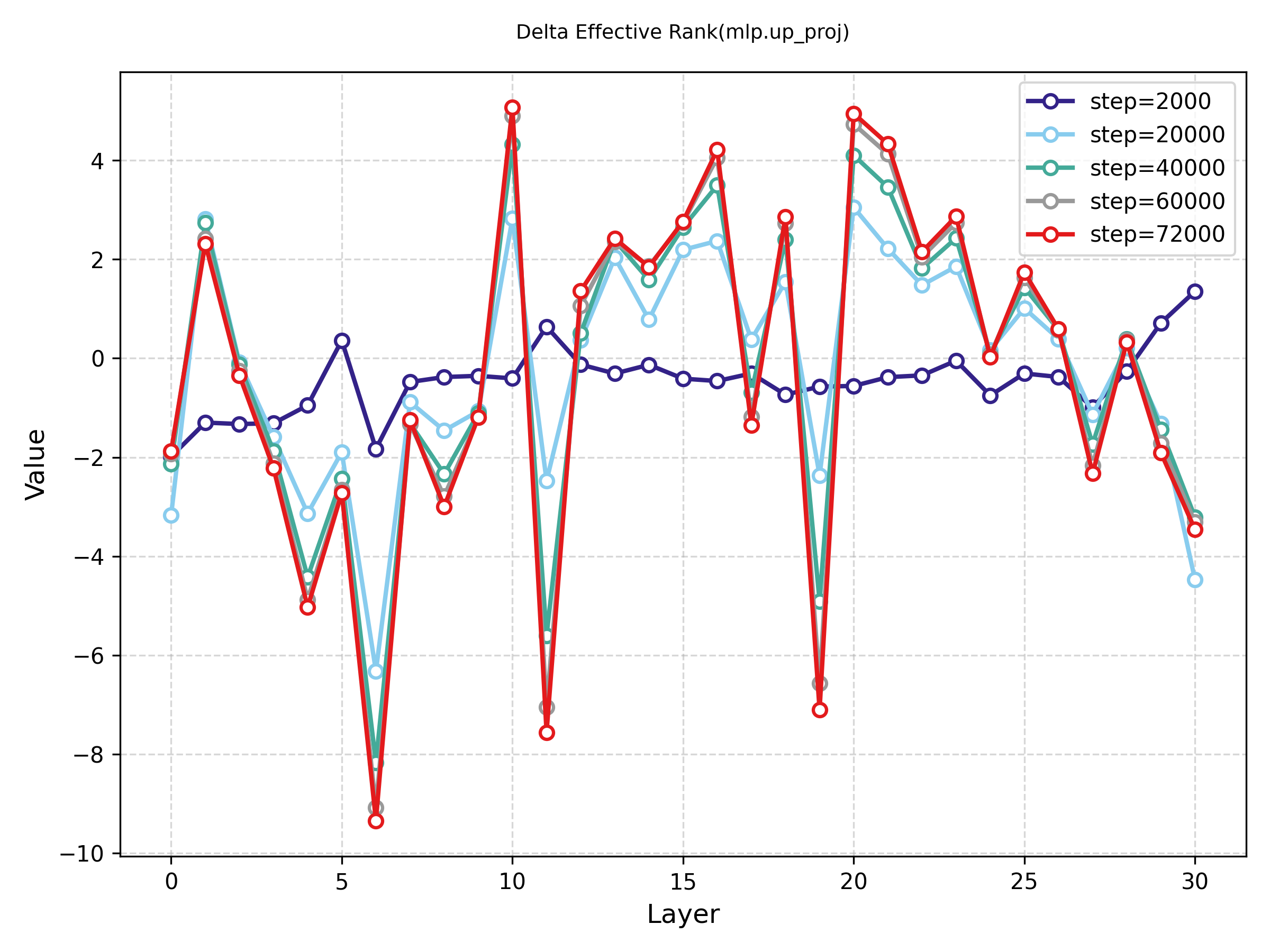}
    \caption{Condition A}
\end{subfigure}
\hfill
\begin{subfigure}[b]{0.45\linewidth}
    \centering
    \includegraphics[width=\linewidth]{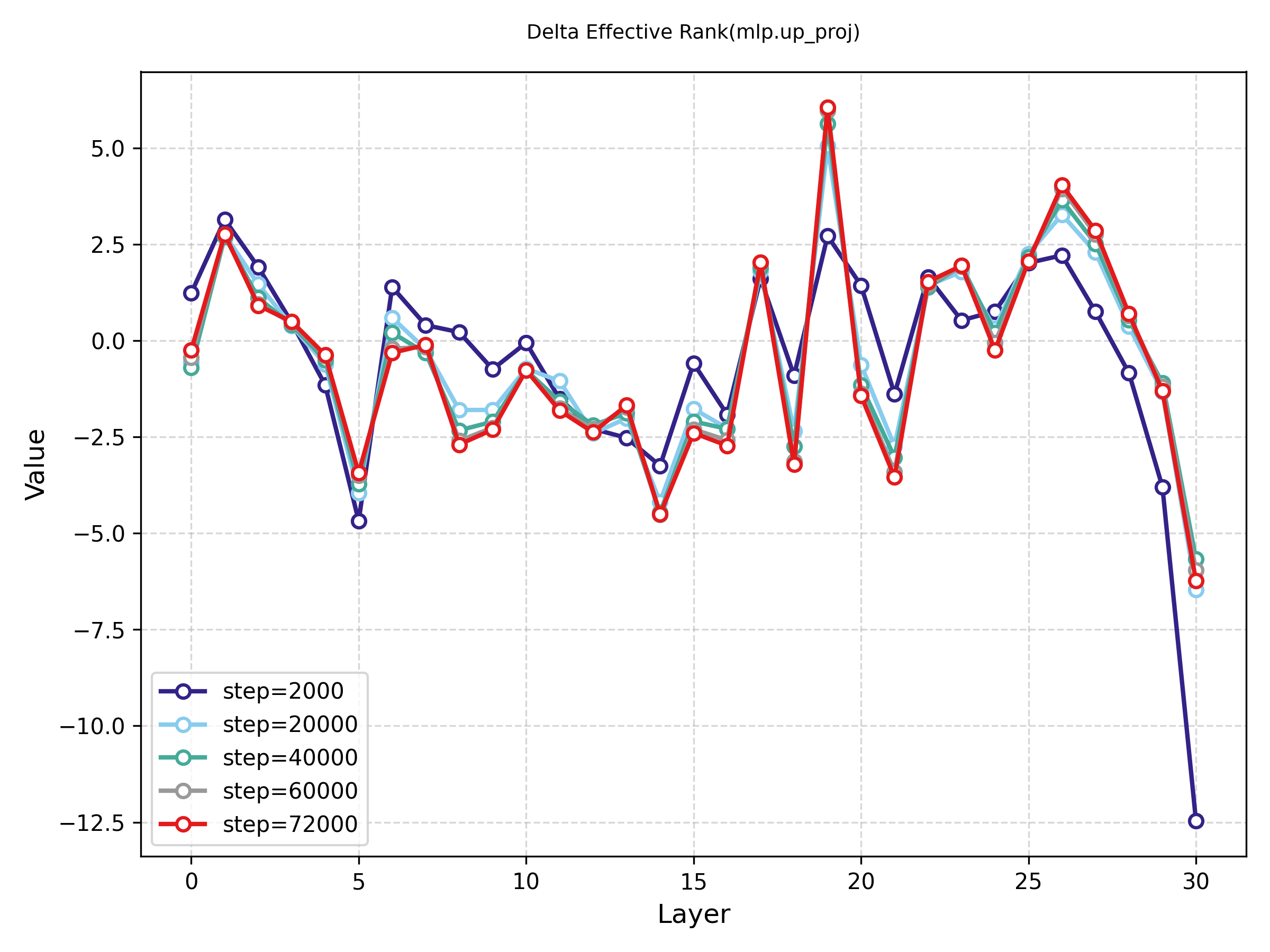}
    \caption{Condition B}
\end{subfigure}

\vspace{0.3cm}

\begin{subfigure}[b]{0.45\linewidth}
    \centering
    \includegraphics[width=\linewidth]{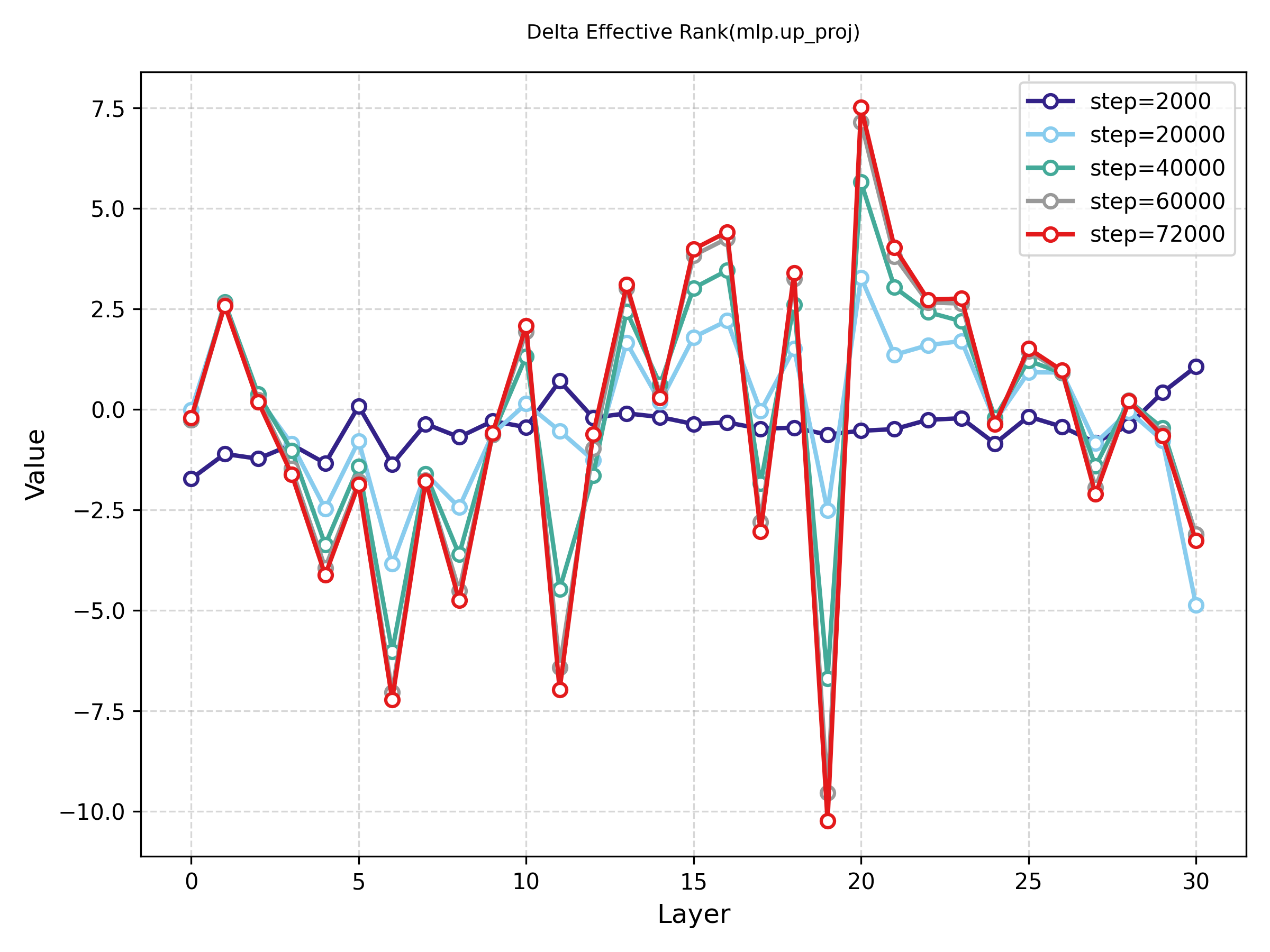}
    \caption{Condition C}
\end{subfigure}
\hfill
\begin{subfigure}[b]{0.45\linewidth}
    \centering
    \includegraphics[width=\linewidth]{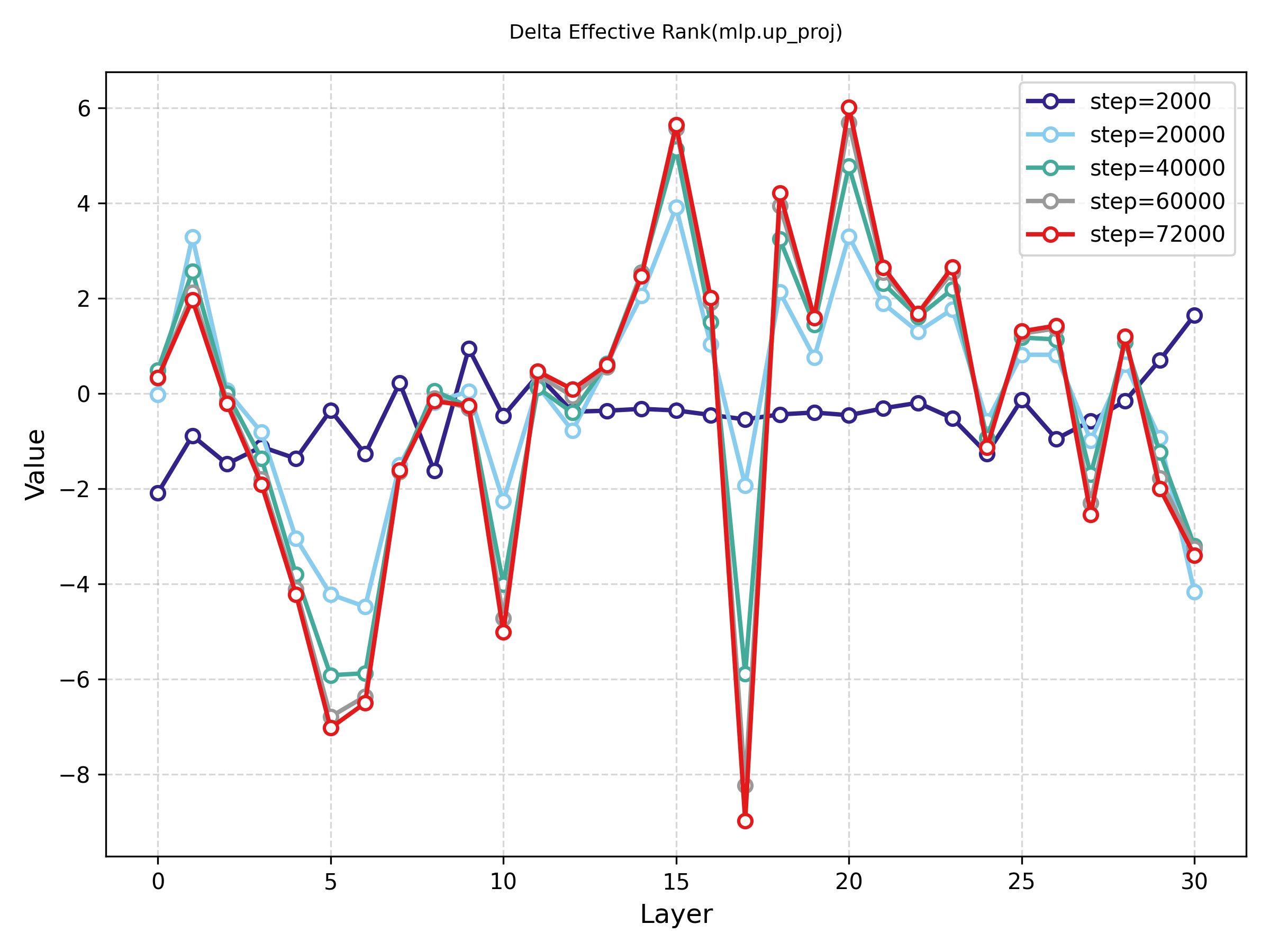}
    \caption{Condition D}
\end{subfigure}

\caption{Delta effective rank in \texttt{mlp.up\_proj} across training steps.
(a)~Baseline: moderate, balanced oscillations at convergence.
(c)~Repeated-bias: pronounced oscillations persist in mid-to-upper layers at 72k despite aggregate [2,6]\% recovery.
(d)~High-freq bias: compression shifts toward mid-network layers, with tighter inter-checkpoint clustering.}
\label{fig:ctrl_data_delta_eff_rank_attn_alpha}
\end{figure}

To further disentangle the effects of optimization and data distribution, we analyze layer-wise parameter dynamics using multiple complementary metrics, including mean change, relative parameter change, correlation, change variance, and delta effective rank.

\paragraph{Change variance as an optimizer diagnostic.}
We first examine change variance across layers, which reflects how update magnitudes are distributed over depth. Figure~\ref{fig:ctrl_data_chg_var} shows representative profiles for different conditions. Under the baseline regime (Condition A), change variance exhibits a characteristic U-shaped pattern across layers, with relatively higher variance in early and late layers and a stable middle region. This profile remains nearly unchanged across Conditions A, C, and D, indicating that change variance is largely invariant to data regime. In contrast, the optimization-control baseline (Condition B) produces amplified and less regular fluctuations. This shows that change variance is primarily governed by optimization schedule rather than data concentration, making it useful as an optimizer diagnostic.

\paragraph{Delta effective rank as a data diagnostic.}
We next analyze delta effective rank, which captures changes in representational dimensionality across layers. Unlike change variance, delta effective rank exhibits strong condition-specific behavior. The baseline shows relatively smooth and distributed changes across layers, consistent with incremental representation refinement. Condition C shows moderate deviations but broadly preserves the overall structure, indicating substantial later recovery. Condition D, however, displays more pronounced and localized disruptions in deeper layers, with sharper increases and decreases that reflect uneven representational restructuring. These patterns suggest that delta effective rank tracks data-induced structural effects that persist beyond optimization dynamics.

\paragraph{Disentangling optimization and data effects.}
Taken together, these results reveal a useful separation between optimizer-driven and data-driven signals. Change variance is primarily controlled by optimization and remains relatively stable across data regimes, whereas delta effective rank is much more sensitive to training data distribution and exhibits condition-specific structural signatures. This distinction provides a practical diagnostic framework: optimization artifacts and data-induced regime shifts can be separated by examining complementary layer-wise metrics rather than relying on aggregate benchmark outcomes alone.

\paragraph{Summary.}
While optimization determines the overall magnitude and smoothness of parameter updates, data distribution governs how representational capacity is allocated across layers. Concentrated data regimes do not merely alter update strength; they reshape the internal structure of representations in a layer-specific and path-dependent manner. Importantly, the two concentration mechanisms are not equivalent: repetition-concentrated training is largely recoverable, whereas diversity-reducing concentration leaves more persistent structural disruption.

This also complements the layer-wise spectral results above: attention pathways remain the main locus of condition-sensitive structural distortion, while MLP dynamics help separate optimizer-driven from data-driven effects.

This addresses the third question: change variance is primarily sensitive to optimization schedule, while delta effective rank and layer-wise alpha track data regime, providing a practical basis for separating the two.

\section{External Validation Across Model Families}
\label{external_validation}

Section~\ref{controlled_data} established, under controlled interventions, that different forms of concentrated training induce distinct parameter-space signatures and differ in recoverability. We now ask whether analogous signatures recur in real systems. Because independently trained models differ simultaneously in data mixture, optimization, training duration, and multimodal design, the analysis in this section is correlational rather than causal. Our goal is therefore not to attribute each observed pattern to a single training factor, but to test whether the signatures identified in controlled settings reappear in practice and whether they align with asymmetric capability profiles across task families.

Although the controlled experiments are conducted on a smaller text-only decoder, the regime mechanisms we study---concentration versus coverage expansion---are not specific to that architecture. The spectral and rank-based diagnostics used here are similarly not tied to a single modality, making them useful for testing whether regime-like structural signatures recur under more realistic multimodal training pipelines. We therefore treat the external comparison not as direct causal transfer, but as a consistency test across independently trained model families.

We focus on a set of open-source 4B-scale models spanning text-only and multimodal settings, with Qwen3-4B-Base serving as a shared reference point for parameter comparison whenever applicable. We look for three forms of external consistency: whether aggregate spectral patterns recur across models, whether layer-wise and parameter-dynamics signatures remain diagnostic outside the controlled setting, and whether these structural differences align with non-uniform benchmark profiles rather than uniform capability gains or losses.

\subsection{Cross-Family Structural Signatures}

\begin{figure}[t]
\centering
\small

\begin{subfigure}[b]{0.45\linewidth}
    \centering
    \includegraphics[width=\linewidth]{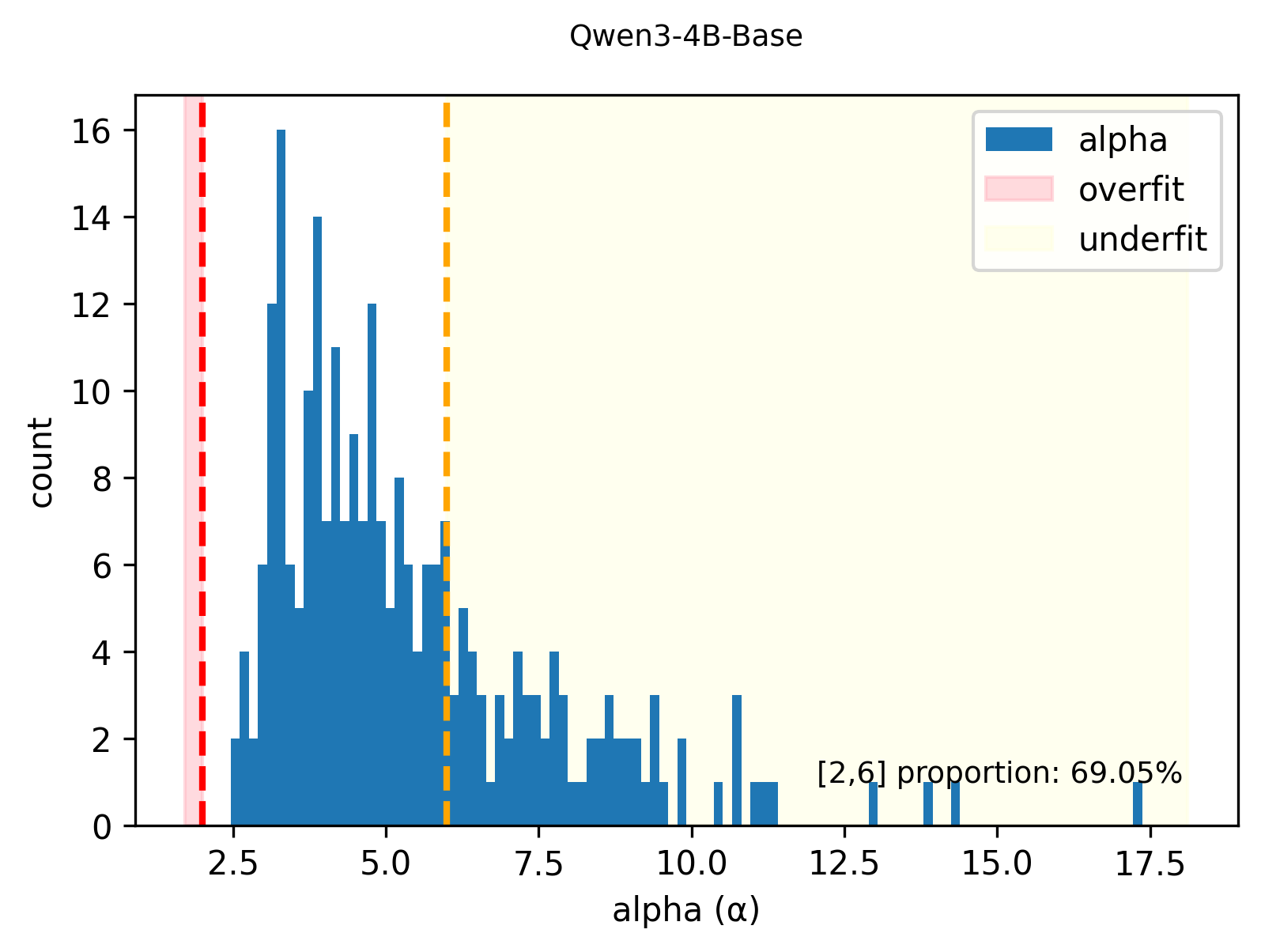}
    \caption{Qwen3-4B-Base ([2,6]: 69.05\%)}
\end{subfigure}
\hfill
\begin{subfigure}[b]{0.45\linewidth}
    \centering
    \includegraphics[width=\linewidth]{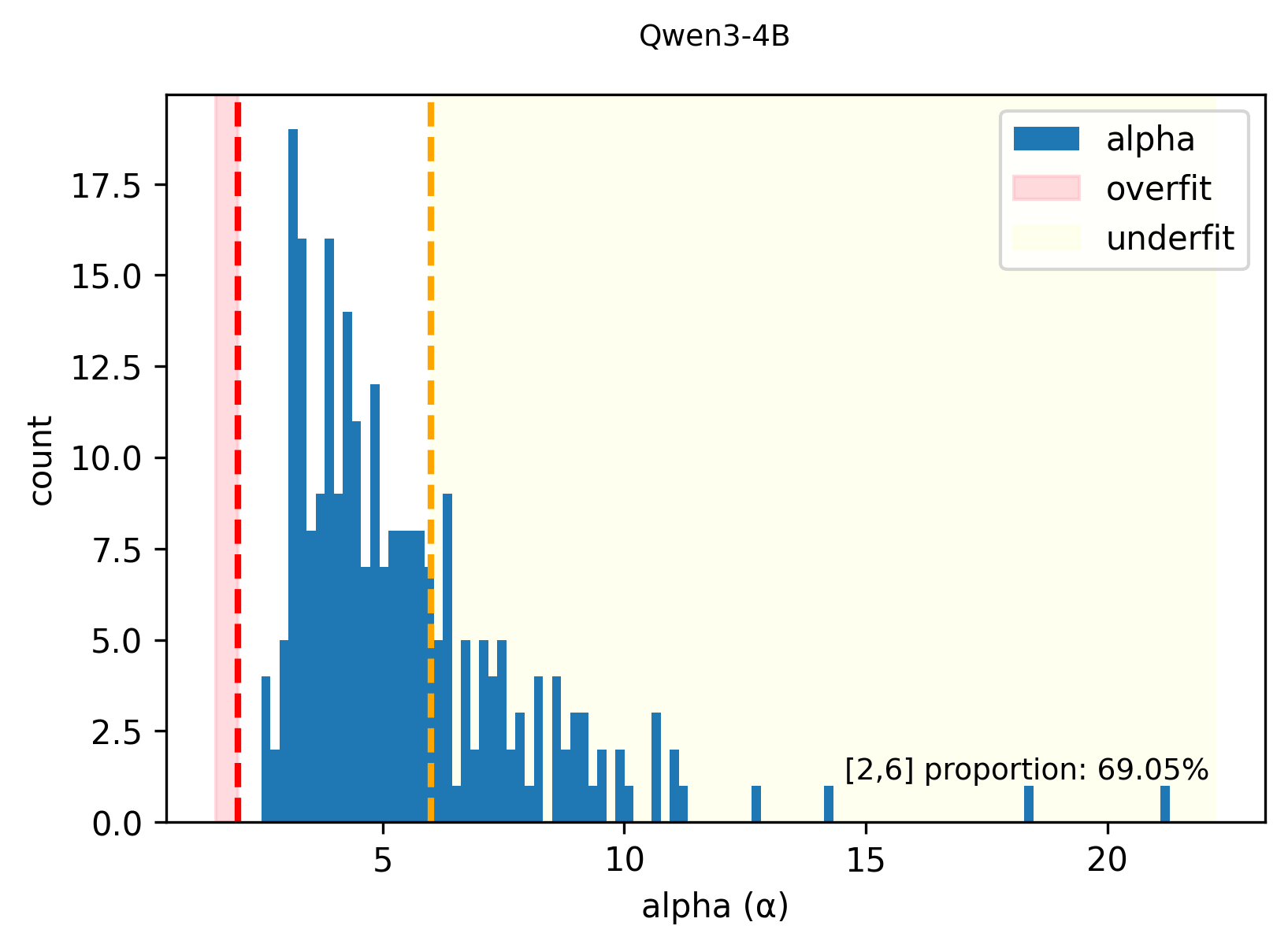 }
    \caption{Qwen3-VL-4B-Instruct ([2,6]: 69.05\%)}
\end{subfigure}

\vspace{0.3cm}

\begin{subfigure}[b]{0.45\linewidth}
    \centering
    \includegraphics[width=\linewidth]{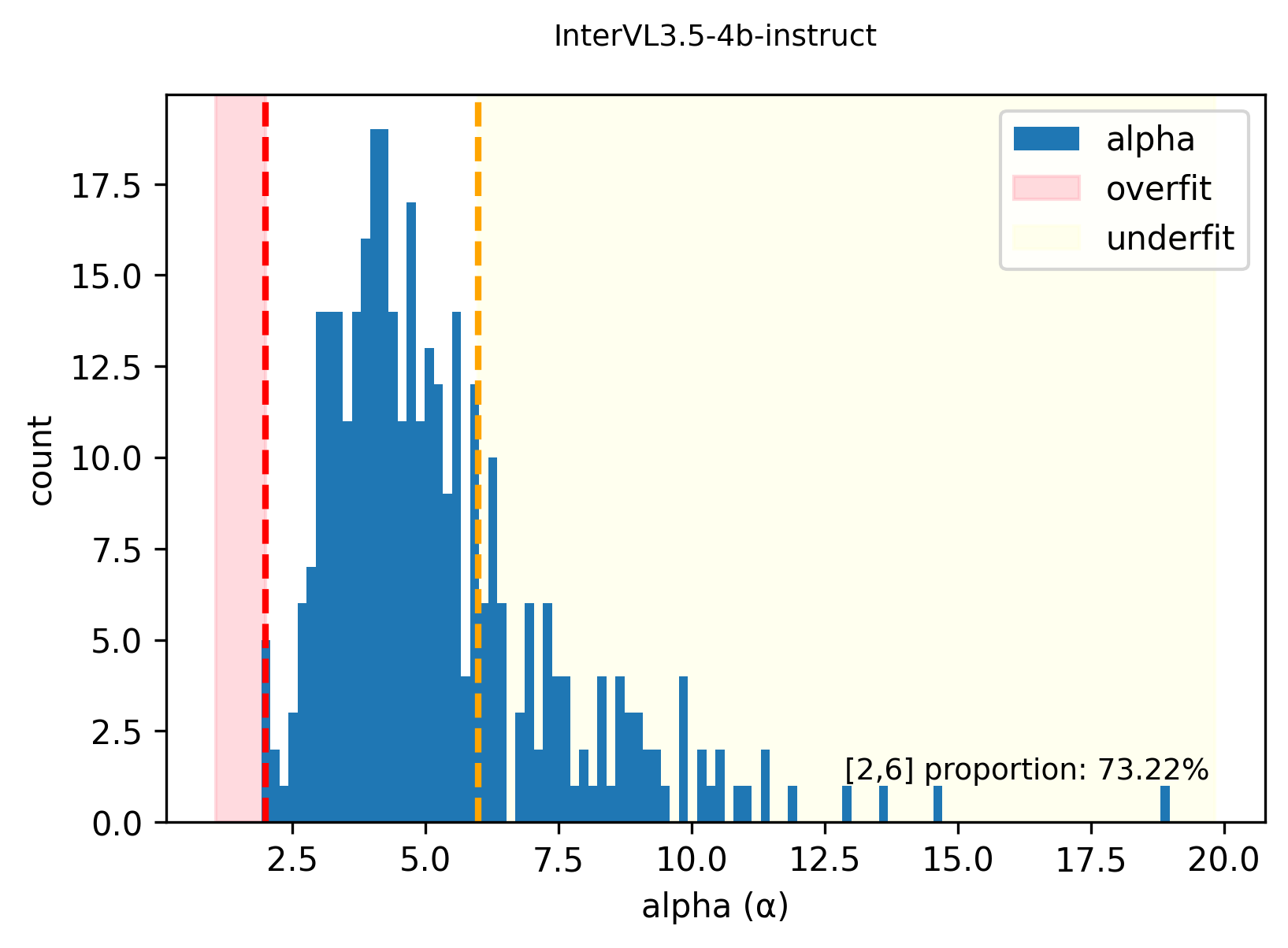}
    \caption{InternVL3.5-4B-Instruct ([2,6]: 73.22\%)}
\end{subfigure}
\hfill
\begin{subfigure}[b]{0.45\linewidth}
    \centering
    \includegraphics[width=\linewidth]{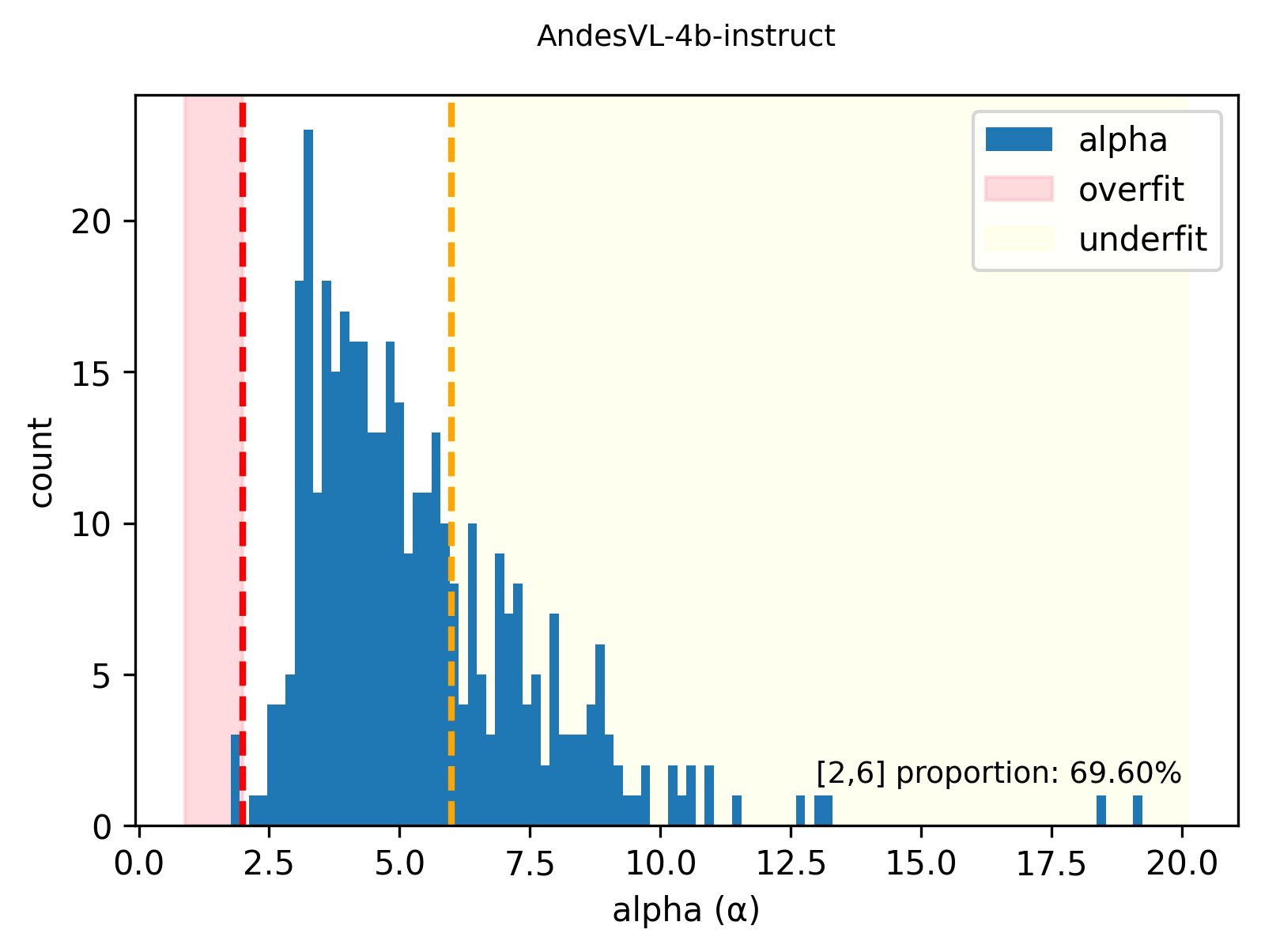}
    \caption{AndesVL-4B-Instruct ([2,6]: 69.60\%)}
\end{subfigure}

\caption{Global alpha distributions for four representative models derived from Qwen3-4B-Base. Histograms show the WeightWatcher power-law exponent ($\alpha$) across all weight matrices. Red dashed line: overfit boundary ($\alpha=2$); orange dashed line: underfit boundary ($\alpha=6$); $[2,6]$ proportion annotated per panel. (a)~Qwen3-4B-Base: tight peak at $\alpha\approx3$--$4$ with sparse outliers. (b)~Qwen3-VL-4B-Instruct: near-identical global shape despite deep decoder reparametrization. (c)~InternVL3.5-4B-Instruct: broader distribution with a secondary hump at $\alpha\approx5$--$6$, together with an elevated $[2,6]$ proportion (73.22\%). (d)~AndesVL-4B-Instruct: comparable $[2,6]$ proportion (69.60\%) but denser tail mass through $\alpha=6$--$10$, indicating spectrally degraded layers that are only weakly reflected in the aggregate metric.}
\label{fig:cross_model_alpha_hist}
\end{figure}

\begin{figure}[t]
\centering
\small
\includegraphics[width=0.75\linewidth]{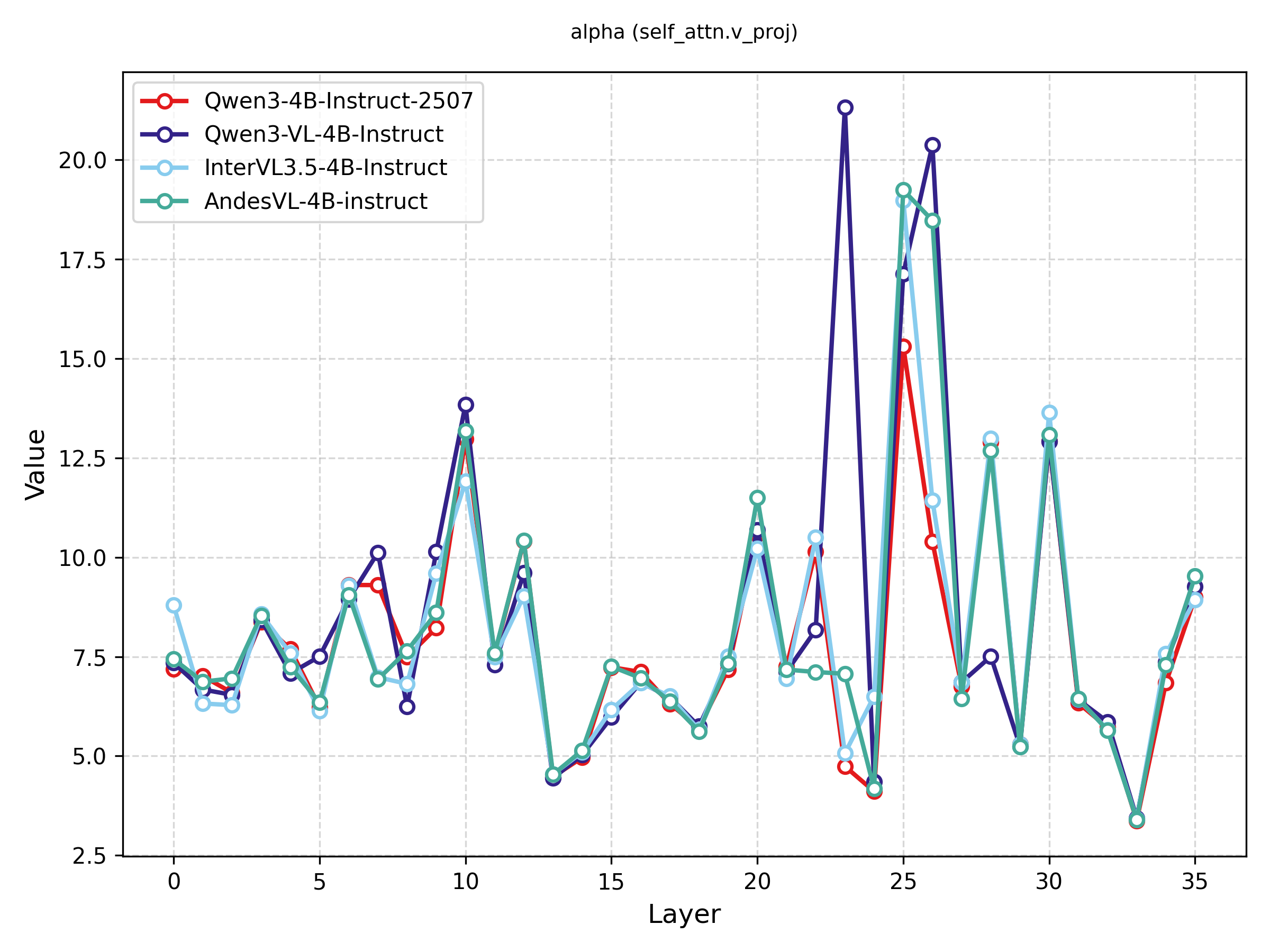}
\caption{Layer-wise alpha ($\alpha$) of \texttt{self\_attn.v\_proj} across 36 decoder layers for four instruct models. All models share a well-conditioned trough at layers 12--18 ($\alpha\approx4$--$7$) and a spike region at layers 24--27. Key divergences include an isolated spike at layer~23 for Qwen3-VL-4B-Instruct and systematic early-layer inflation for AndesVL-4B-Instruct ($\alpha\approx7.5$--$9$ at layers 0--5, versus $6.5$--$7.5$ for the other models). Qwen3-4B-Instruct (LLM-only) remains the flattest, suggesting that instruction tuning without VL data is comparatively spectrally conservative. The concentration of cross-model divergences in \texttt{v\_proj}---with much weaker separation in \texttt{mlp.up\_proj} (see Appendix)---suggests that the attention value projection is the clearest locus of cross-model VL adaptation differences.}
\label{fig:cross_model_attn_alpha_instruct}
\end{figure}

\begin{figure}[t]
\centering
\small
\includegraphics[width=0.75\linewidth]{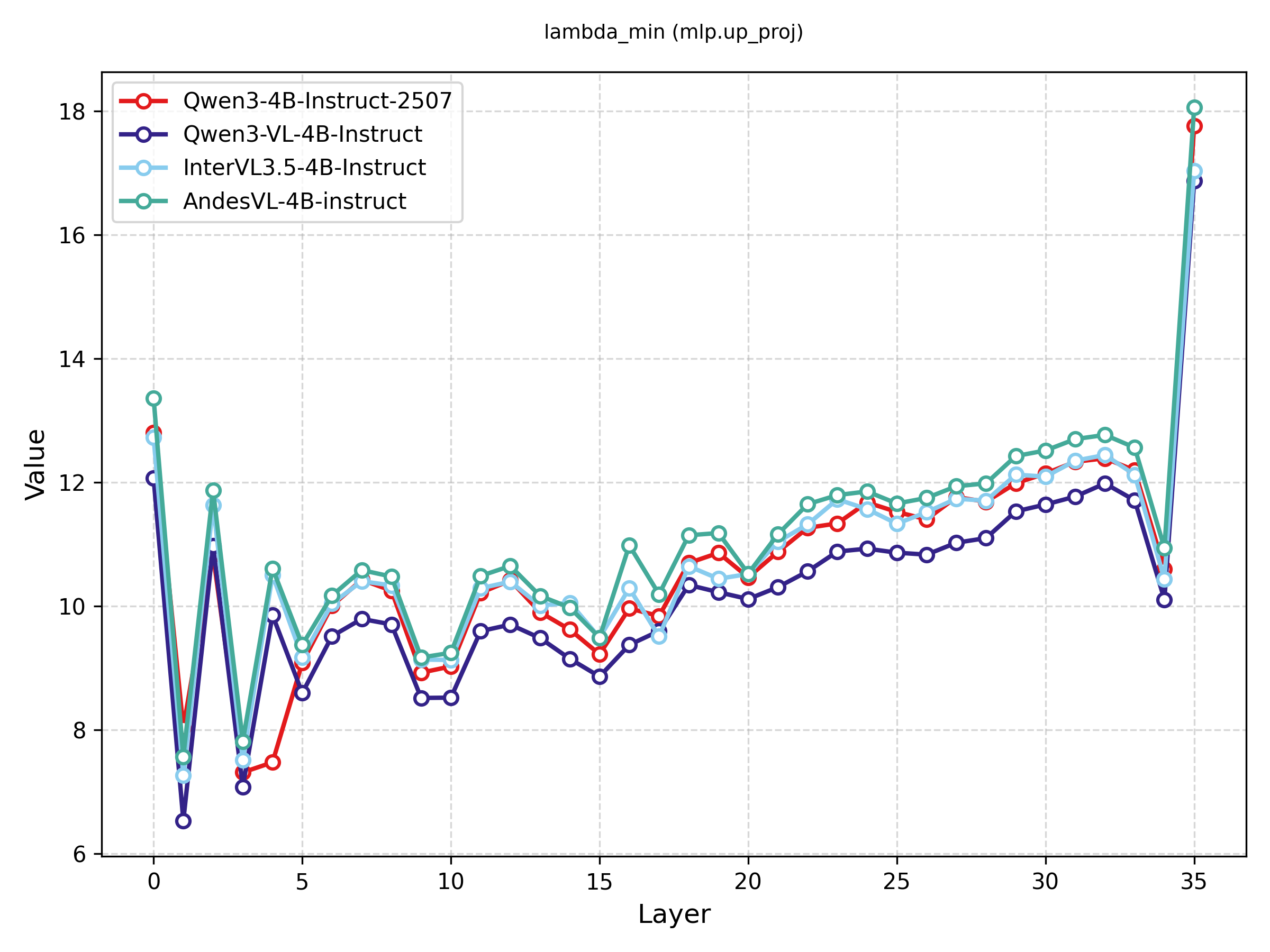}
\caption{Layer-wise $\lambda_{\min}$ of \texttt{mlp.up\_proj} across 36 decoder layers for four instruct models. All models share a monotonically rising trend from $\sim$8 (early layers) to $\sim$13 (layer~33), with a terminal spike at layer~35, but a consistent rank order separates them: Qwen3-VL-4B-Instruct sits lowest, followed by Qwen3-4B-Instruct, InternVL3.5-4B-Instruct, and AndesVL-4B-Instruct highest. The separation widens from roughly $\sim$0.5 in early layers to $\sim$1.5 beyond layer~20. This rank order is stable across instruct and thinking variants (see Appendix). Since alpha and effective feature number in \texttt{mlp.up\_proj} are nearly identical across models (see Appendix), $\lambda_{\min}$ is the main differentiating metric in the MLP pathway: lower values are consistent with broader spectral resolution, whereas higher values indicate a more compressed operating regime.}
\label{fig:cross_model_mlp_lam_min_instruct}
\end{figure}

\begin{figure}[t]
\centering
\small
\includegraphics[width=0.75\linewidth]{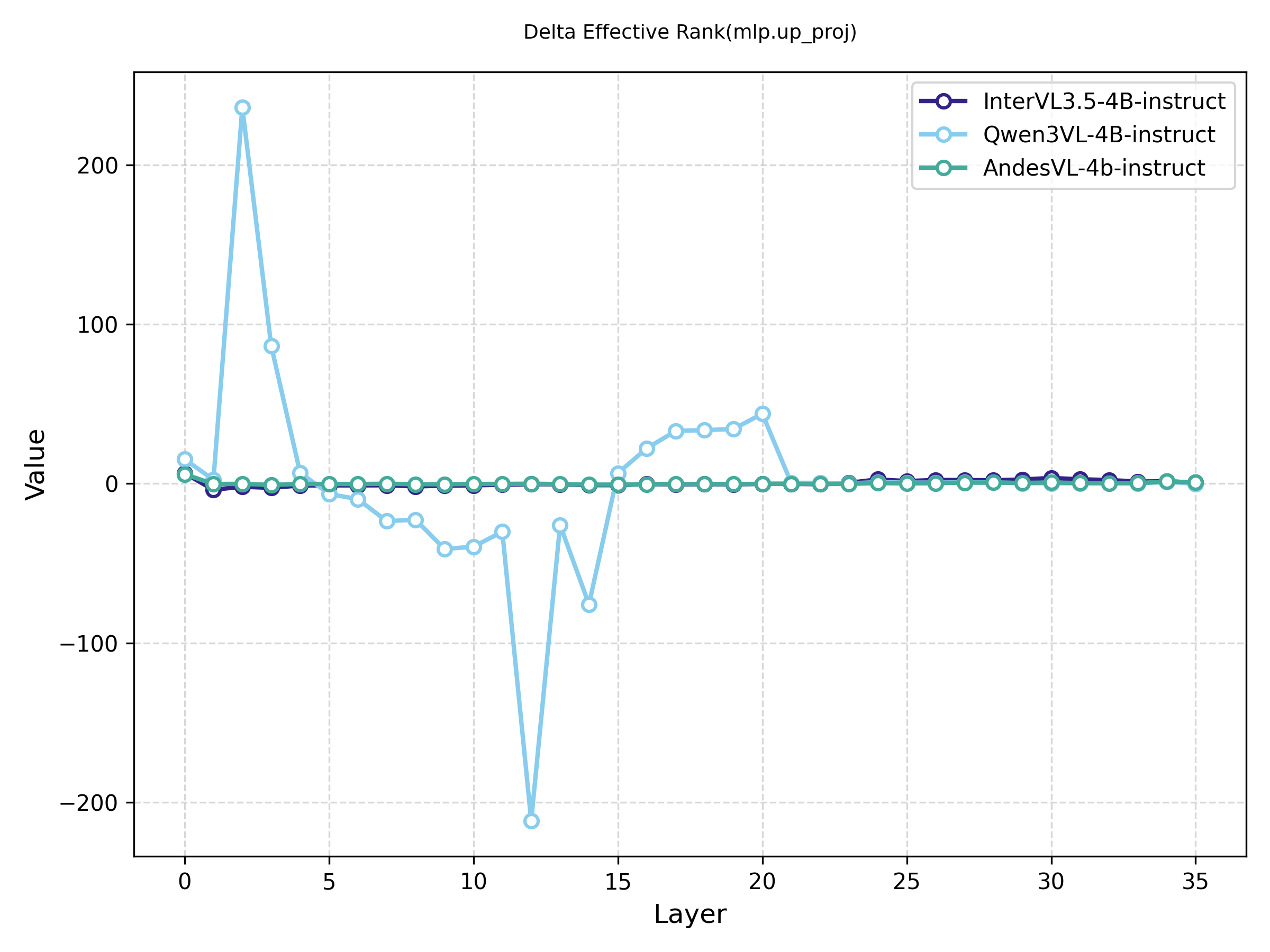}
\caption{Delta effective rank in \texttt{mlp.up\_proj} between instruct and thinking checkpoints for four models. Qwen3-VL-4B shows large layer-specific restructuring, with both positive and negative spikes, indicating targeted rank expansion and contraction. InternVL3.5-4B shows similarly strong oscillations. AndesVL-4B-Instruct remains near zero across all layers: thinking training changes parameter magnitudes (see Appendix) but leaves rank geometry comparatively unchanged. This spectral inertness under reasoning alignment is more consistent with structurally limited adaptation than with broad representational restructuring.}
\label{fig:cross_model_param_delta_mlp_delta_eff_rank_instruct}
\end{figure}

\begin{figure}[t]
\centering
\small
\includegraphics[width=0.75\linewidth]{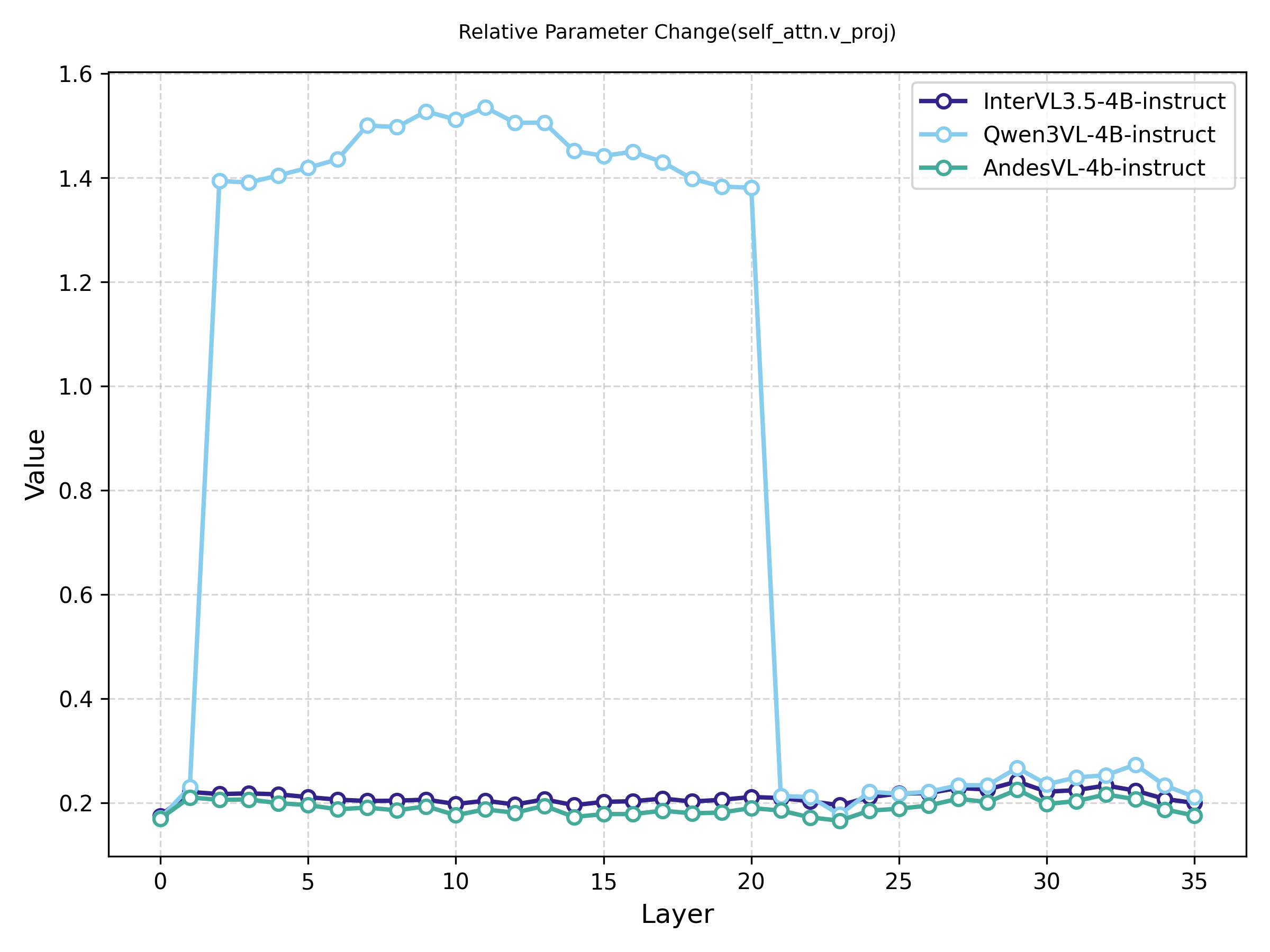}
\caption{Relative parameter change in \texttt{self\_attn.v\_proj} measured against the shared ancestor Qwen3-4B-Base for three MLLM instruct models. Qwen3-VL-4B-Instruct exhibits a bimodal structure: layers 2--20 are modified by roughly $\sim$140--150\% relative to the base, while layers 0--1 and 21--35 remain near $\sim$20\%, with a sharp transition at layer~21. InternVL3.5-4B-Instruct and AndesVL-4B-Instruct remain close to $\sim$20\% across all layers. This large gap suggests three broad modification regimes from the shared base: minimal perturbation (InternVL base, see Appendix), conservative fine-tuning (InternVL/AndesVL instruct), and deep reparametrization (Qwen3-VL layers 2--20).}
\label{fig:cross_model_param_delta_attn_rel_instruct}
\end{figure}

\begin{figure}[t]
\centering
\small
\includegraphics[width=0.75\linewidth]{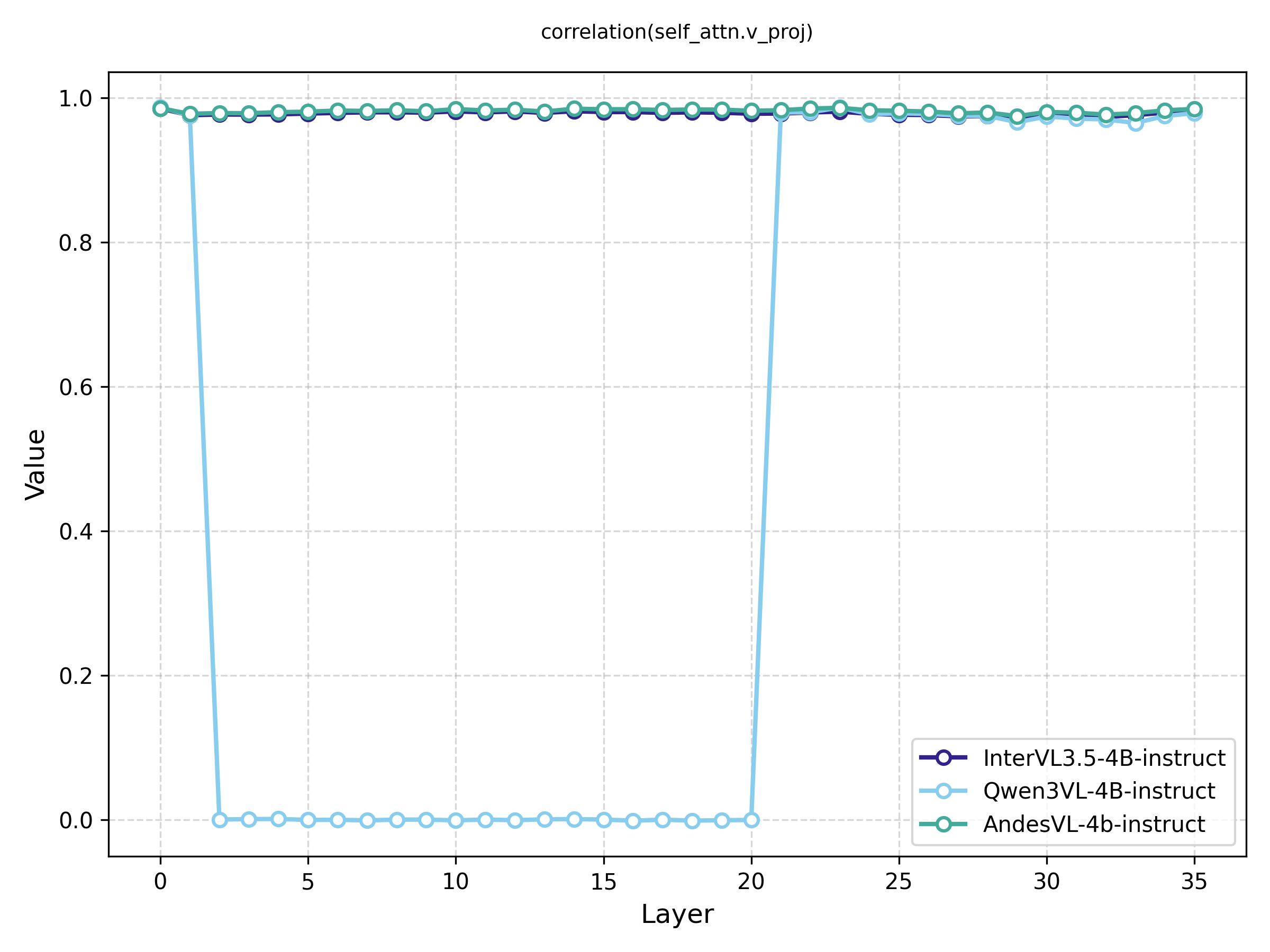}
\caption{Weight correlation with Qwen3-4B-Base in \texttt{self\_attn.v\_proj} for three MLLM instruct models. InternVL3.5-4B-Instruct and AndesVL-4B-Instruct maintain $r\approx0.97$--$0.98$ at every layer, preserving much of the original backbone structure. Qwen3-VL-4B-Instruct drops to near-zero correlation at layers 2--20, then recovers to $\sim$0.97 at layers 21+, mirroring the transition in Figure~\ref{fig:cross_model_param_delta_attn_rel_instruct}. The simultaneous near-zero correlation and very large relative change is more consistent with deep reparametrization than with incremental fine-tuning within the modified zone.}
\label{fig:cross_model_param_delta_attn_corr_instruct}
\end{figure}

\begin{figure}[t]
\centering
\small
\includegraphics[width=0.75\linewidth]{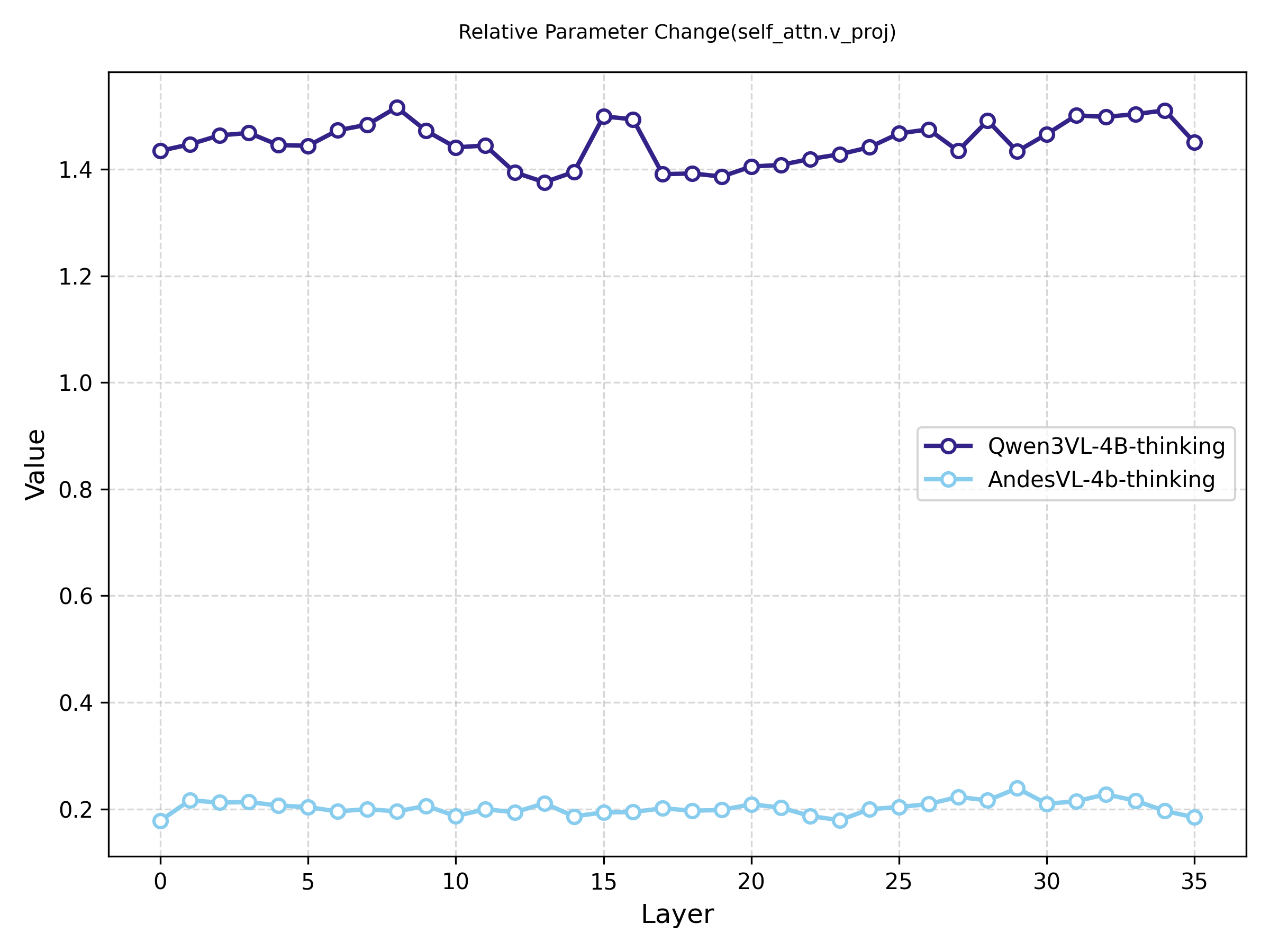}
\caption{Relative parameter change in \texttt{self\_attn.v\_proj} from Qwen3-4B-Base for two MLLM thinking models. Qwen3-VL-4B-Thinking shows roughly $\sim$140--150\% relative change across all 36 layers, indicating that the transition at layer~21 observed in the instruct variant has disappeared. AndesVL-4B-Thinking remains near $\sim$20\%, unchanged from its instruct variant. The persistent gap suggests that the modification regime is largely established during VL training and is not substantially altered by later reasoning alignment.}
\label{fig:cross_model_param_delta_attn_rel_think}
\end{figure}

We begin with model-level spectral statistics. Figure~\ref{fig:cross_model_alpha_hist} shows the distribution of $\alpha$ values across weight matrices for representative models. At the aggregate level, all four models occupy a relatively narrow range in the proportion of layers with $\alpha \in [2,6]$ (69.0\%--73.2\%), indicating that model-level summaries alone are insufficient to distinguish internal training behavior. The main differences instead lie in the shape of the distribution tails. Qwen3-4B-Base and Qwen3-VL-4B-Instruct exhibit concentrated distributions with sparse high-$\alpha$ outliers, whereas AndesVL-4B-Instruct shows a visibly denser tail in the $\alpha=6$--10 range, indicating the presence of spectrally degraded layers that are only weakly reflected in the aggregate $[2,6]$ proportion. InternVL3.5-4B-Instruct occupies an intermediate position, with a broader distribution and a secondary hump near $\alpha\approx5$--6. These patterns mirror the main lesson from Section~\ref{controlled_data}: aggregate conditioning can conceal meaningful regime-level differences in internal structure.

We next examine layer-wise spectral behavior in the decoder. Figure~\ref{fig:cross_model_attn_alpha_instruct} shows that the strongest cross-model divergences concentrate in \texttt{self\_attn.v\_proj}. Qwen3-4B-Instruct, which does not undergo multimodal adaptation, remains the flattest across depth. Qwen3-VL-4B-Instruct is broadly smooth but exhibits an isolated spike at layer~23. AndesVL-4B-Instruct shows systematic early-layer inflation and stronger heterogeneity across depth, while InternVL3.5-4B-Instruct lies between these extremes. This concentration of cross-model differences in the attention value projection is consistent with the controlled findings, where regime-specific distortions also emerged most clearly in attention pathways rather than in model-level summaries alone.

In contrast, the feed-forward pathway is much more stable. As shown in Figure~\ref{fig:cross_model_mlp_lam_min_instruct}, $\alpha$ and effective feature number in \texttt{mlp.up\_proj} are nearly invariant across models, while $\lambda_{\min}$ provides the main differentiating signal. The observed rank order is stable: Qwen3-VL-4B-Instruct has the lowest $\lambda_{\min}$, followed by Qwen3-4B-Instruct, then InternVL3.5-4B-Instruct and AndesVL-4B-Instruct at higher values. We interpret this pattern conservatively: lower $\lambda_{\min}$ is consistent with a broader spectral floor and richer representational resolution, whereas higher values indicate a more compressed operating regime. Together, these results suggest that cross-family regime differences are expressed primarily through attention-layer adaptation, with MLP differences appearing in a narrower but still informative form.

\paragraph{Summary.}
The structural signatures identified in controlled experiments recur in recognizable form across real model families. However, the recurring signal is not captured well by aggregate metrics alone. Instead, it appears most clearly in layer-wise attention statistics and in tail behavior of spectral distributions, reinforcing the need for fine-grained diagnostics when comparing training outcomes.

\subsection{Parameter Modification Depth}

\begin{figure}[t]
\centering
\small
\includegraphics[width=0.75\linewidth]{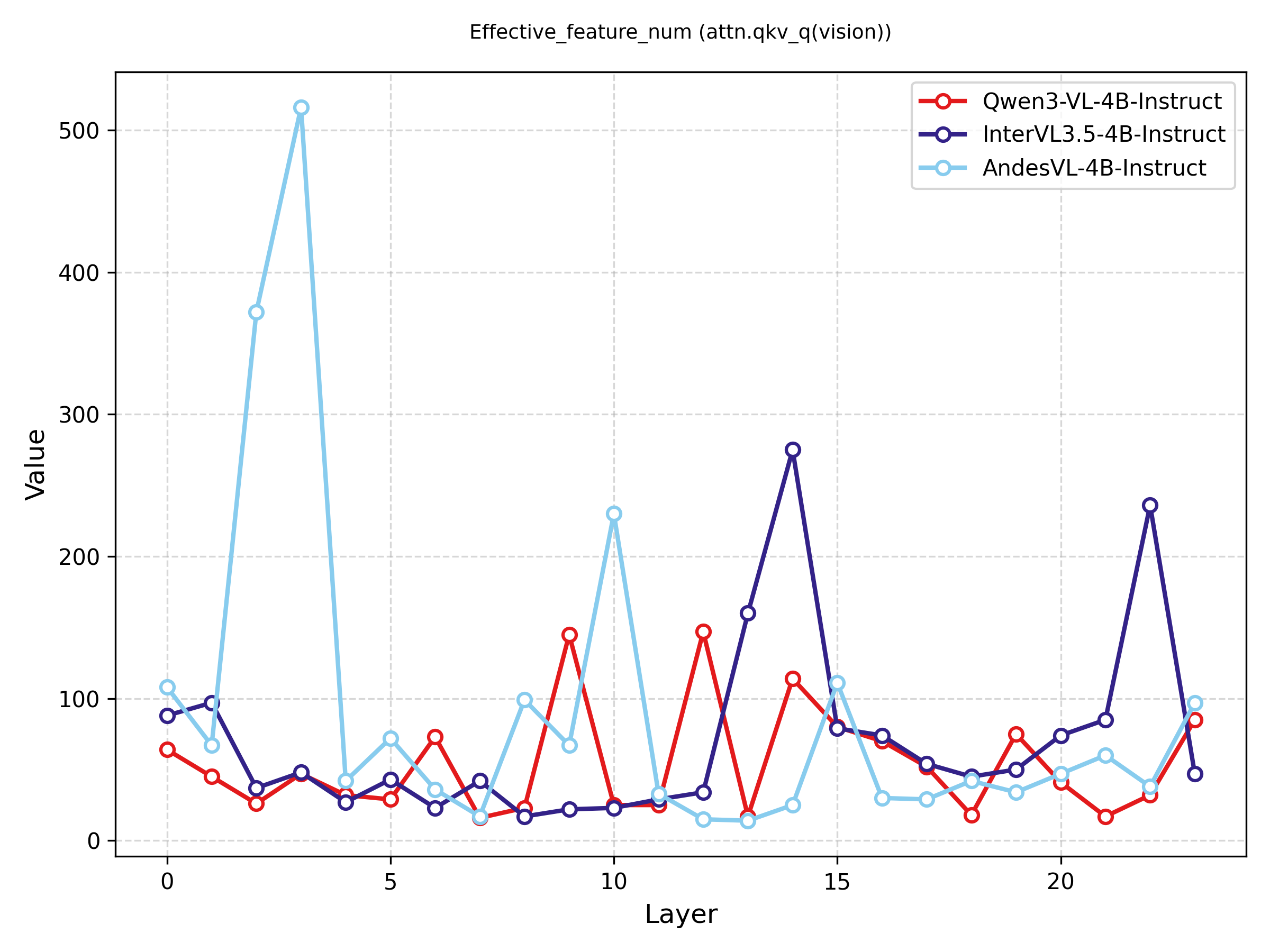}
\caption{Layer-wise effective feature number of the vision encoder attention query weight (\texttt{attn.qkv\_q}) across 24 layers for three MLLM instruct models. AndesVL-4B-Instruct exhibits a sharp early-layer spike at layer~3 ($\sim$520), substantially larger than the peak values of Qwen3-VL-4B-Instruct ($\sim$145 at layer~10) and InternVL3.5-4B-Instruct ($\sim$275 at layer~14). At the same layer, AndesVL also shows very low $\alpha$ and near-zero $\lambda_{\min}$ (see Appendix), consistent with a flat singular-value spectrum and poorly differentiated representations. A secondary spike at layer~10 further distinguishes AndesVL from the other models. This pattern also appears in the corresponding thinking variant, suggesting that it is established during the initial VL training stage rather than repaired by subsequent reasoning alignment.}
\label{fig:cross_model_spec_cmp_q_eff_feat_instruct}
\end{figure}

To quantify how strongly different models depart from the shared base, we compare parameters against Qwen3-4B-Base. Rather than treating all adaptation as a single continuum, the resulting patterns suggest three broad regimes of backbone modification: \emph{minimal perturbation}, \emph{conservative fine-tuning}, and \emph{deep reparametrization}. These are descriptive categories rather than causal claims; they summarize how far the learned parameters move from the original backbone and how much of the original structure is preserved.

Figure~\ref{fig:cross_model_param_delta_attn_rel_instruct} shows the relative parameter change in \texttt{self\_attn.v\_proj} for three multimodal instruct models. InternVL3.5-4B-Instruct and AndesVL-4B-Instruct both remain near $\sim$20\% across depth, consistent with a conservative fine-tuning regime. By contrast, Qwen3-VL-4B-Instruct exhibits a bimodal pattern: layers 2--20 differ from the base by roughly $140$--$150\%$, while layers 0--1 and 21--35 remain much closer to baseline. The corresponding correlation analysis in Figure~\ref{fig:cross_model_param_delta_attn_corr_instruct} sharpens this picture: InternVL3.5 and AndesVL maintain $r\approx0.97$--$0.98$ throughout, whereas Qwen3-VL drops to near-zero correlation across the same 2--20 layer range. This combination of very large relative change and near-zero correlation is more consistent with deep reparametrization than with incremental fine-tuning.

The difference persists under reasoning alignment. As shown in Figure~\ref{fig:cross_model_param_delta_attn_rel_think}, Qwen3-VL-4B-Thinking extends this deep reparametrization across all 36 layers, while AndesVL-4B-Thinking remains close to the same conservative regime as its instruct counterpart. We interpret this cautiously: the depth of adaptation appears to be largely established during the primary multimodal training stage and is not substantially altered by later reasoning alignment in all model families. This does not by itself identify the cause of the difference, but it does show that post-training alignment does not erase the structural distinction.

We also examine representational restructuring through delta effective rank between instruct and thinking checkpoints. Figure~\ref{fig:cross_model_param_delta_mlp_delta_eff_rank_instruct} shows that Qwen3-VL-4B and InternVL3.5-4B undergo substantial layer-specific changes, with both positive and negative spikes that suggest non-uniform reallocation of representational capacity. AndesVL-4B-Instruct, in contrast, remains near zero across all layers. We refer to this pattern as \emph{spectral inertness}: parameters may continue to change in magnitude, but representational geometry remains comparatively unchanged across depth. In the language of Section~\ref{controlled_data}, this behavior is more consistent with structurally limited adaptation than with broad representational restructuring.

Beyond the decoder, Figure~\ref{fig:cross_model_spec_cmp_q_eff_feat_instruct} shows that regime-like irregularity can also appear in the vision encoder. AndesVL exhibits a sharp early-layer anomaly in the attention query projection, together with a weaker secondary spike at a later layer, while the corresponding decoder remains in a conservative modification regime. This cross-submodule decoupling suggests that structural distortion in multimodal systems need not be distributed uniformly across components: some models may retain comparatively shallow decoder modification while localizing stronger irregularity in modality-specific submodules.

\paragraph{Summary.}
Across real model families, adaptation depth is highly non-uniform. Some models remain close to the shared base under conservative fine-tuning, whereas others undergo deep reparametrization over large portions of the decoder. These differences do not by themselves establish which training choice caused them, but they show that the controlled signatures of shallow versus deeply restructured adaptation recur in practice.

\subsection{Capability Asymmetry Across Benchmarks}
\label{sec:benchmark_validation}

\begin{table*}[t]
\centering
\caption{Benchmark comparison of 4B-scale vision-language models. * are from AndesVL's evaluation}
\label{tab:benchmark_comparison}
\small
\begin{tabular}{lccc}
\toprule
\textbf{Benchmark} & \textbf{Qwen3-VL-4B Instruct} & \textbf{InternVL3.5-4B} & \textbf{AndesVL-4B-Instruct} \\
\midrule
MMMU (val)                  & 67.4 & 66.6            & 58.0  \\
MMMU-Pro                    & 53.2 & 53.5            & 37.6  \\
MathVista (mini)            & 73.7 & 77.1            & 73.3  \\
MathVision                  & 51.6 & 54.4            & 27.1  \\
MathVerse (vision-only)     & 46.8 & 61.7            & 34.3  \\
DynaMath                    & 65.3 & 35.7            & 21.2  \\
LogicVista                  & 53.2 & 56.4            & 41.6  \\
WeMath                      & --   & 50.1            & 33.7  \\
AI2D (w M)                  & 84.1 & 82.6            & 84.5  \\
ChartQA (test)              & 84.6 & 86              & 87.8  \\
TextVQA (val)               & --   & 77.9            & 81.6  \\
DocVQA (test)               & 95.3 & 92.4            & 96    \\
InfoVQA (test)              & 80.3 & 78              & 81    \\
OCRBench                    & 881  & 822             & 861   \\
MMBench v1.1 (EN)           & 83.9 & 80.3 *           & 81.2  \\
MMStar                      & 69.8 & 65              & 66.1  \\
RealWorldQA                 & 70.9 & 66.3            & 72.2  \\
MMVet                       & --   & 76.6            & 61.2  \\
MME (sum)                   & --   & 2272.3            & 2345  \\
HallusionBench              & 57.6 & 44.8            & 54.7  \\
POPE (avg)                  & --   & 88.9            & 88.5  \\
BLINK                       & 65.8 & 58.1            & 58.2  \\
MuirBench                   & 63.8 & 53.1            & 55.5  \\
ScreenSpot                  & --   & 83.6            & 84.3  \\
ScreenSpot-v2               & --   & 85.1            & 86.1  \\
ScreenSpot Pro              & 59.5 & 18.1 *           & 28.2  \\
ERQA                        & 41.3 & 38.5            & --    \\
VSI-Bench                   & 59.3 & 54.9            & --    \\
MVBench                     & 68.9 & 71.2            & --    \\
Video-MME (wo sub)          & 69.3 & 65.4 *           & --    \\
MLVU (M-Avg)                & 75.3 & 70.4            & --    \\
RefCOCO-avg                 & 89   & 92.5/94.3/88.2  & --    \\
\bottomrule
\end{tabular}
\end{table*}

Structural differences are only meaningful if they align with observable external behavior. We therefore compare benchmark performance across representative 4B-scale multimodal models, not to claim that spectral structure alone determines capability, but to test whether different structural regimes co-occur with different distributions of task performance. The central question is whether external performance appears broad and transferable, or instead concentrated on a narrower subset of benchmarks.

The benchmark results are highly non-uniform across task families. Qwen3-VL-4B-Instruct shows the broadest high-performing profile among the reported results, especially on reasoning-intensive evaluations such as MMMU-Pro, MathVision, and DynaMath. InternVL3.5-4B remains competitive, particularly on mathematical and structured reasoning tasks. AndesVL-4B-Instruct, however, exhibits a markedly asymmetric profile: it performs strongly on more perception- and document-oriented tasks such as ChartQA, DocVQA, AI2D, RealWorldQA, and OCRBench, but shows large deficits on reasoning-heavy benchmarks including MMMU-Pro, MathVision, and DynaMath. This asymmetry is the key result. The relevant contrast is not simply ``high score'' versus ``low score,'' but whether performance remains broad across task types or concentrates on benchmarks that are more distribution-matched to the training regime.

Viewed together with the structural analyses above, these results are consistent with the regime-centric interpretation developed earlier in the paper. Models exhibiting broader or deeper representational restructuring also tend to maintain stronger performance on reasoning-intensive and compositionally demanding tasks. Models with more limited backbone modification can still achieve competitive results on tasks that are closer to the dominant training distribution, but their capability profile is narrower. We stress again that this is an association rather than a proof of mechanism. Nevertheless, the alignment between structural signatures and benchmark asymmetry provides external support for the claim that benchmark gains do not necessarily imply uniform capability growth.

\paragraph{Summary.}
Across independently trained models, benchmark performance exhibits structured asymmetry rather than uniform improvement. This pattern aligns with the external-validation role of this section: the regime-like signatures found in controlled interventions recur in real systems and are associated with uneven capability allocation across task families.

\subsection{Discussion}

Taken together, the analyses in this section provide external consistency rather than causal identification. The controlled experiments in Section~\ref{controlled_data} showed that concentrated training can induce distinct and partially persistent structural signatures. Here, we observe that analogous signatures recur across real model families, especially in attention-layer adaptation depth, spectral tail behavior, the presence or absence of representational restructuring during later alignment, and localized irregularity in the vision encoder. These structural differences, in turn, align with asymmetric benchmark profiles rather than uniform capability gains.

The broader implication is that independently trained models need not fail in the same way to reflect a similar underlying regime pressure. Some exhibit shallow but stable adaptation, others deep reparametrization, and others localized degradation in specific submodules such as the vision encoder. What unifies them is not a single visible failure mode, but the recurrence of internal signatures that resemble the patterns isolated under controlled data interventions. This supports the broader claim of the paper: benchmark shadows are better understood as regime-level distortions in representation formation than as a simple matter of benchmark contamination or score inflation alone.

\section{Case Study: Prompt Repetition as a Distributional Artifact}
\label{case_study}

The regime framework developed above suggests that not all forms of data redundancy have the same effect on learning. To illustrate this boundary, we examine prompt repetition in vision-language (VL) training data as a practical case of distributional artifact. Unlike the concentrated regimes in Section~\ref{controlled_data}, prompt duplication increases local repetition without substantially narrowing semantic support, making it a useful contrast case for the regime-centric view.

\subsection{Prompt Repetition in VL Pipelines}

In multimodal training corpora, instruction prompts are often duplicated across multiple fields, including system messages, user inputs, captions, metadata, and OCR-derived text. These repetitions arise naturally from data collection and preprocessing pipelines. While such datasets may appear diverse at the sample level, prompt duplication introduces redundancy at the structural level by repeatedly exposing the model to identical or near-identical instruction patterns. This makes prompt repetition a useful contrast case: it introduces template-level redundancy while largely preserving the underlying semantic supervision.

\subsection{Intervention}

We construct a modified dataset by removing duplicated prompt segments while preserving semantic content. Specifically, we define duplicated prompt segments as verbatim instruction strings or template fragments repeated across multiple fields within the same sample. The de-duplication procedure preserves one canonical instance of each segment while removing repeated copies, leaving the image, task target, and core semantic supervision unchanged. We then maintain the overall dataset size and task distribution so that the intervention primarily affects prompt redundancy rather than coverage.

This results in two conditions:
\begin{itemize}
\item \textbf{Baseline}: original dataset with repeated prompts (approximately 75\% duplication)
\item \textbf{De-duplicated condition}: modified dataset with redundant prompt segments removed
\end{itemize}

\subsection{Spectral Effects of Duplication}

\begin{figure}[t]
\centering

\begin{subfigure}[b]{0.45\linewidth}
    \centering
    \includegraphics[width=\linewidth]{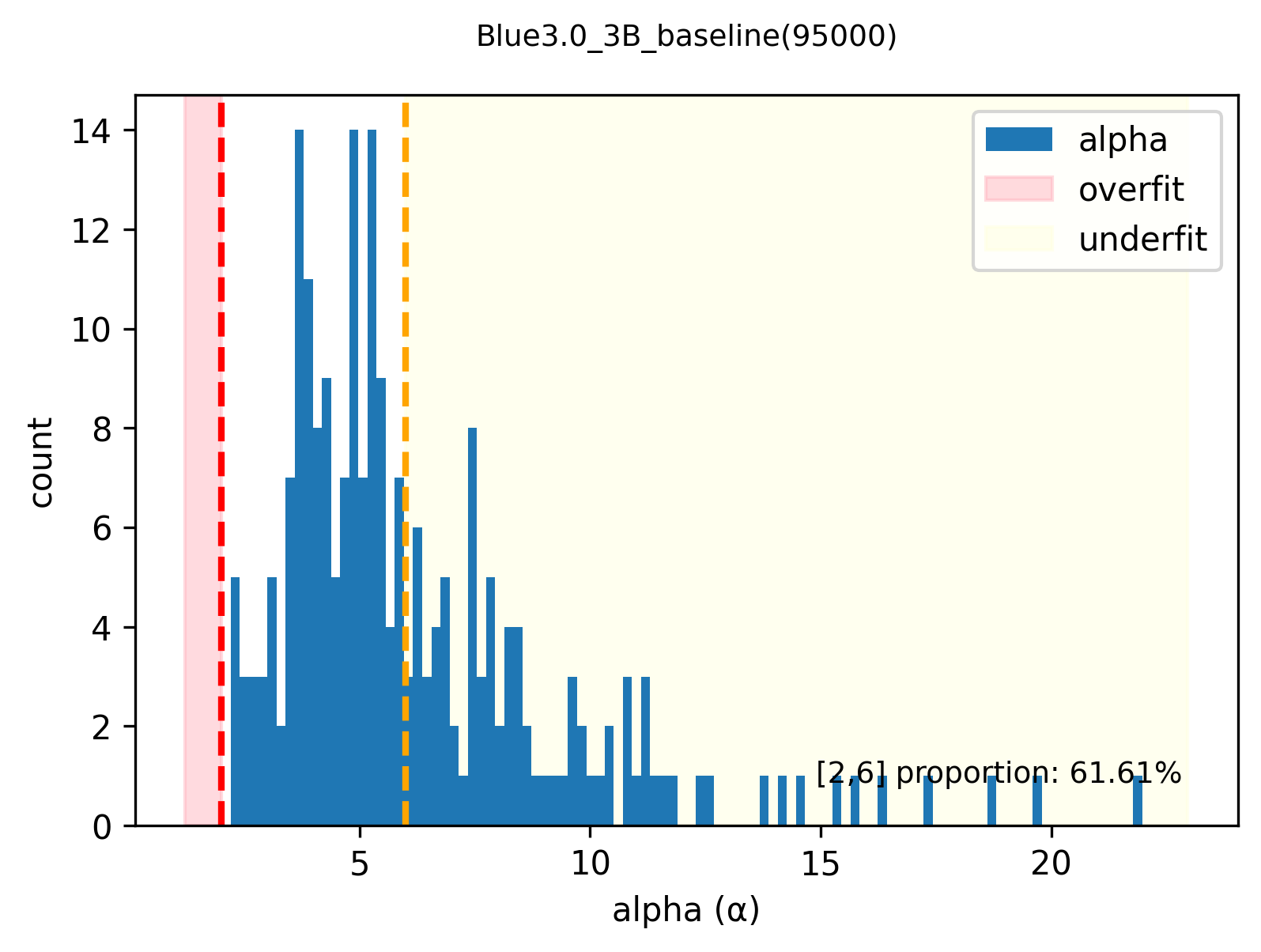}
    \caption{Baseline (with prompt duplication). $\alpha \in [2,6]$: 61.61\%.}
\end{subfigure}
\hfill
\begin{subfigure}[b]{0.45\linewidth}
    \centering
    \includegraphics[width=\linewidth]{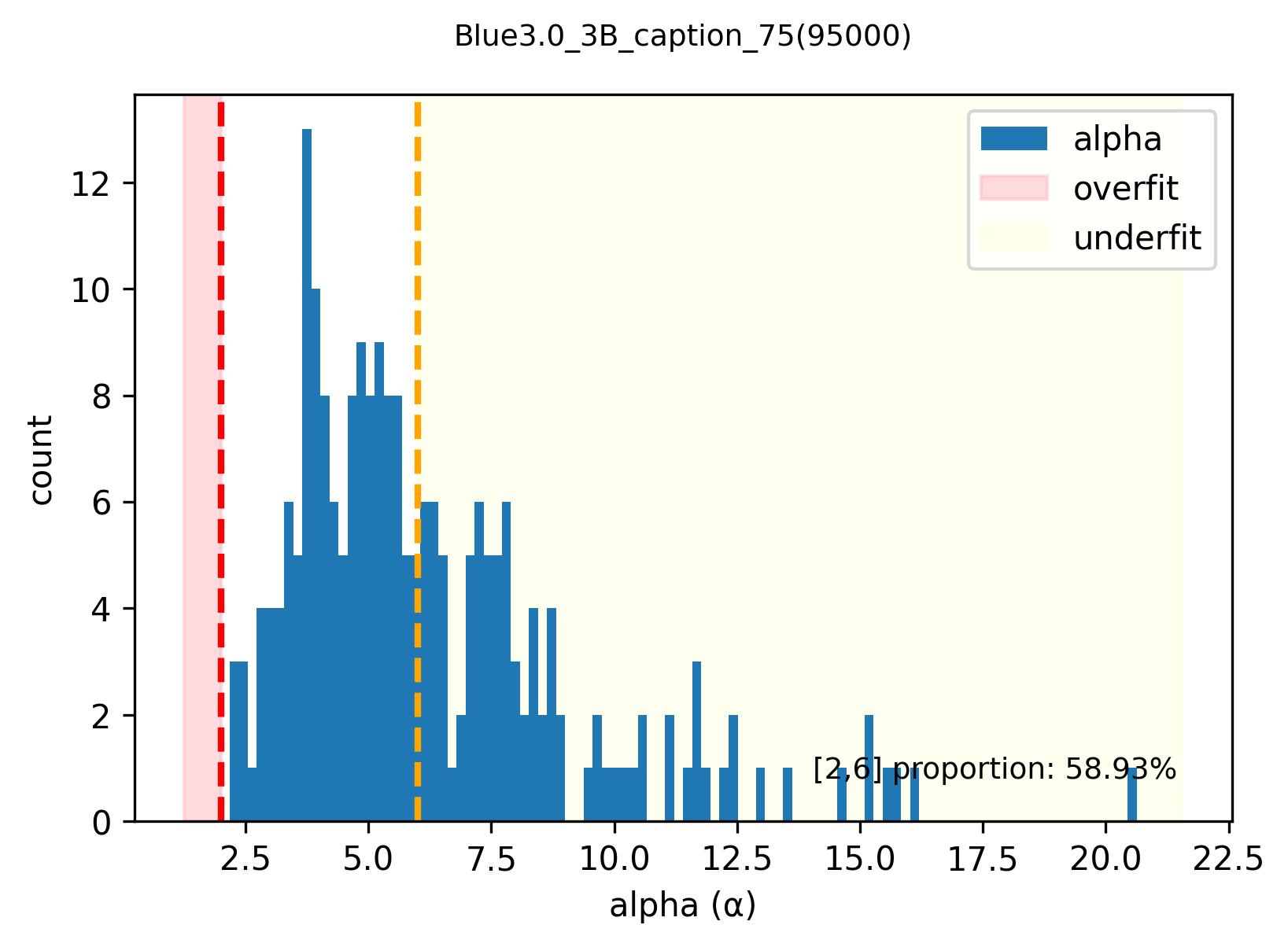}
    \caption{De-duplicated variant (75\% of repeated prompts removed). $\alpha \in [2,6]$: 58.93\%.}
\end{subfigure}

\caption{Model-level $\alpha$ distributions for the baseline with prompt duplication (left) and the de-duplicated condition with 75\% of repeated prompts removed (right), both at checkpoint 95{,}000. The dashed red line marks $\alpha = 2$ (overfit boundary) and the dashed orange line marks $\alpha = 6$ (underfit boundary). The proportion of well-conditioned layers ($\alpha \in [2,6]$) is 61.61\% for the duplicated baseline and 58.93\% for the de-duplicated variant. Both distributions retain similar overall shape and tail behavior, indicating that prompt repetition introduces only mild model-level spectral differences rather than a regime-level shift in global conditioning.}
\label{fig:dedup_mdl_lvl_base_vs_dedup}

\end{figure}

\begin{figure}[t]
\centering
\small
\includegraphics[width=0.75\linewidth]{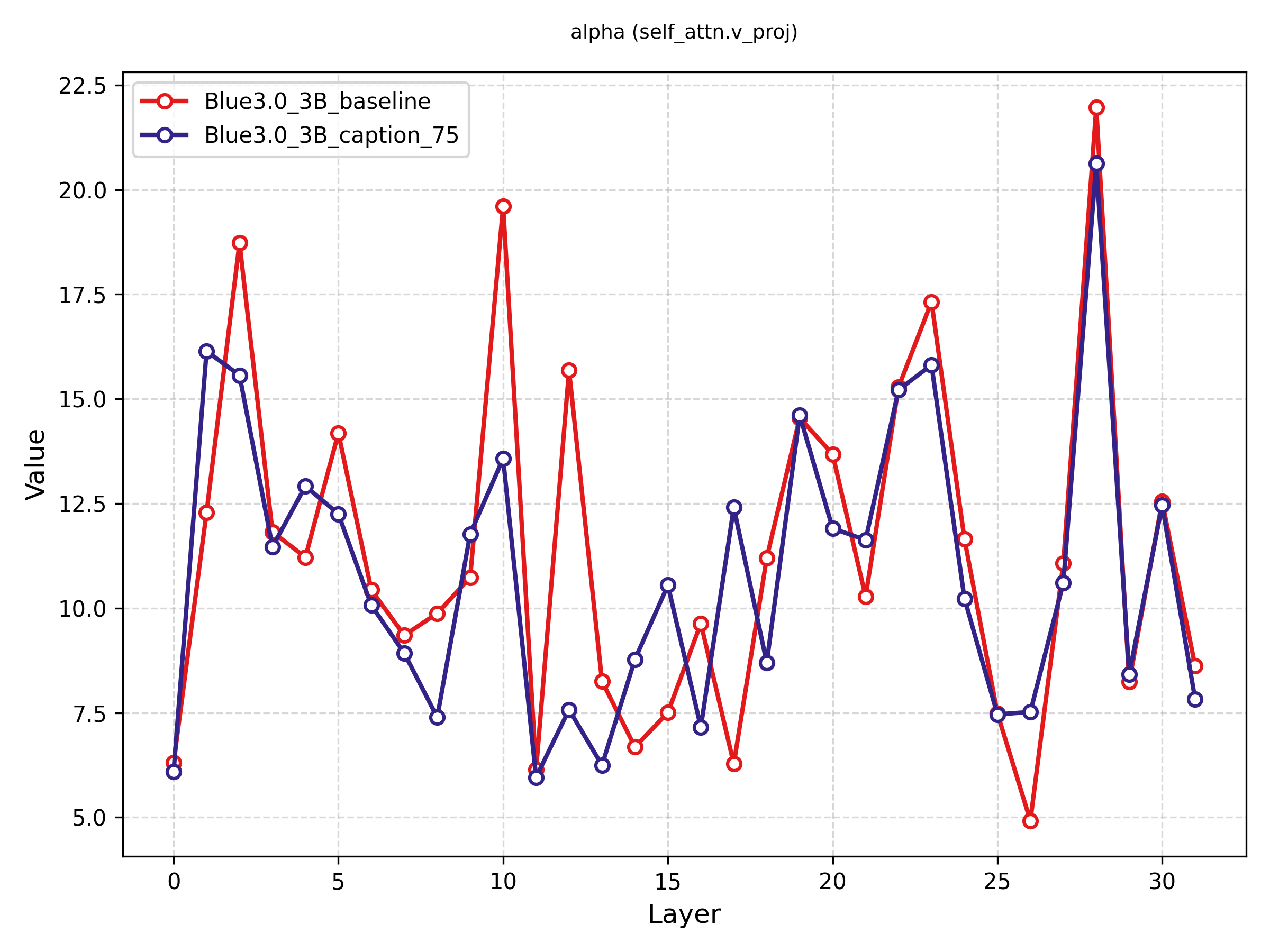}
\caption{Layer-wise $\alpha$ values for \texttt{self\_attn.v\_proj} across 32 layers, comparing the duplicated baseline (red) and the de-duplicated variant (blue) at checkpoint 95{,}000. The two profiles are highly correlated overall, but localized divergences appear at layers 2, 10, and 27, where the baseline exhibits elevated $\alpha$ values (roughly 18--22). These selective deviations indicate that prompt duplication introduces layer-specific distortions in attention projections while leaving the global spectral structure largely intact.}
\label{fig:dedup_spec_cmp_attn_alpha}
\end{figure}

\begin{figure}[t]
\centering
\small
\includegraphics[width=0.75\linewidth]{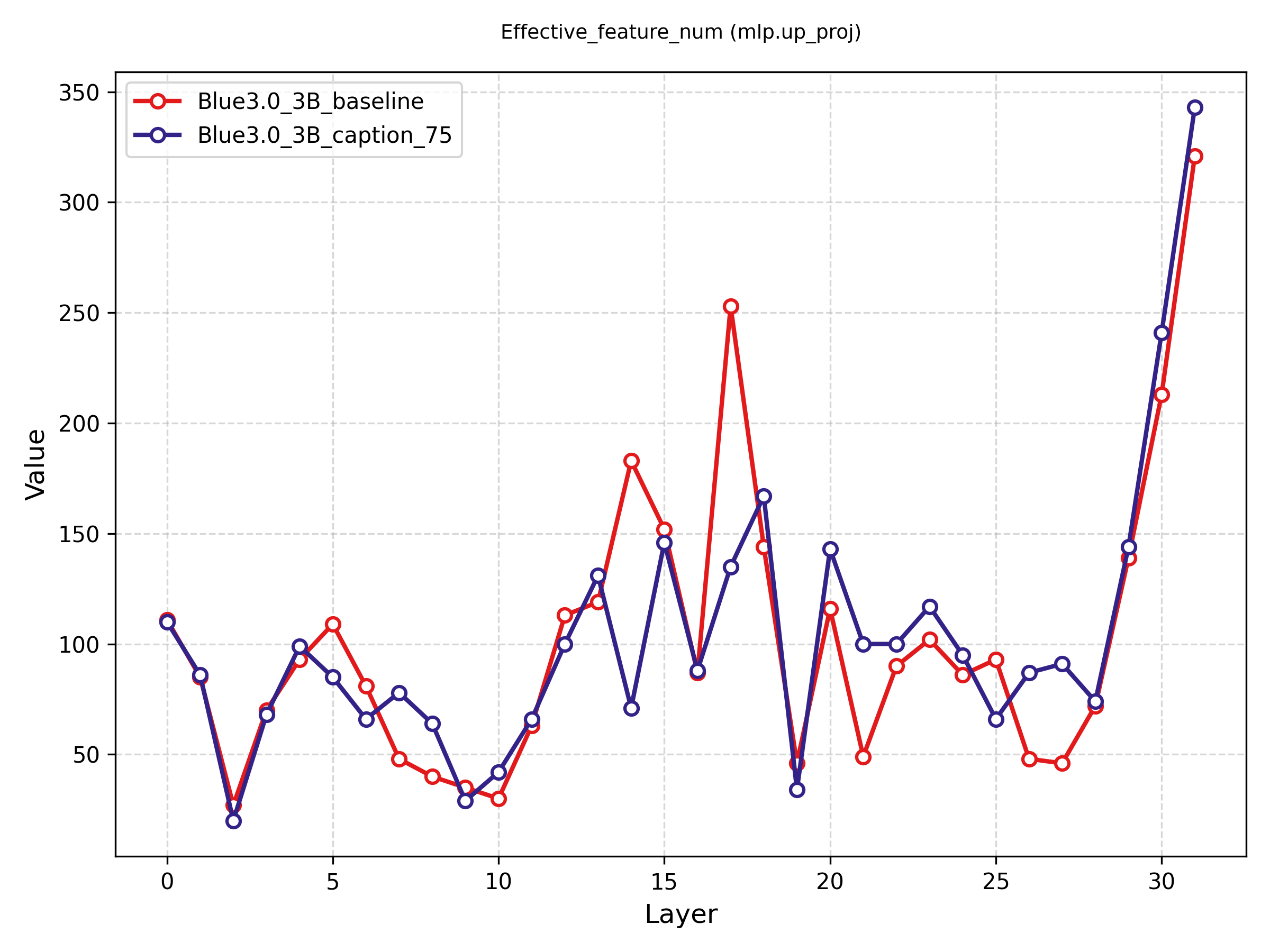}
\caption{Layer-wise effective feature number for \texttt{mlp.up\_proj} across 32 layers, comparing the duplicated baseline (red) and the de-duplicated variant (blue) at checkpoint 95{,}000. The two conditions track closely throughout, with both sharing the same characteristic rising trend toward late layers. This near-invariance contrasts with the selective divergences observed in \texttt{v\_proj}, suggesting that feed-forward layers are relatively insensitive to prompt duplication while attention projections absorb most of the structural impact.}
\label{fig:dedup_spec_cmp_mlp_eff_feat}
\end{figure}

\paragraph{Model-level impact.}
At the model level, prompt duplication does not produce the broad spectral degradation observed under the concentrated regimes in Section~\ref{controlled_data}. Figure~\ref{fig:dedup_mdl_lvl_base_vs_dedup} shows that the duplicated baseline and de-duplicated variant retain highly similar global $\alpha$ distributions. Although the de-duplicated condition has a slightly lower aggregate proportion of layers with $\alpha \in [2,6]$, the overall histogram changes only mildly, indicating that prompt repetition does not fundamentally alter the global conditioning profile.

\paragraph{Layer-wise effects.}
The layer-wise results show a more informative pattern. In \texttt{self\_attn.v\_proj}, the two conditions are highly correlated overall but differ at a small number of layers, where the duplicated baseline exhibits elevated $\alpha$ values. By contrast, \texttt{mlp.up\_proj} remains nearly unchanged across conditions. This separation suggests that prompt duplication primarily introduces localized distortions in attention-mediated instruction processing rather than a broad reorganization of representational structure.

\paragraph{Interpretation.}
This behavior differs qualitatively from the benchmark-shadow conditions studied earlier. In Section~\ref{controlled_data}, concentrated regimes altered both global spectral behavior and recovery dynamics because they narrowed effective support over semantic or structural pattern space. Prompt duplication, by contrast, mainly increases local redundancy while leaving the underlying semantic distribution largely intact. Accordingly, its effects are selective rather than regime-forming: it perturbs specific attention layers without producing the broader path-dependent signatures associated with support collapse or concentrated reinforcement. Consistent with this interpretation, delta effective rank---the diagnostic most sensitive to data-regime differences in the controlled experiments---remains nearly unchanged between the two conditions (see Appendix), further confirming that prompt duplication does not induce regime-level restructuring.

Notably, de-duplication does not improve the aggregate proportion of well-conditioned layers and slightly lowers the global $[2,6]$ ratio, yet it still improves downstream benchmark performance. This apparent mismatch is informative rather than contradictory. It suggests that prompt repetition primarily creates localized structural inefficiencies that are not well captured by coarse model-level summaries. More broadly, it highlights an important limitation of aggregate spectral metrics: global conditioning can remain nearly unchanged, or even move slightly in the opposite direction, while task-relevant attention pathways become cleaner after de-duplication.

\subsection{Benchmark Effects}

\begin{table}[t]
\centering
\caption{Benchmark performance comparison between the duplicated baseline and the de-duplicated variant across five evaluation categories. De-duplication improves four out of five categories, with the largest gains on business tasks requiring flexible interpretation. General-LLM benchmarks are largely unaffected, consistent with the intervention targeting multimodal prompt structure rather than broad language-domain knowledge.}
\label{tab:dedup_benchmark_summary}
\small
\begin{tabular}{lccc}
\toprule
\textbf{Category} & \textbf{Baseline} & \textbf{De-duplicated} & \textbf{$\Delta$} \\
\midrule
General-MLLM       & 0.5694 & 0.5892 & +0.0198 \\
General-LLM        & 0.3842 & 0.3708 & $-$0.0134 \\
Business-MLLM      & 0.3498 & 0.3764 & +0.0266 \\
Business-LLM       & 0.4799 & 0.5883 & +0.1084 \\
Instruct-Following & 0.5416 & 0.5623 & +0.0208 \\
\bottomrule
\end{tabular}
\end{table}

To assess whether the structural changes induced by de-duplication translate into measurable capability differences, we evaluate model performance across multiple benchmark categories. As shown in Table~\ref{tab:dedup_benchmark_summary}, de-duplication consistently improves performance on multimodal and instruction-following tasks, with the largest gains observed in business-oriented benchmarks that require flexible interpretation. In contrast, general LLM benchmarks remain largely unaffected, indicating that the intervention primarily affects multimodal instruction structure rather than broad language knowledge.

These results align with the structural analysis above. Prompt repetition does not induce the kind of regime shift observed under benchmark-shadow conditions, but it does introduce localized distortions in attention layers that affect how instruction signals are processed. Removing such redundancy therefore improves performance on several multimodal and instruction-following benchmarks without requiring additional data or model capacity.

\subsection{Summary}

This case study clarifies an important boundary of the regime-centric view. Prompt duplication introduces structural redundancy and can degrade local parameter quality, but it does not substantially narrow semantic support or alter the overall learning regime. Accordingly, de-duplication improves several multimodal and instruction-following benchmarks without producing the regime-level dynamics observed under benchmark-shadow conditions. The critical factor in regime formation is therefore not redundancy alone, but whether the data construction process collapses effective coverage over semantic and task space.

\section{Discussion}
\label{discussion}

\subsection{Implications for Data-Centric Training}

Our results support a regime-centric view of training: data distribution shapes not only what models score well on, but how representational capacity is allocated during learning. Training effectiveness therefore depends not only on data scale, but also on the regime induced by data composition.

First, optimizing for benchmark performance alone can implicitly encourage distributional concentration. When training data closely resembles evaluation settings, optimization may favor shortcut adaptation rather than broad representation learning, leading to improvements on narrow metrics without corresponding gains in general capability. Increasing coverage across domains, tasks, and formats is therefore essential for promoting robust generalization.

Second, regime effects are path-dependent. Our phase-shift experiments show that early exposure to benchmark-aligned or support-concentrated data leaves persistent structural footprints that are only partially corrected by later training on diverse data. This suggests that early-stage data composition is disproportionately important, and that late-stage correction is relatively inefficient. Designing regime-aware training curricula may therefore be more effective than post hoc data balancing.

The prompt-duplication case study further clarifies that regime formation is not caused by redundancy alone. Structural redundancy can degrade local processing efficiency, but the more consequential failure mode is concentration that narrows semantic support and constrains representational development. In other words, the key factor in harmful regime formation is not repetition per se, but collapse of effective coverage over tasks, formats, and knowledge.

\subsection{Monitoring Learning Dynamics During Training}

Parameter-space diagnostics provide complementary signals beyond conventional evaluation metrics.

Layer-wise update profiles, variance distributions, and representational rank changes reveal how learning signals are allocated across the network. Unlike benchmark scores, these structural signals are difficult to optimize directly and therefore provide a more reliable indication of underlying learning behavior.

In particular, we observe that benchmark-aligned regimes are often associated with limited representational restructuring despite substantial parameter updates in magnitude, a pattern we refer to as \textbf{spectral inertness}. Parameters change, but the internal geometry of representations remains comparatively constrained across layers. Monitoring such signatures during training could enable earlier detection of undesirable regimes and help guide data mixture design before benchmark effects become visible.

More broadly, our results suggest a three-level diagnostic hierarchy. Aggregate spectral summaries provide a coarse view of overall parameter quality, layer-wise spectral metrics expose regime-specific distortions that global statistics can hide, and dynamic diagnostics such as delta effective rank help distinguish data-induced restructuring from optimizer-driven effects. Taken together, these signals provide a more informative picture of learning dynamics than benchmark evaluation alone.

These observations suggest a promising direction for training-time monitoring tools that combine performance evaluation with structural diagnostics.

\subsection{Evaluation Beyond Benchmarks}

The discrepancy between benchmark performance, coverage-sensitive validation, and structural diagnostics highlights an important limitation of current evaluation practice.

Standard benchmarks emphasize relatively narrow task distributions and may fail to capture broader capability. Our results show that benchmark gains can coexist with structurally constrained adaptation and spectral inertness, indicating that benchmark success alone is not a reliable proxy for broad capability growth. This issue is especially visible when training data induces concentrated regimes: performance may improve on benchmark-matched tasks while generalization degrades outside the dominant support.

More comprehensive evaluation protocols should therefore incorporate:
\begin{itemize}
\item coverage-sensitive validation across domains and knowledge types,
\item robustness and transfer tasks,
\item structural diagnostics of parameter behavior.
\end{itemize}

Such multi-dimensional evaluation provides a more faithful measure of model capability and helps distinguish broad capability growth from narrow benchmark-aligned improvement.

\subsection{Limitations}

This study has several limitations.

First, while the controlled experiments isolate the effect of data distribution, the analysis of third-party models remains correlational. Factors such as model scale, architecture, optimization strategy, and multimodal design may also contribute to the observed differences.

Second, our diagnostics rely on spectral and structural proxies rather than direct measurement of gradient distributions. Although consistent patterns are observed, further work is needed to establish tighter theoretical connections between these metrics and underlying learning dynamics.

Third, our controlled experiments focus on a single model scale and architecture. Extending the analysis to larger models and more diverse architectures would strengthen the generality of the findings.

Fourth, the prompt de-duplication case study examines a single de-duplication setting. The relationship between duplication rate, semantic support, and regime formation remains underexplored, and future work should test whether the observed effects are stable across different thresholds and intervention strengths.

\subsection{Future Directions}

Several directions remain for future work. An immediate next step is to develop quantitative tools for real-time regime identification during training, especially diagnostics that can detect support collapse before benchmark effects appear. It is also important to extend coverage-sensitive evaluation to multimodal and long-context settings, where support concentration may take different forms. On the training side, regime-aware curricula that explicitly balance coverage and task alignment may offer a more principled alternative to benchmark-driven data selection. Finally, a natural extension of this work is to study whether similar regime effects arise under reinforcement learning, reasoning-focused post-training, and other settings where optimization pressure may amplify narrow pattern families.

\section{Conclusion}

We have shown that benchmark-oriented data regimes induce distinct and measurable parameter-space signatures. Controlled experiments demonstrate that concentration by support collapse produces more persistent structural distortion than repetition alone, while the prompt-duplication case study shows that redundancy without semantic narrowing does not induce the same regime-level effects. External validation further shows that analogous structural signatures recur across independently trained model families and align with asymmetric capability profiles across task types. Together, these findings suggest that benchmark performance is an incomplete proxy for model capability, and that parameter-space diagnostics provide complementary signals for assessing training quality and generalization.

\bibliographystyle{icml2025}
\bibliography{main}

\newpage
\appendix
\onecolumn

\section{Appendix}
\label{appendix}

\subsection{Rewriting Prompt for Condition D (Frequency-Concentrated Regime)}
\label{app:condition_d_prompt}

The following prompt is used in the teacher-based rewriting procedure for Condition~D (frequency-concentrated regime). The teacher model (Qwen3-30B-A3B-Instruct-2507) receives each training sample with this instruction, which classifies the content and applies format-specific simplification. English and Chinese variants are used depending on detected language.

\paragraph{English variant.}
\begin{mdframed}
\small
\ttfamily
{
Please rewrite the following content according to these instructions:\\
1. First, classify the content as [Text], [Exercise], or [Code].\\
2. If it is [Text], try to use very simple words and sentences to express the same meaning.\\
3. If it is [Code], rewrite it in JavaScript format.\\
4. If it is [Exercise], output it exactly as it is without any changes.\\
5. Keep formatting symbols in the original text, such as `\textbackslash n', and do not change formatting based on them.\\
6. Do not add extra content or modify the original data.\\
7. Only output the rewritten content; do not mention the classification result.\\
8. Format the output as: <output> your output </output>.\\
Content to rewrite: \{text\}
}
\end{mdframed}

\FloatBarrier

\subsection{Prompt De-duplication: Benchmark Scores}

\FloatBarrier

\subsubsection{General MLLM Benchmarks}

\begin{table}[t]
\centering
\caption{General-MLLM benchmark scores for the duplicated baseline and de-duplicated variant. Scores are averaged over multiple runs where applicable.}
\label{tab:dedup_general_mllm}
\scriptsize
\small
\begin{tabular}{lccc}
\toprule
\textbf{Benchmark} & \textbf{Baseline} & \textbf{De-duplicated} & \textbf{$\Delta$} \\
\midrule
MIA-Bench (acc)       & 0.7558 & 0.7801 & +0.0243 \\
MMMU (pass@4)         & 0.4756 & 0.5016 & +0.0260 \\
MMBench-dev (acc)     & 0.7147 & 0.7240 & +0.0093 \\
MMVet                 & 0.4965 & 0.5412 & +0.0448 \\
MMStar                & 0.5160 & 0.5451 & +0.0291 \\
MathVision (avg@4)    & 0.1264 & 0.1381 & +0.0117 \\
MathVista (pass@4)    & 0.6143 & 0.6213 & +0.0070 \\
AI2D-test             & 0.6803 & 0.7004 & +0.0201 \\
OCRBench              & 0.7453 & 0.7507 & +0.0053 \\
\midrule
\textbf{Average}      & 0.5694 & 0.5892 & +0.0198 \\
\bottomrule
\end{tabular}
\end{table}

\FloatBarrier

\subsubsection{General LLM Benchmarks}

\begin{table}[t]
\centering
\caption{General-LLM benchmark scores for the duplicated baseline and de-duplicated variant.}
\label{tab:dedup_general_llm}
\scriptsize
\small
\begin{tabular}{lccc}
\toprule
\textbf{Benchmark} & \textbf{Baseline} & \textbf{De-duplicated} & \textbf{$\Delta$} \\
\midrule
MMLU-Pro (micro acc)    & 0.1763 & 0.1850 & +0.0087 \\
GPQA-Diamond (pass@4)   & 0.6330 & 0.5808 & $-$0.0522 \\
CFBench (psr)            & 0.3433 & 0.3467 & +0.0033 \\
\midrule
\textbf{Average}         & 0.3842 & 0.3708 & $-$0.0134 \\
\bottomrule
\end{tabular}
\end{table}

\FloatBarrier

\subsubsection{Business MLLM Benchmarks}

\begin{table}[t]
\centering
\caption{Business-MLLM benchmark scores for the duplicated baseline and de-duplicated variant.}
\label{tab:dedup_business_mllm}
\scriptsize
\small
\begin{tabular}{lccc}
\toprule
\textbf{Benchmark} & \textbf{Baseline} & \textbf{De-duplicated} & \textbf{$\Delta$} \\
\midrule
OCR-GRD (avg score)             & 0.7185 & 0.8214 & +0.1029 \\
OCR-REC (score)                 & 0.6899 & 0.7161 & +0.0262 \\
Screen-ViVO-Pro (avg score)     & 0.0724 & 0.0589 & $-$0.0135 \\
Screen-ViVO-v2.1 (avg score)    & 0.1870 & 0.1717 & $-$0.0153 \\
Object Recognition (avg)        & 0.3589 & 0.3250 & $-$0.0339 \\
Info Extraction (acc)            & 0.2041 & 0.3722 & +0.1681 \\
Doc Understanding (attr)         & 0.2835 & 0.2994 & +0.0159 \\
Doc Understanding (table)        & 0.4826 & 0.4728 & $-$0.0099 \\
Image Description (overall)      & 0.3795 & 0.3830 & +0.0036 \\
Image Grounding (avg score)      & 0.1214 & 0.1435 & +0.0221 \\
\midrule
\textbf{Average}                 & 0.3498 & 0.3764 & +0.0266 \\
\bottomrule
\end{tabular}
\end{table}

\FloatBarrier

\subsubsection{Business LLM Benchmarks}

\begin{table}[t]
\centering
\caption{Business-LLM benchmark scores for the duplicated baseline and de-duplicated variant.}
\label{tab:dedup_business_llm}
\scriptsize
\small
\begin{tabular}{lccc}
\toprule
\textbf{Benchmark} & \textbf{Baseline} & \textbf{De-duplicated} & \textbf{$\Delta$} \\
\midrule
WritingBench (avg)      & 0.4813 & 0.5001 & +0.0187 \\
Summary (avg overall)   & 0.7057 & 0.7120 & +0.0063 \\
Scene NER (overall F1)  & 0.5539 & 0.5515 & $-$0.0024 \\
Keywords (F1)           & 0.1786 & 0.5896 & +0.4110 \\
\midrule
\textbf{Average}        & 0.4799 & 0.5883 & +0.1084 \\
\bottomrule
\end{tabular}
\end{table}

\FloatBarrier

\subsubsection{Instruction Following Benchmarks}

\begin{table}[t]
\centering
\caption{Instruction-following benchmark scores for the duplicated baseline and de-duplicated variant.}
\label{tab:dedup_instruct_following}
\scriptsize
\small
\begin{tabular}{lccc}
\toprule
\textbf{Benchmark} & \textbf{Baseline} & \textbf{De-duplicated} & \textbf{$\Delta$} \\
\midrule
IF-v2 (acc)                      & 0.2817 & 0.2867 & +0.0050 \\
CFBench (csr)                    & 0.6067 & 0.6133 & +0.0067 \\
CFBench (isr)                    & 0.2500 & 0.2567 & +0.0067 \\
CFBench (psr)                    & 0.3433 & 0.3467 & +0.0033 \\
IF-Creativity (loose acc)        & 0.5772 & 0.6296 & +0.0525 \\
IF-Info Extraction (loose acc)   & 0.7225 & 0.8197 & +0.0973 \\
IF-Summary (loose acc)           & 0.5666 & 0.6150 & +0.0485 \\
IFEval (inst-level loose)        & 0.7034 & 0.6870 & $-$0.0164 \\
IFEval (prompt-level loose)      & 0.6087 & 0.5884 & $-$0.0203 \\
MIA-Bench (acc)                  & 0.7558 & 0.7801 & +0.0243 \\
\midrule
\textbf{Average}                 & 0.5416 & 0.5623 & +0.0208 \\
\bottomrule
\end{tabular}
\end{table}

\FloatBarrier

\subsection{Controlled Data Intervention: Additional Figures}
\label{app:ctrl_data_figures}

\begin{figure*}[htbp!]
\centering
\small

\begin{subfigure}[b]{0.3\linewidth}
    \centering
    \includegraphics[width=\linewidth]{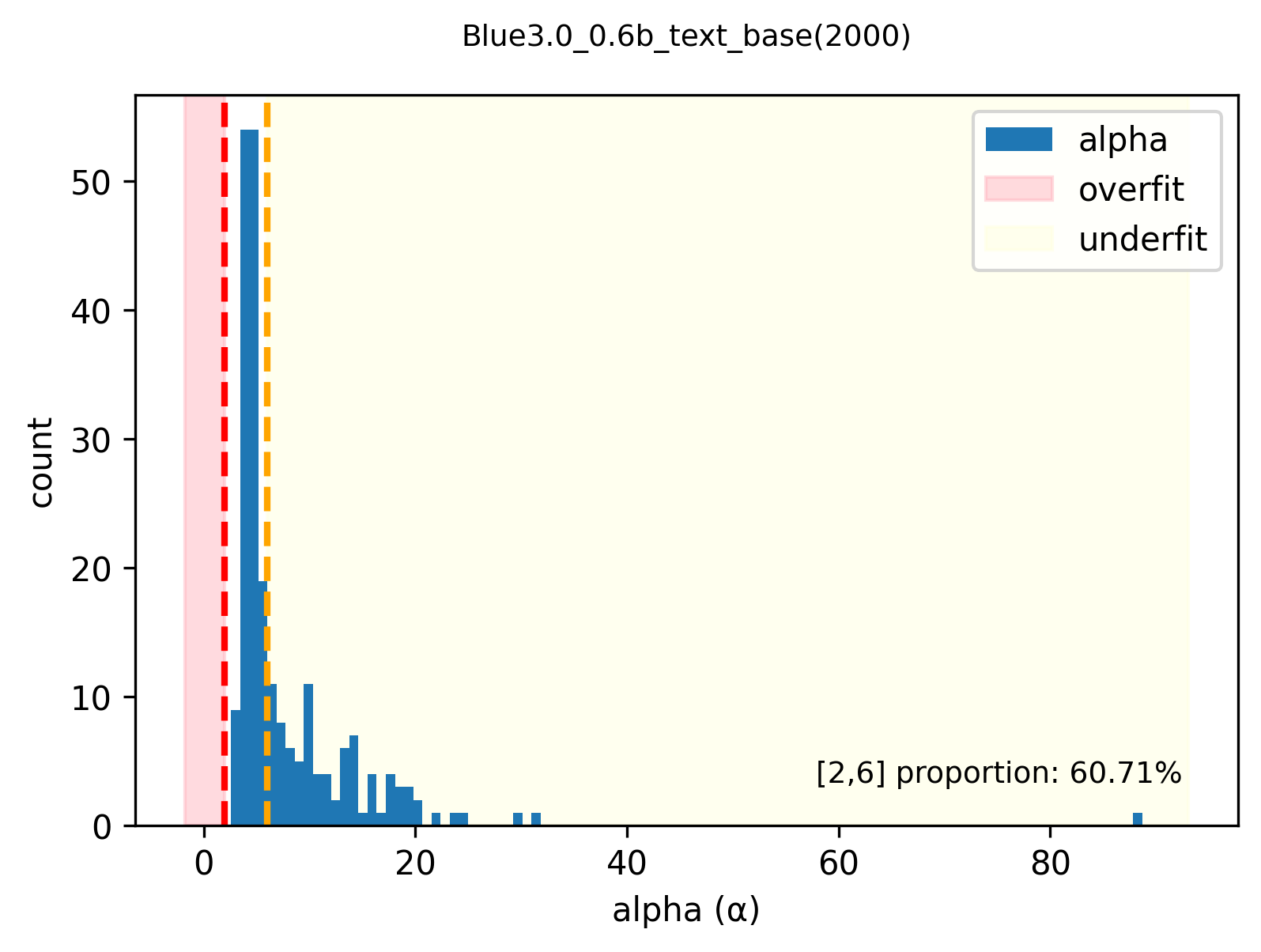}
    \caption{Condition A, 2k}
\end{subfigure}
\hfill
\begin{subfigure}[b]{0.3\linewidth}
    \centering
    \includegraphics[width=\linewidth]{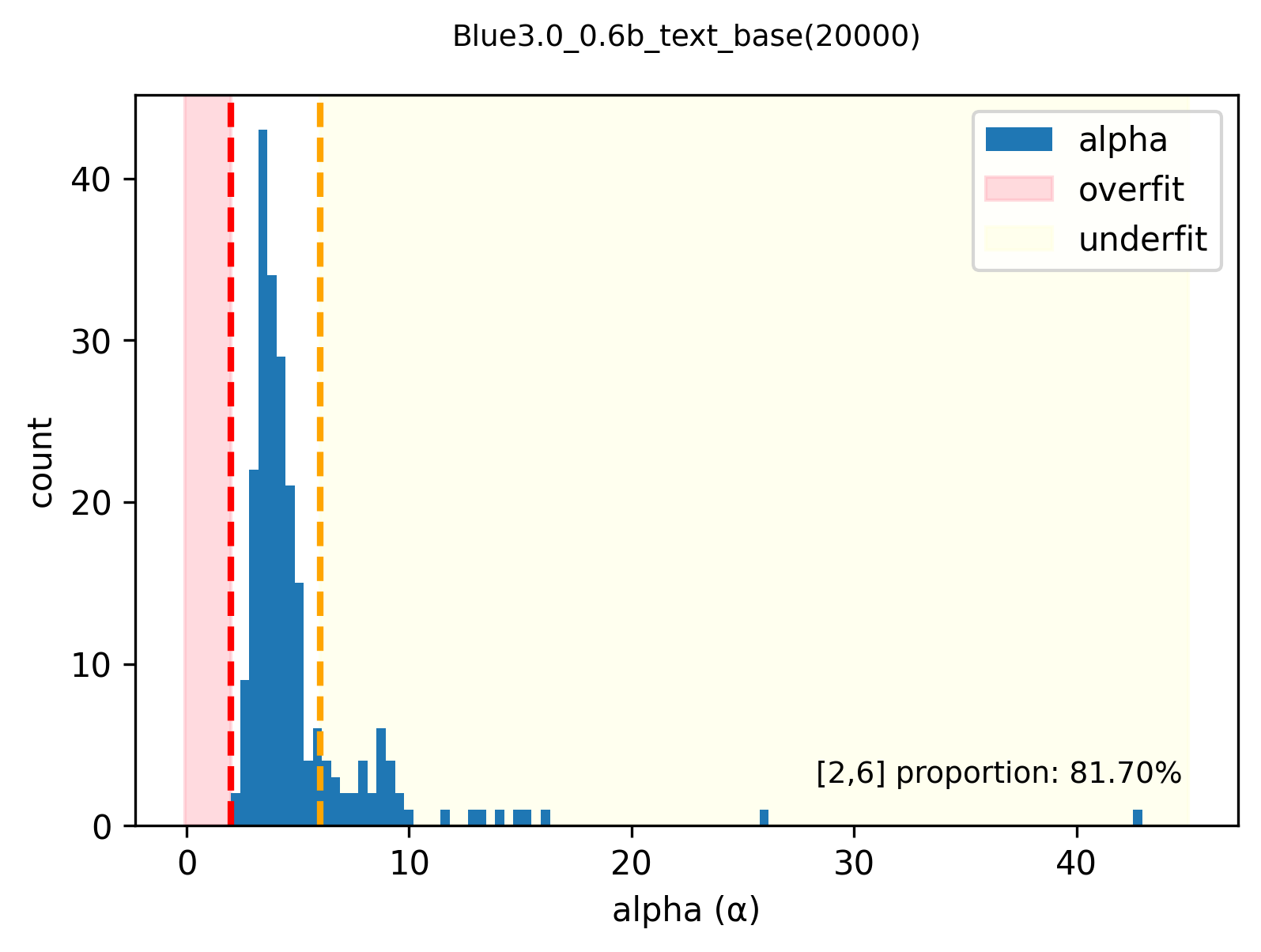}
    \caption{Condition A, 20k}
\end{subfigure}
\hfill
\begin{subfigure}[b]{0.3\linewidth}
    \centering
    \includegraphics[width=\linewidth]{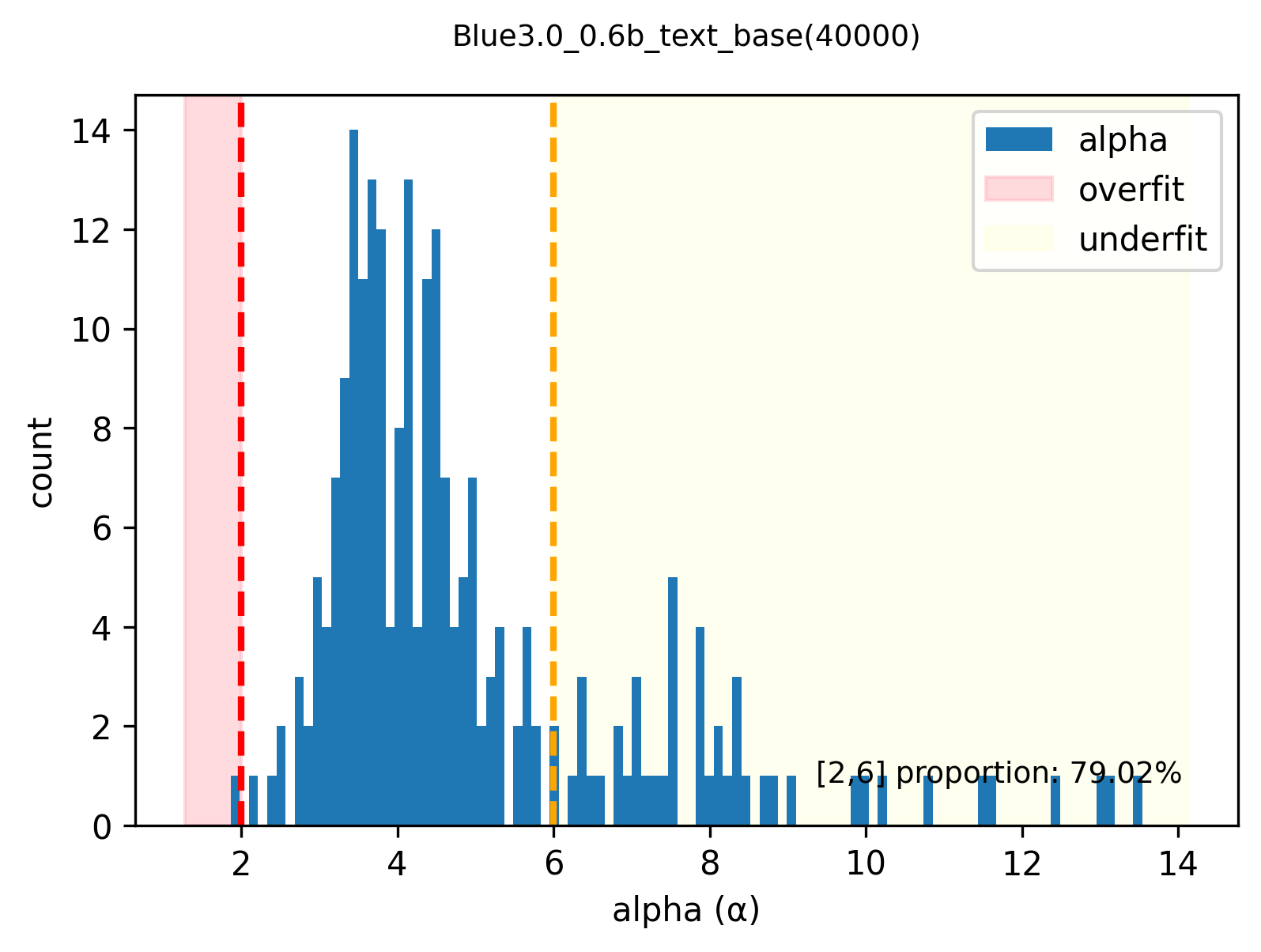}
    \caption{Condition A, 40k}
\end{subfigure}

\vspace{0.3cm}

\begin{subfigure}[b]{0.3\linewidth}
    \centering
    \includegraphics[width=\linewidth]{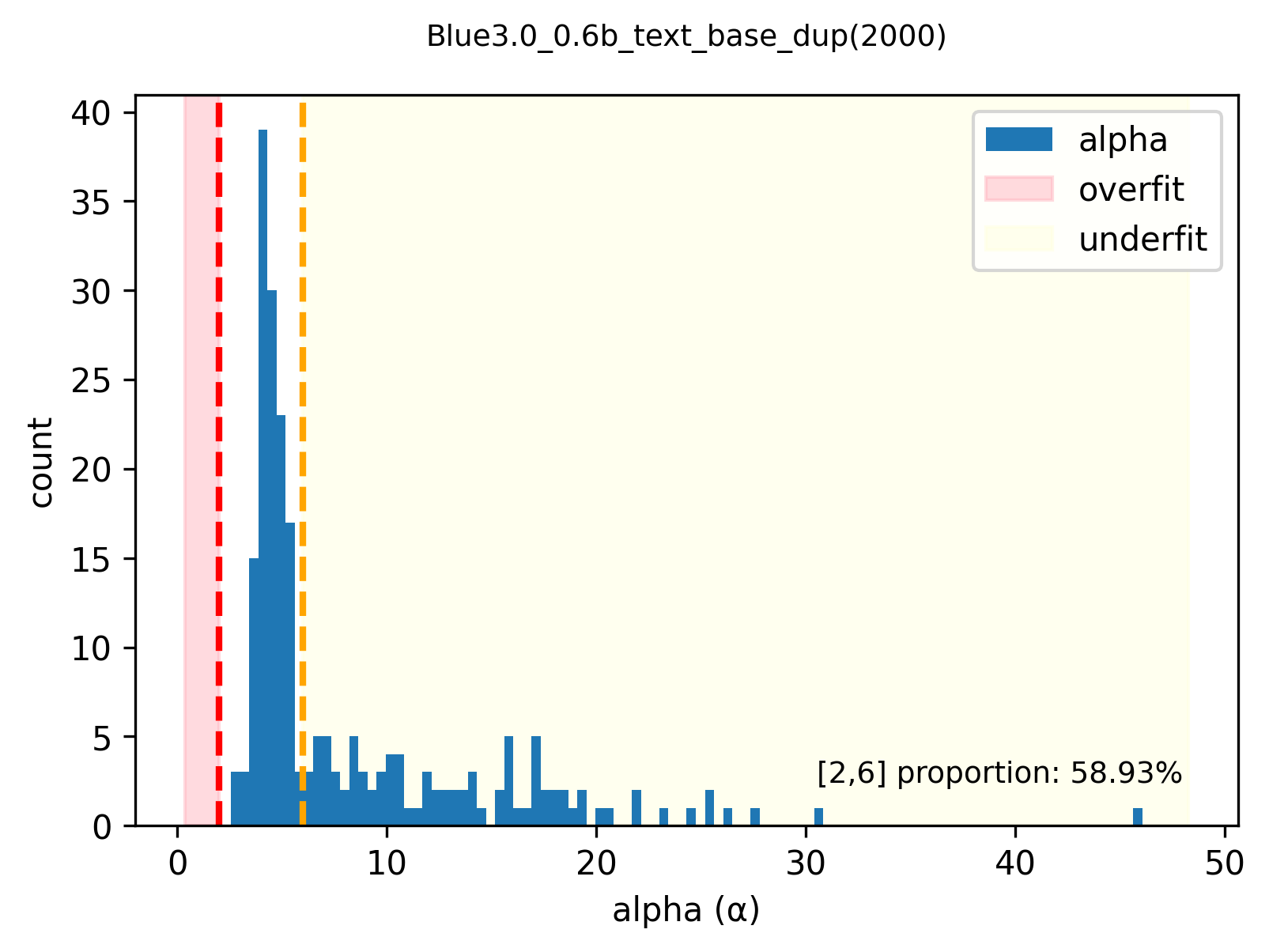}
    \caption{Condition C, 2k}
\end{subfigure}
\hfill
\begin{subfigure}[b]{0.3\linewidth}
    \centering
    \includegraphics[width=\linewidth]{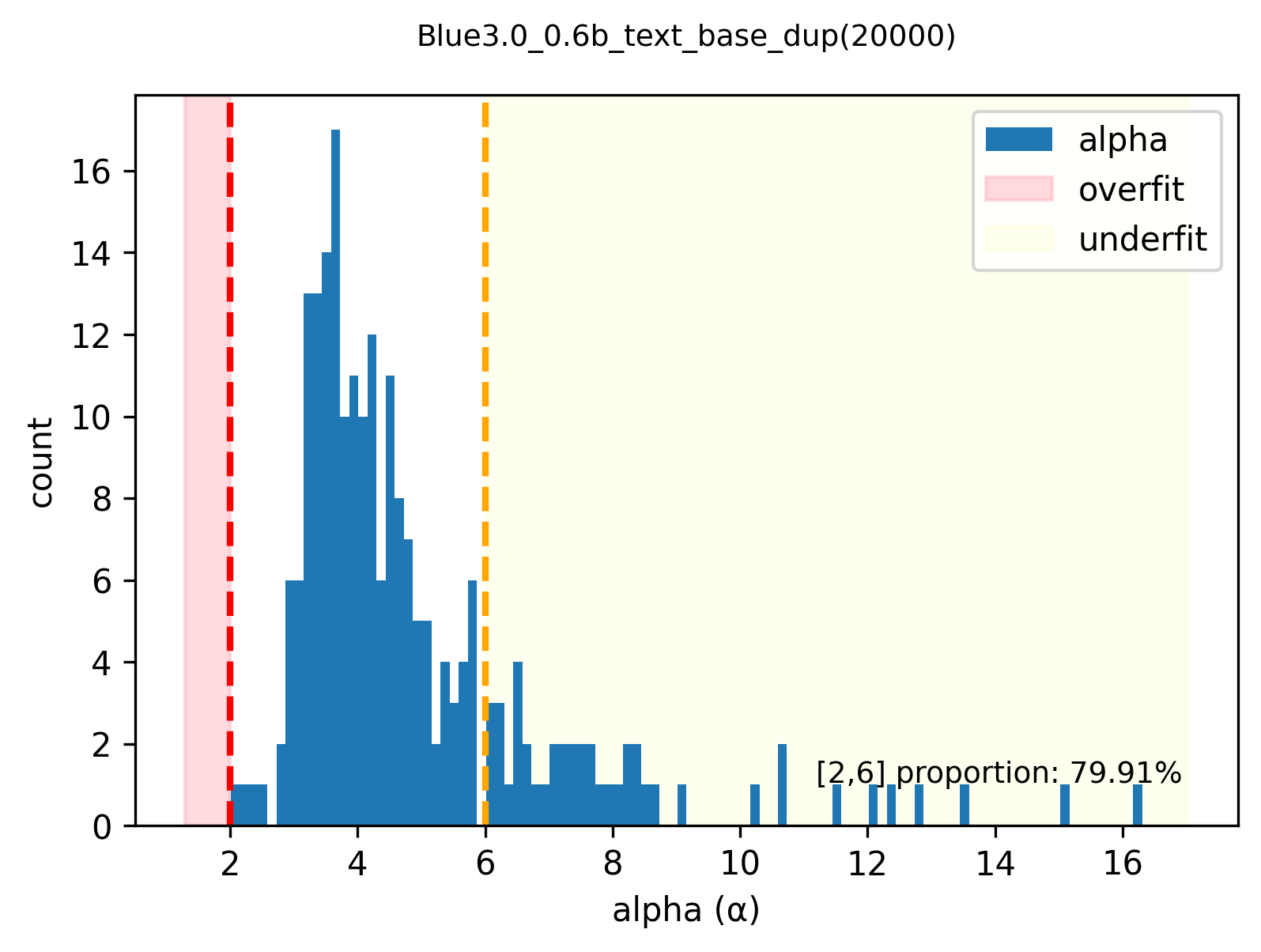}
    \caption{Condition C, 20k}
\end{subfigure}
\hfill
\begin{subfigure}[b]{0.3\linewidth}
    \centering
    \includegraphics[width=\linewidth]{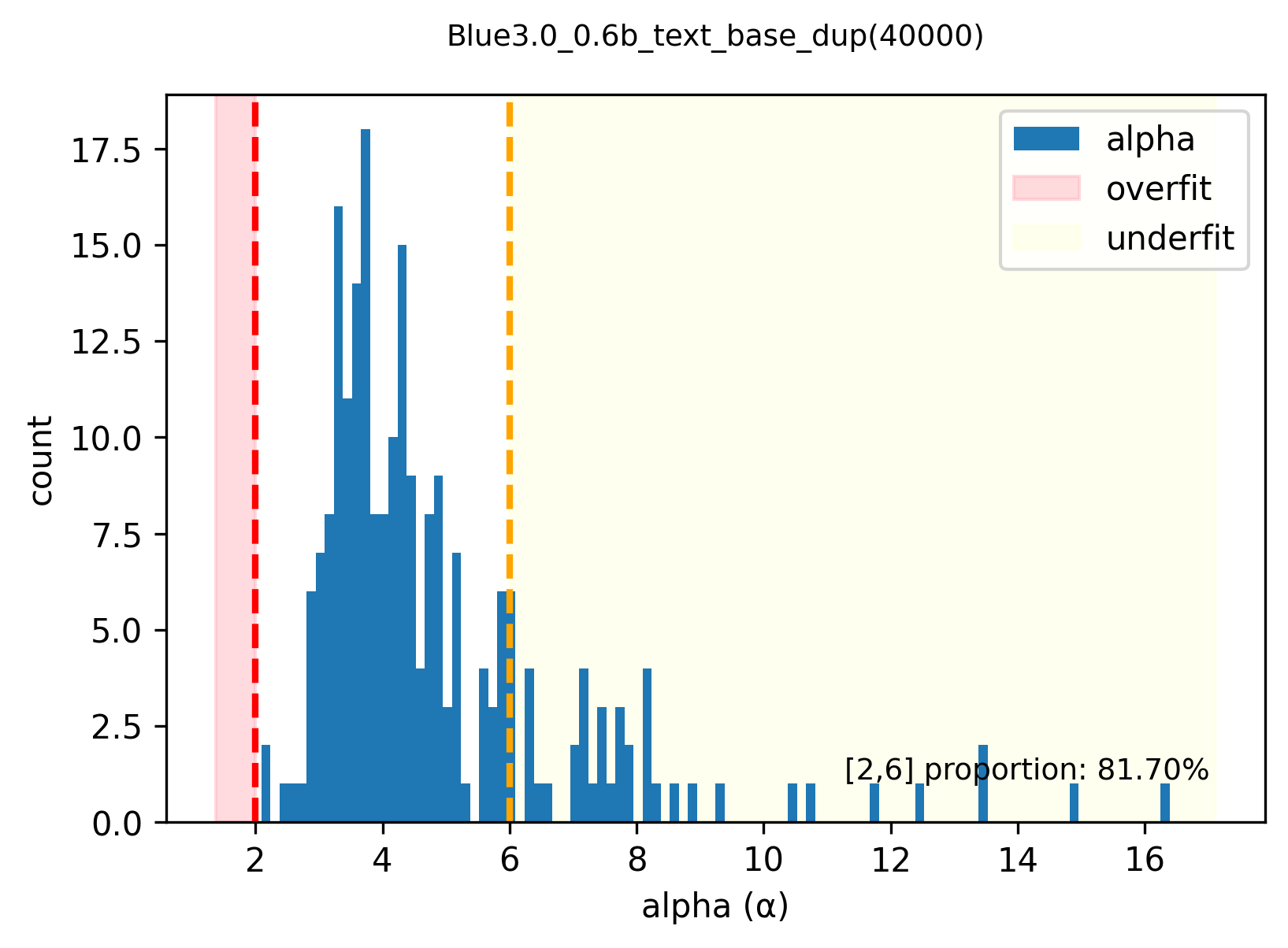}
    \caption{Condition C, 40k}
\end{subfigure}

\vspace{0.3cm}

\begin{subfigure}[b]{0.3\linewidth}
    \centering
    \includegraphics[width=\linewidth]{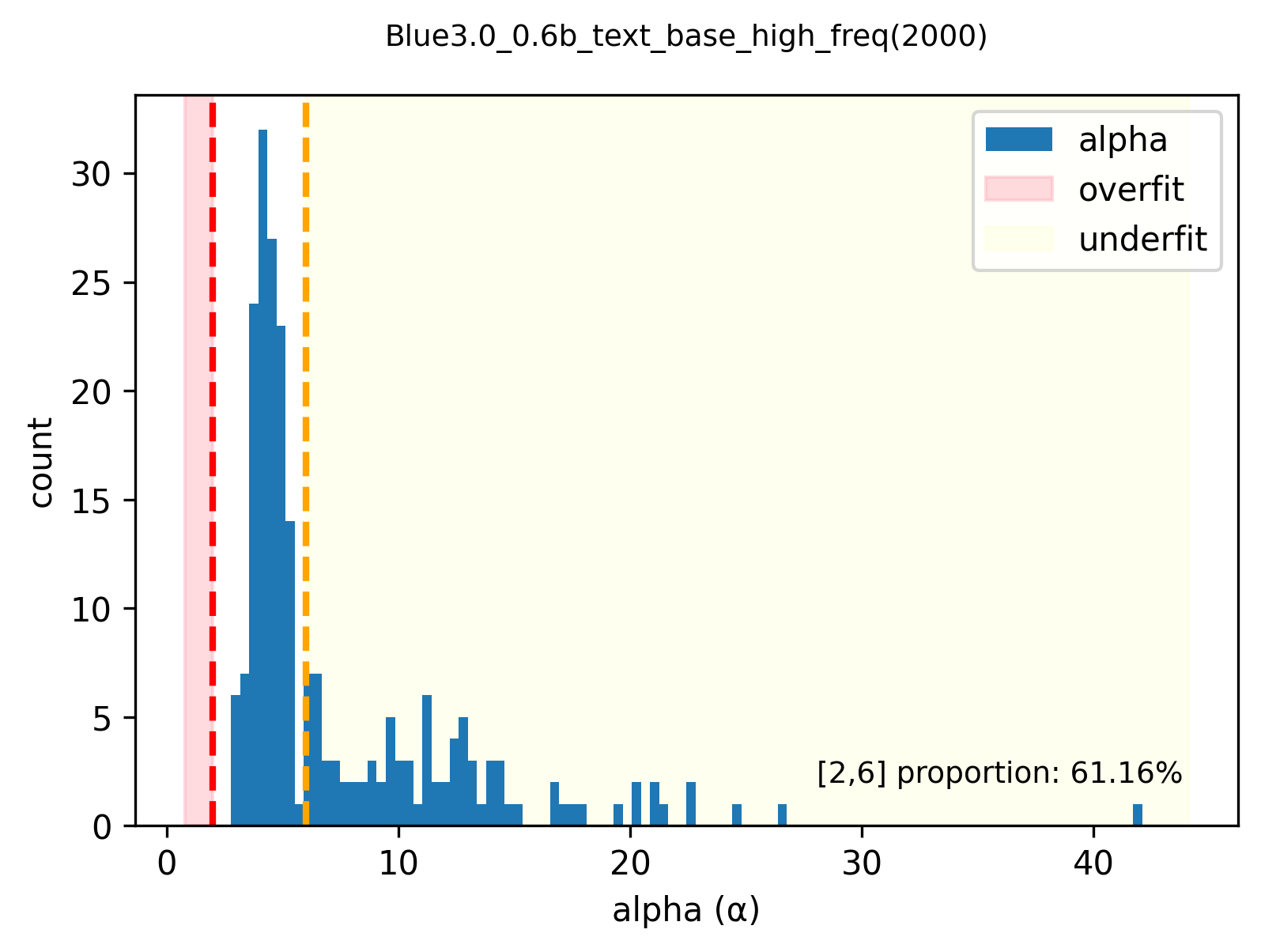}
    \caption{Condition D, 2k}
\end{subfigure}
\hfill
\begin{subfigure}[b]{0.3\linewidth}
    \centering
    \includegraphics[width=\linewidth]{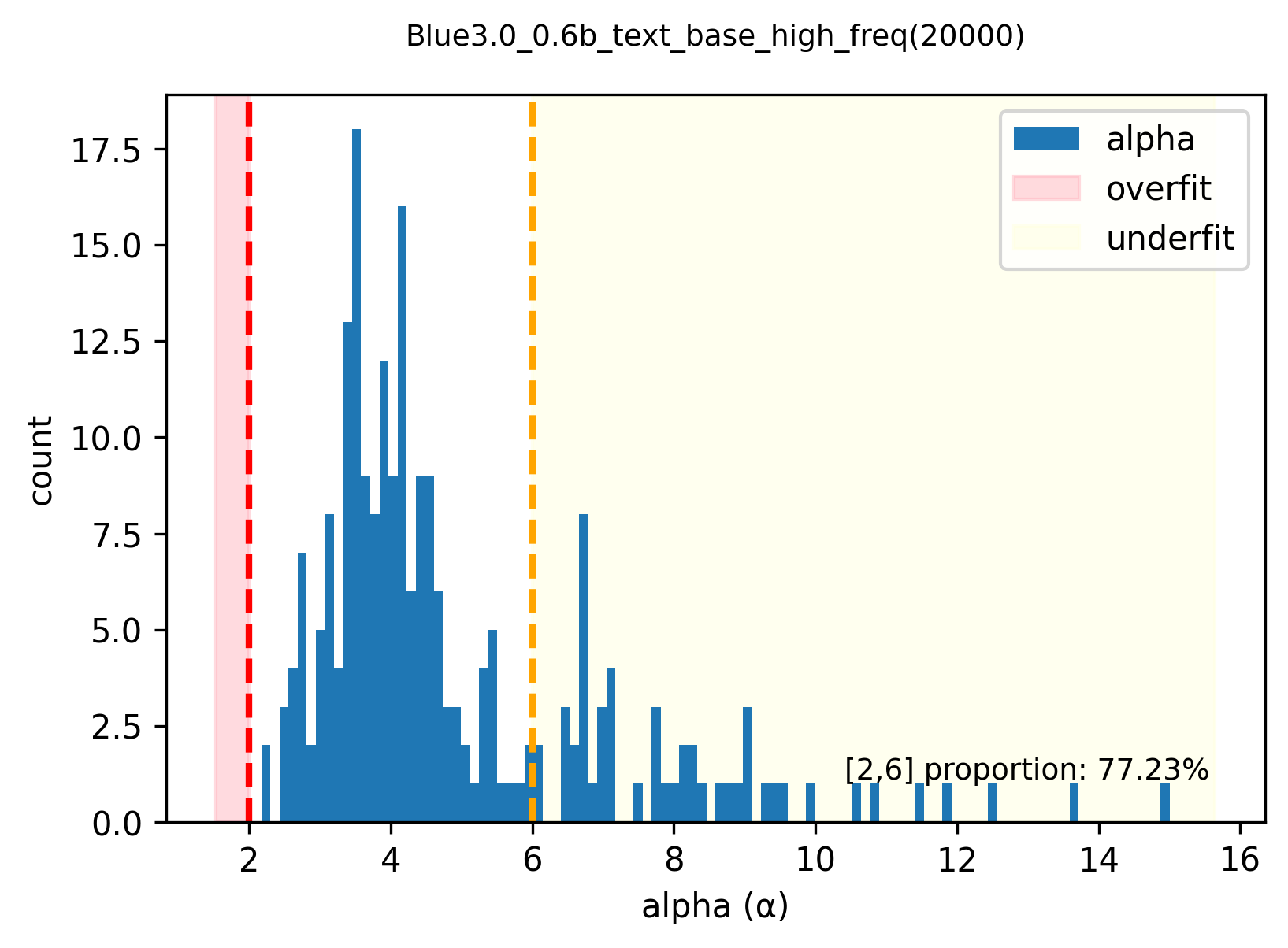}
    \caption{Condition D, 20k}
\end{subfigure}
\hfill
\begin{subfigure}[b]{0.3\linewidth}
    \centering
    \includegraphics[width=\linewidth]{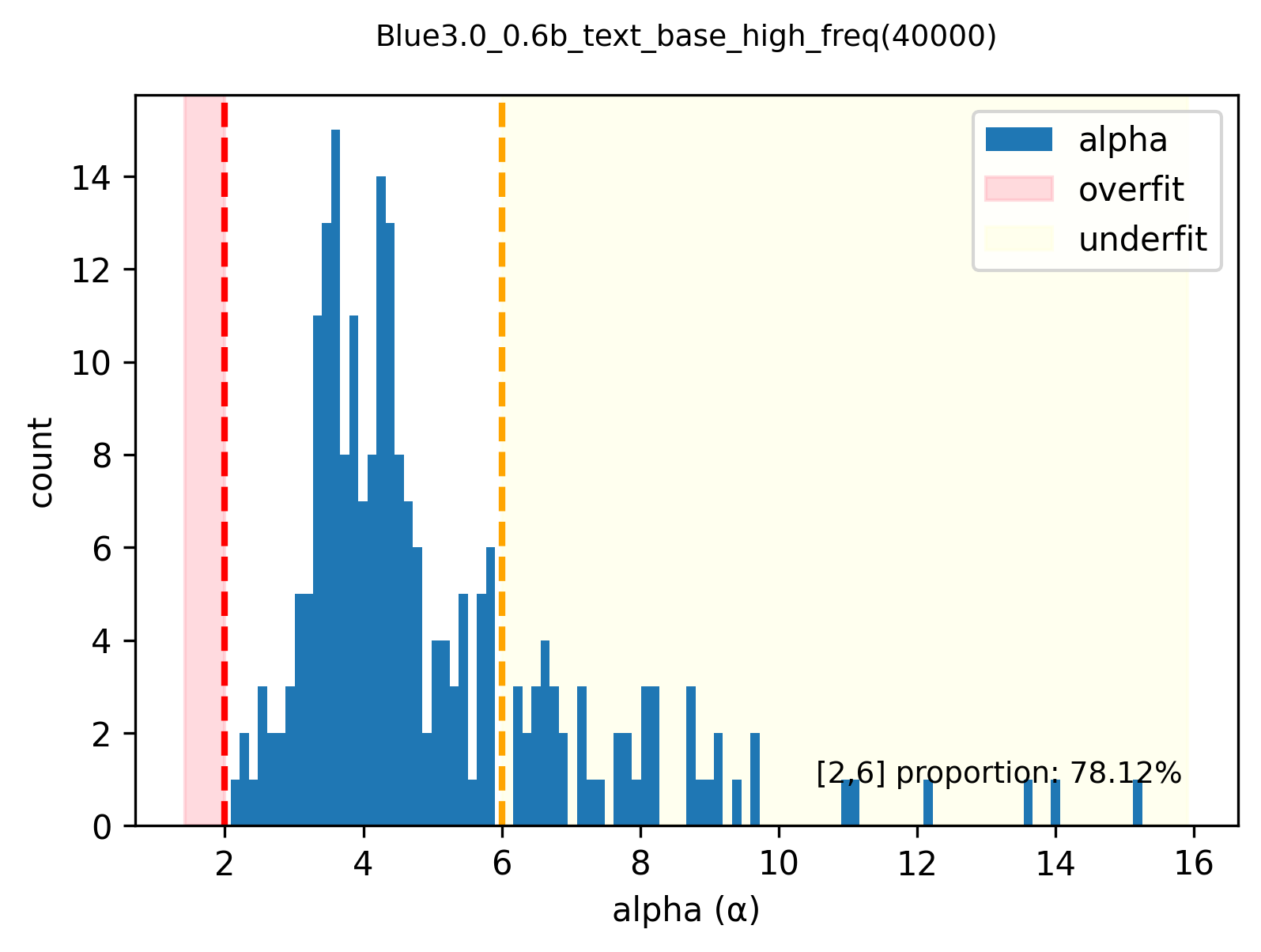}
    \caption{Condition D, 40k}
\end{subfigure}

\caption{Temporal evolution of model-level $\alpha$ distributions under the controlled data interventions. Rows correspond to the coverage-expanding baseline (Condition A), repetition-concentrated regime (Condition C), and frequency-concentrated regime (Condition D). Columns correspond to checkpoints at 2k, 20k, and 40k steps. These intermediate checkpoints make the recovery asymmetry discussed in Section~\ref{controlled_data} visually explicit: Condition C shows stronger later recovery, whereas Condition D remains more persistently offset.}
\label{fig:app_ctrl_data_temporal_mdl_lvl}
\end{figure*}

\begin{figure*}[htbp!]
\centering
\small

\begin{subfigure}[b]{0.32\linewidth}
    \centering
    \includegraphics[width=\linewidth]{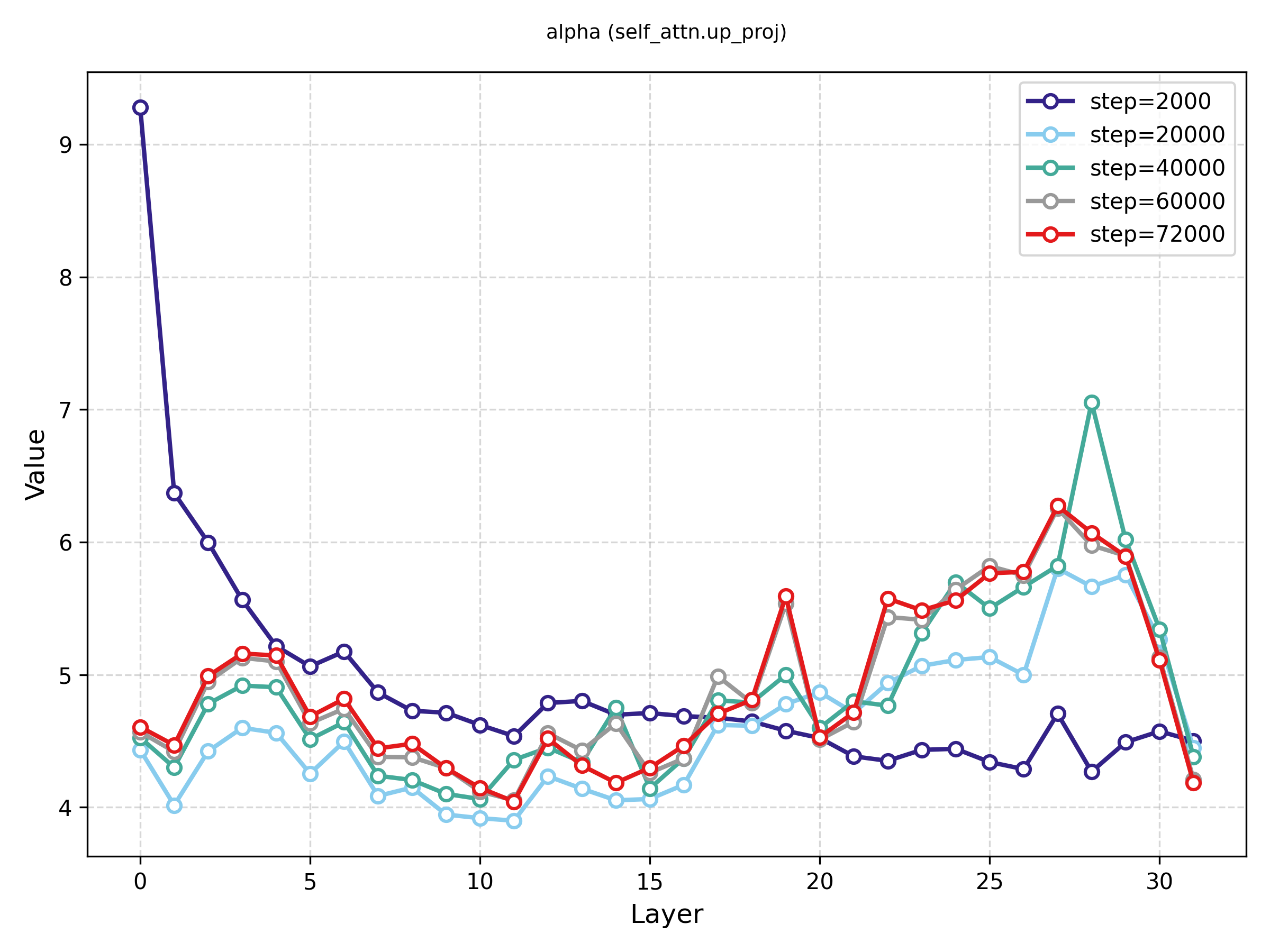}
    \caption{Condition A}
\end{subfigure}
\hfill
\begin{subfigure}[b]{0.32\linewidth}
    \centering
    \includegraphics[width=\linewidth]{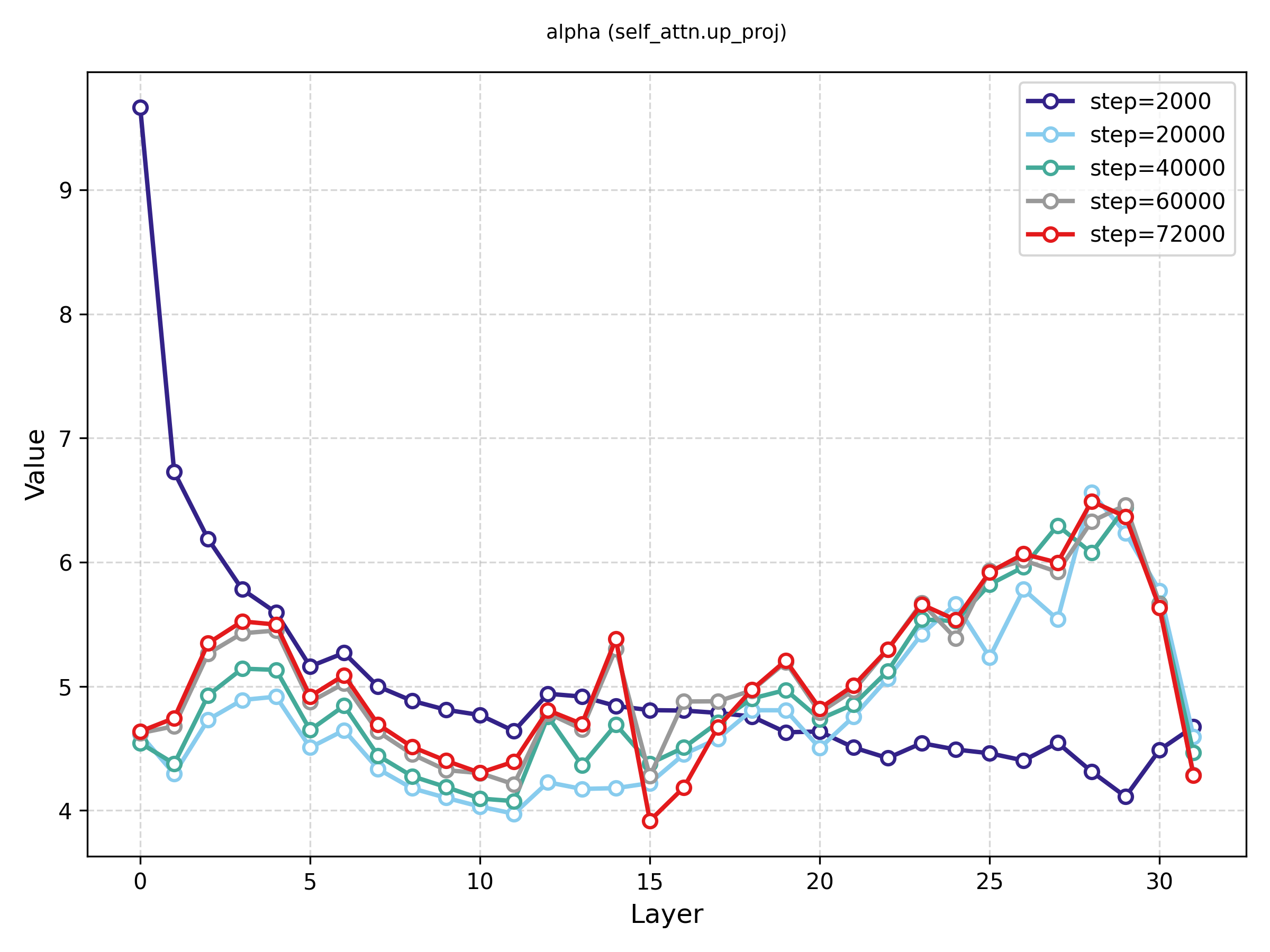}
    \caption{Condition C}
\end{subfigure}
\hfill
\begin{subfigure}[b]{0.32\linewidth}
    \centering
    \includegraphics[width=\linewidth]{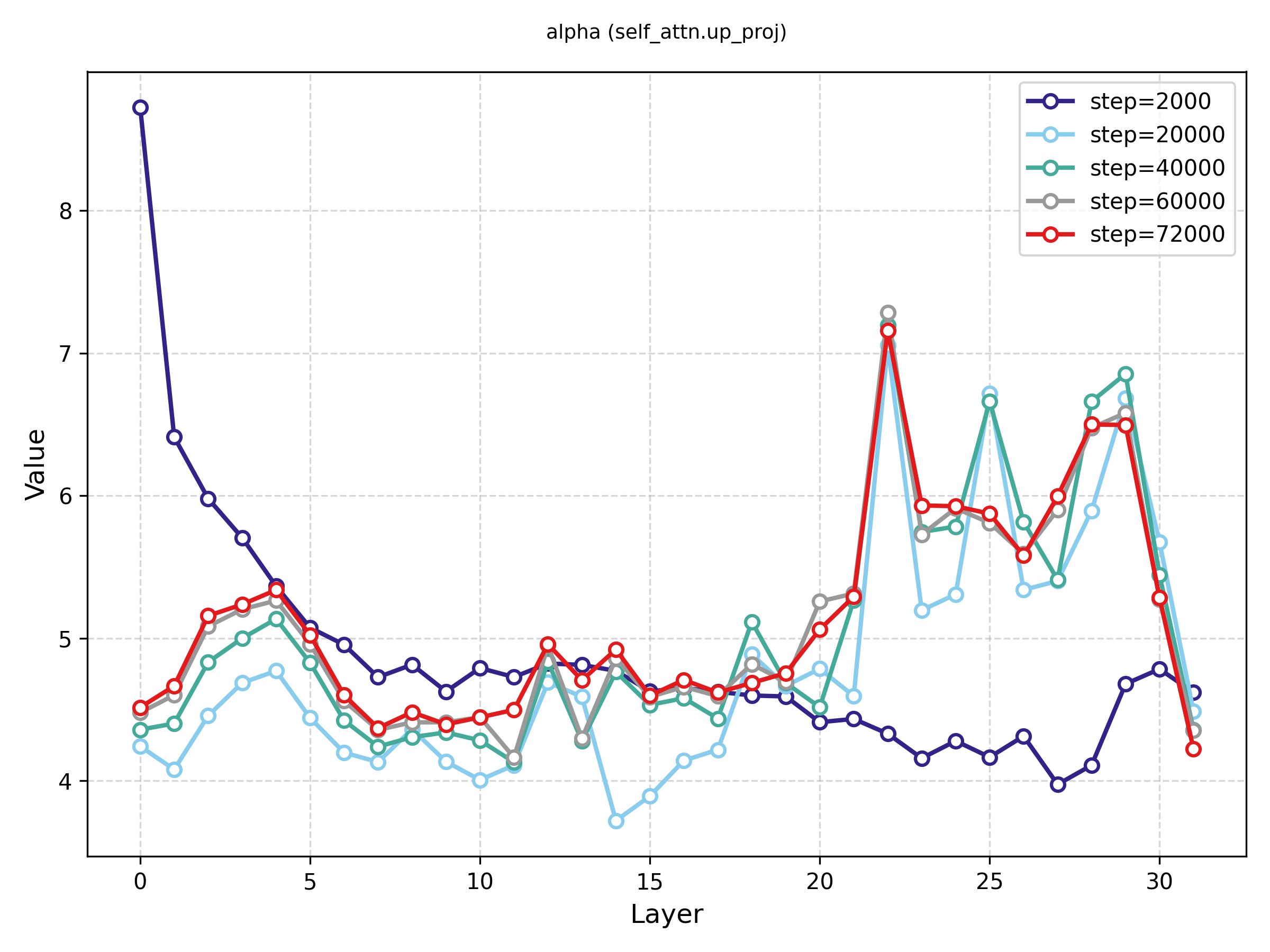}
    \caption{Condition D}
\end{subfigure}

\caption{Layer-wise $\alpha$ values in \texttt{mlp.up\_proj} at the final training stage for three representative data conditions. In contrast to the stronger condition-specific separation observed in attention projections, MLP $\alpha$ profiles remain comparatively similar across conditions, supporting the attention--MLP asymmetry discussed in Section~\ref{controlled_data}.}
\label{fig:app_ctrl_data_mlp_alpha}
\end{figure*}

\begin{figure}[htbp!]
\centering
\small
\includegraphics[width=0.45\linewidth]{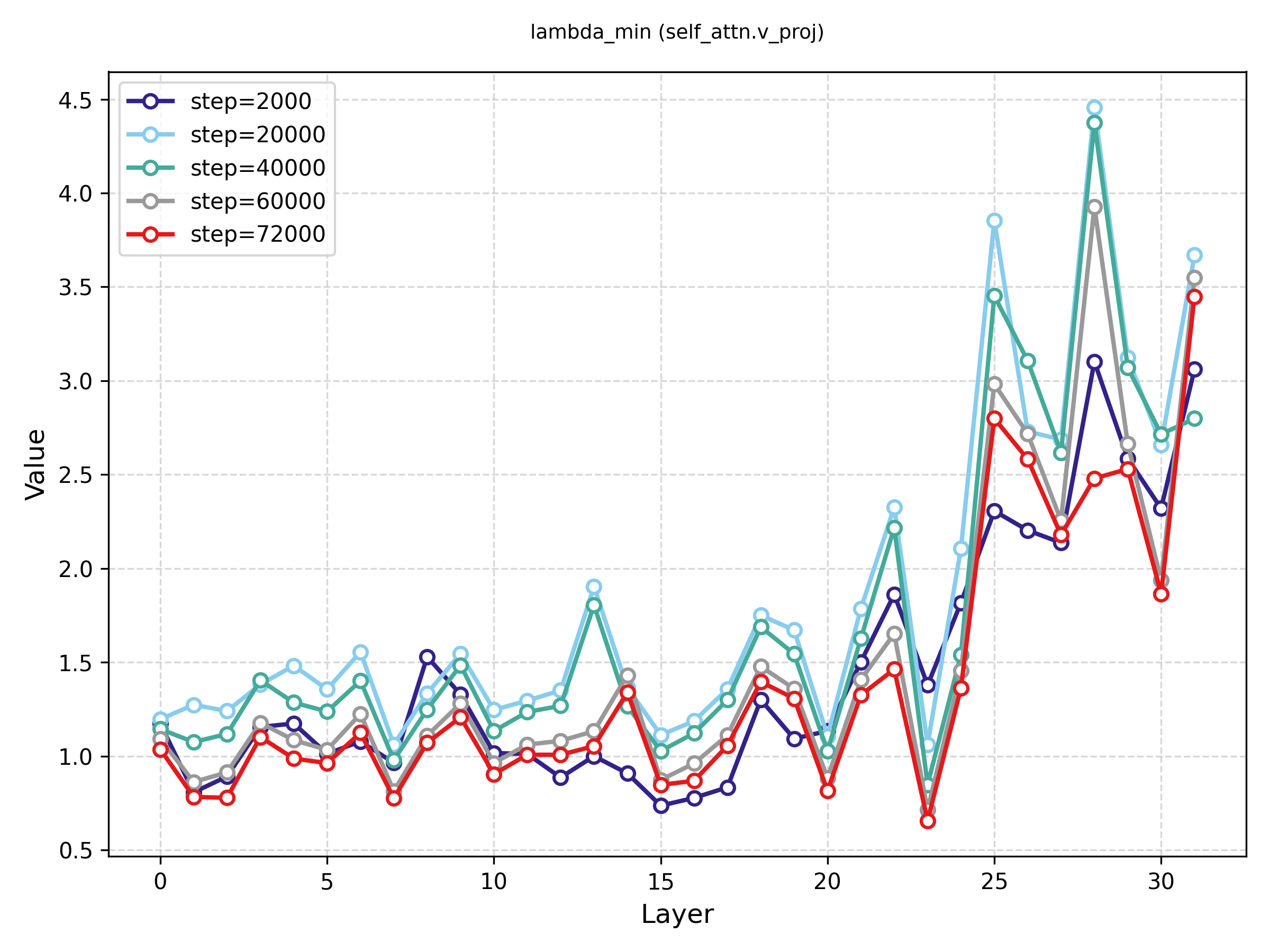}
\caption{Layer-wise $\lambda_{\min}$ in \texttt{self\_attn.v\_proj} for the optimization-control baseline (Condition B). The monotonic trend provides additional evidence that this condition reflects an optimization-driven effect rather than the irregular, data-induced layer-specific distortions observed under concentrated data regimes.}
\label{fig:app_ctrl_data_attn_lammin_b}
\end{figure}

\begin{figure*}[htbp!]
\centering
\small

\begin{subfigure}[b]{0.45\linewidth}
    \centering
    \includegraphics[width=\linewidth]{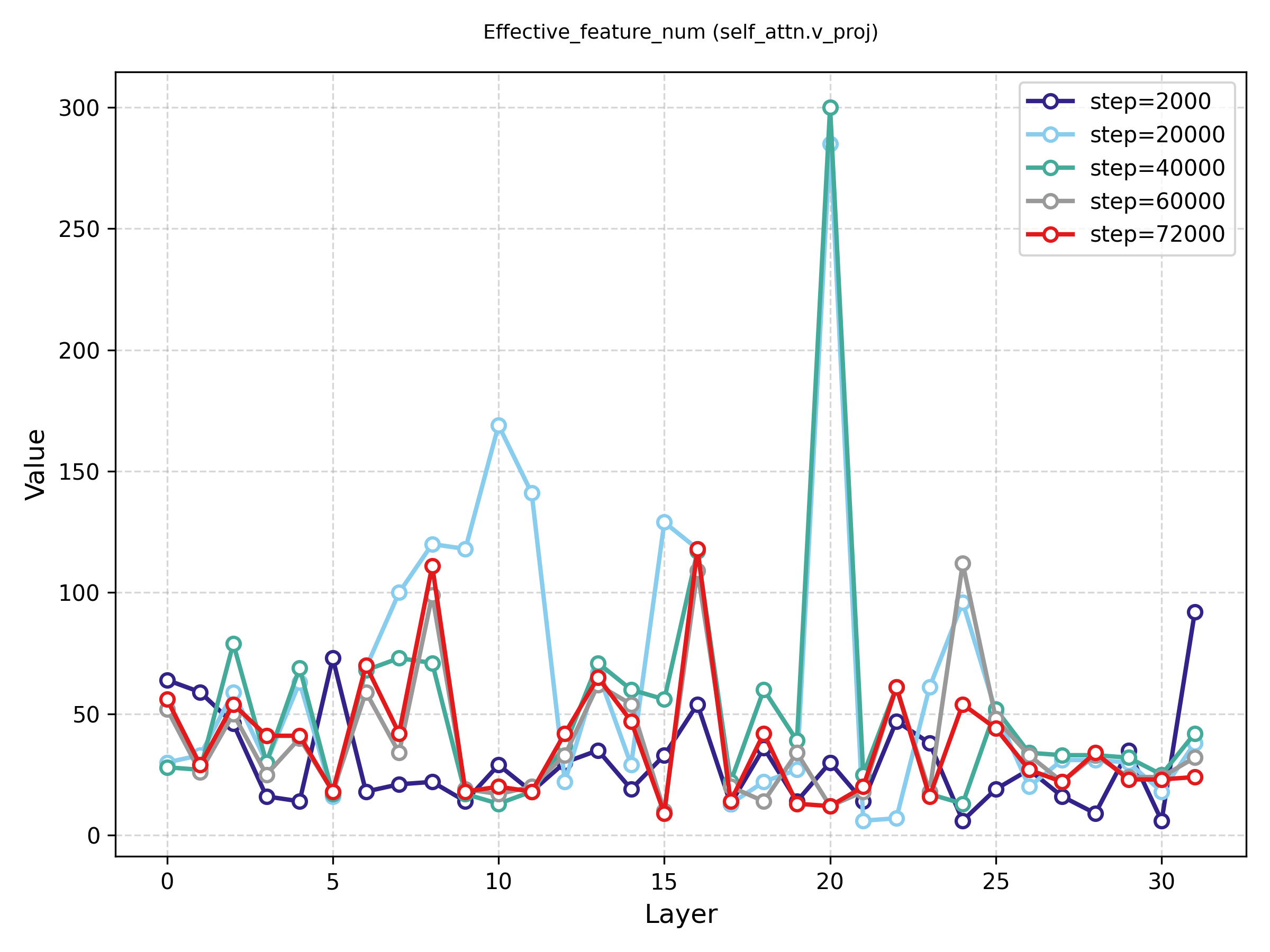}
    \caption{Condition A}
\end{subfigure}
\hfill
\begin{subfigure}[b]{0.45\linewidth}
    \centering
    \includegraphics[width=\linewidth]{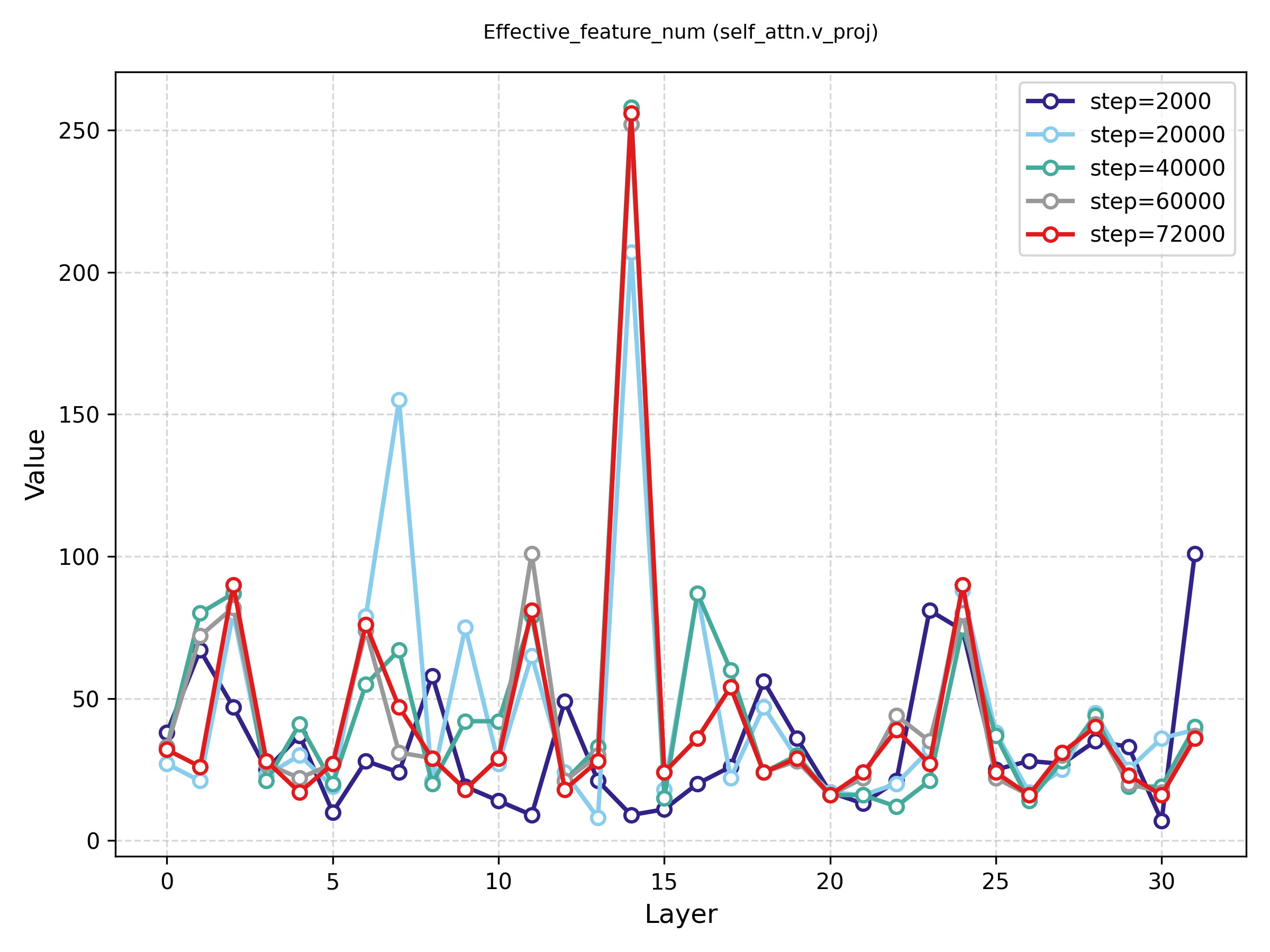}
    \caption{Condition D}
\end{subfigure}

\caption{Layer-wise effective feature number in \texttt{self\_attn.v\_proj} for the coverage-expanding baseline (Condition A) and the frequency-concentrated regime (Condition D). Condition D exhibits stronger inhomogeneity, especially in upper layers, consistent with uneven allocation of representational capacity under support collapse.}
\label{fig:app_ctrl_data_attn_eff_feat}
\end{figure*}

\begin{figure*}[htbp!]
\centering
\small

\begin{subfigure}[b]{0.45\linewidth}
    \centering
    \includegraphics[width=\linewidth]{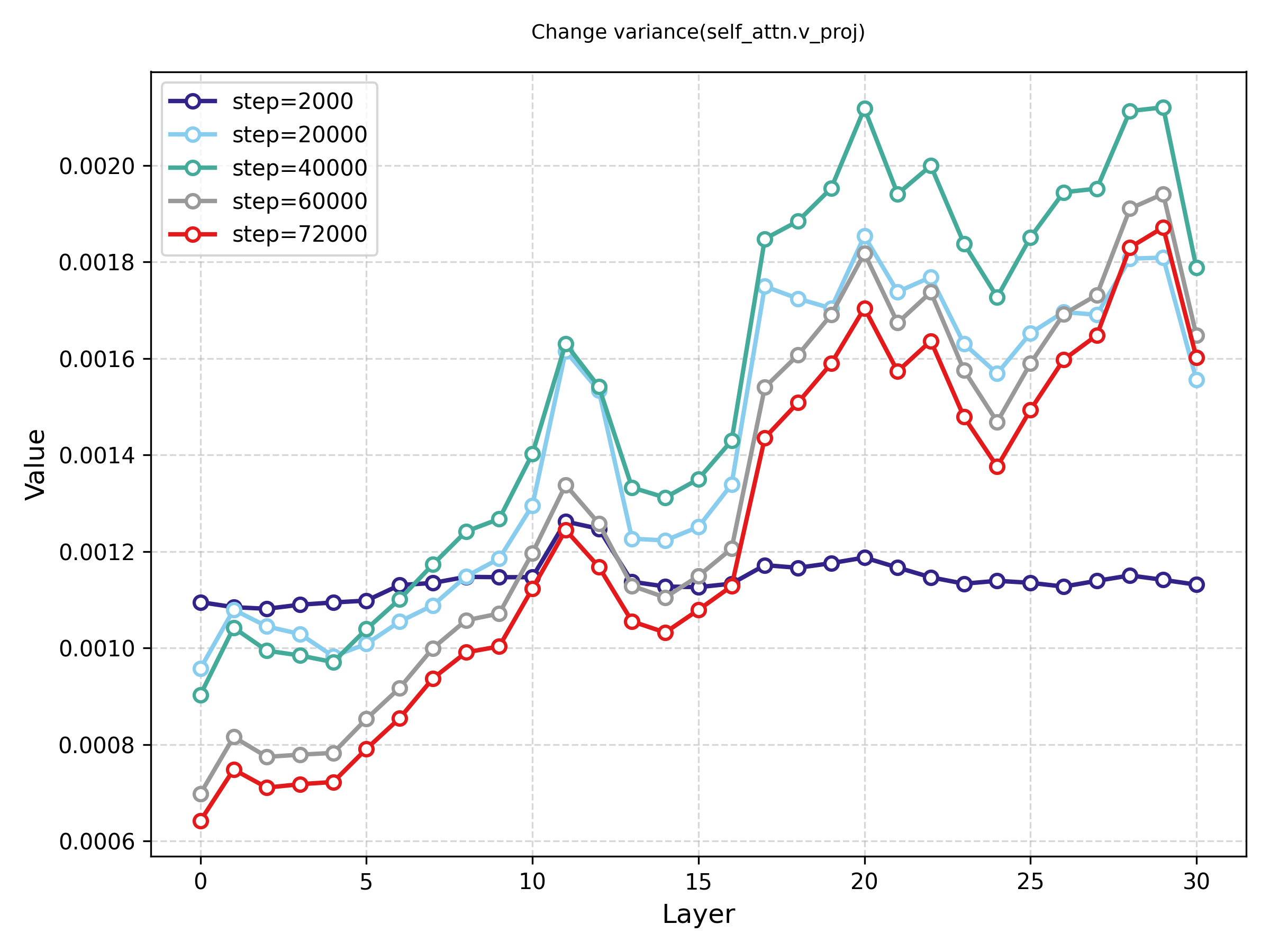}
    \caption{Condition C}
\end{subfigure}
\hfill
\begin{subfigure}[b]{0.45\linewidth}
    \centering
    \includegraphics[width=\linewidth]{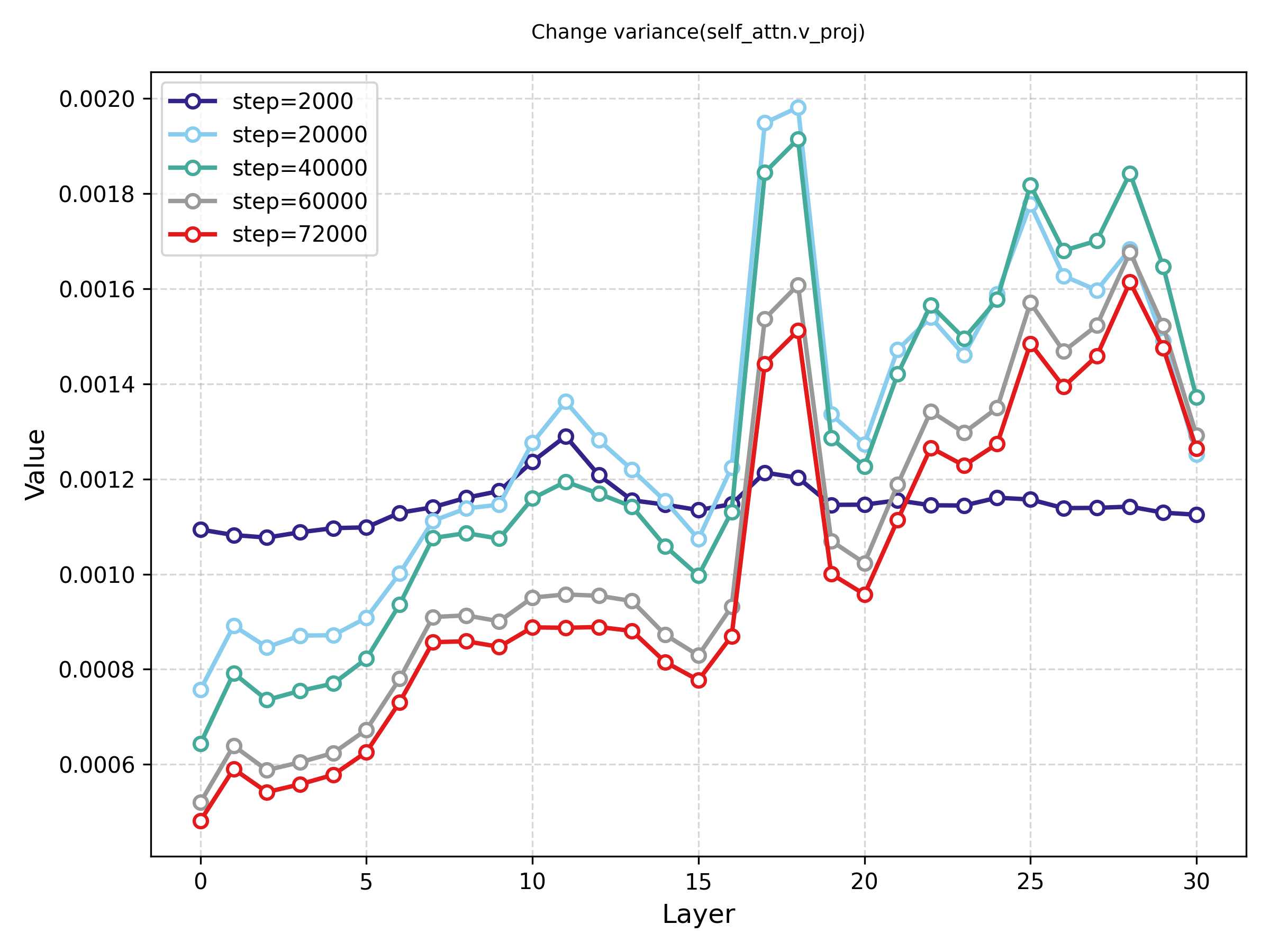}
    \caption{Condition D}
\end{subfigure}

\caption{Change variance in \texttt{mlp.up\_proj} for the two concentrated data regimes. Both conditions remain close to the baseline U-shaped profile reported in the main text, reinforcing the conclusion that change variance is comparatively insensitive to data regime and primarily reflects optimization schedule.}
\label{fig:app_ctrl_data_chg_var_cd}
\end{figure*}

\FloatBarrier

\subsection{External Validation: Additional Figures}
\label{app:ext_val_figures}

\begin{figure*}[htbp!]
\centering
\small

\begin{subfigure}[b]{0.45\linewidth}
    \centering
    \includegraphics[width=\linewidth]{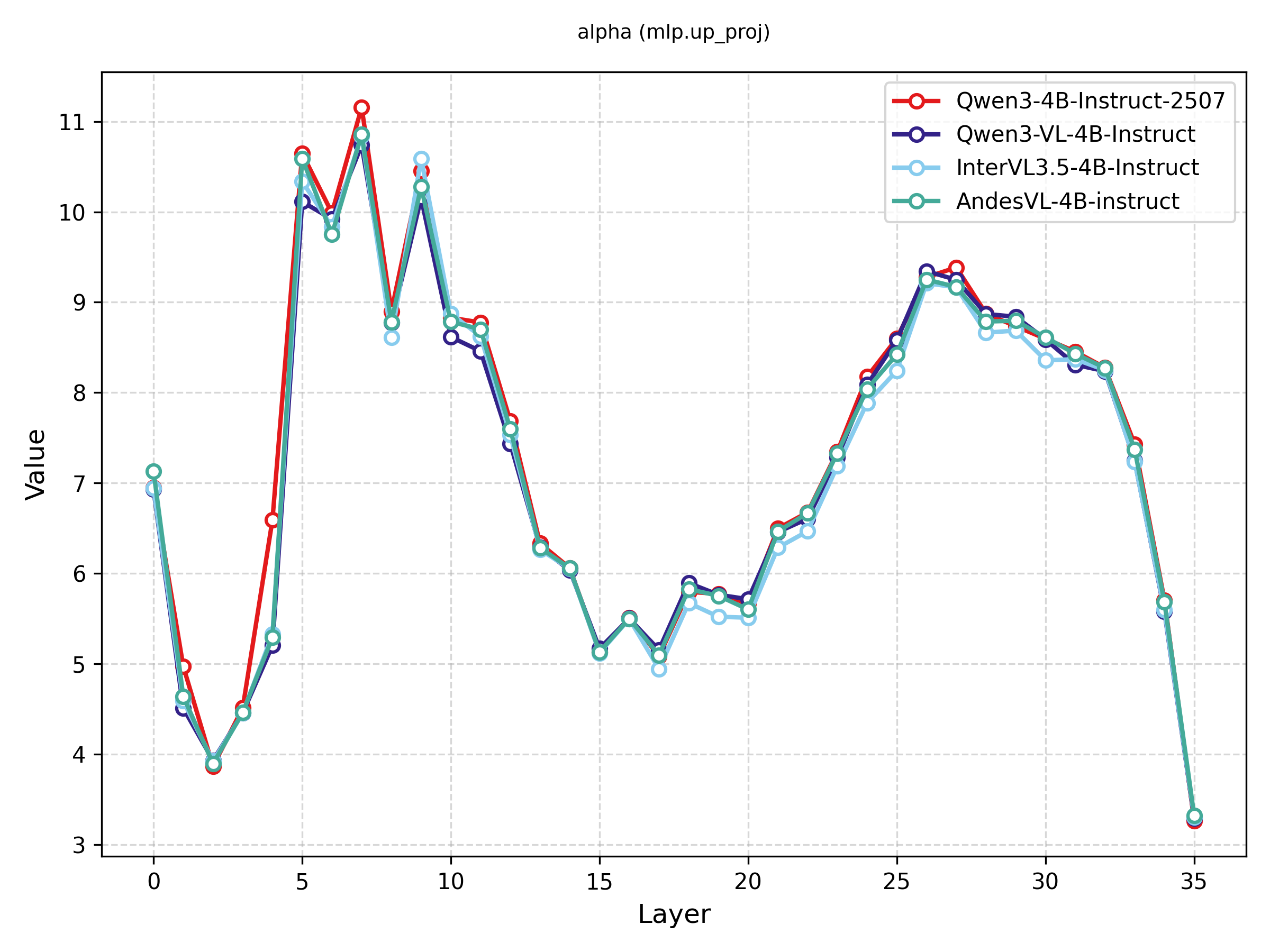}
    \caption{MLP $\alpha$}
\end{subfigure}
\hfill
\begin{subfigure}[b]{0.45\linewidth}
    \centering
    \includegraphics[width=\linewidth]{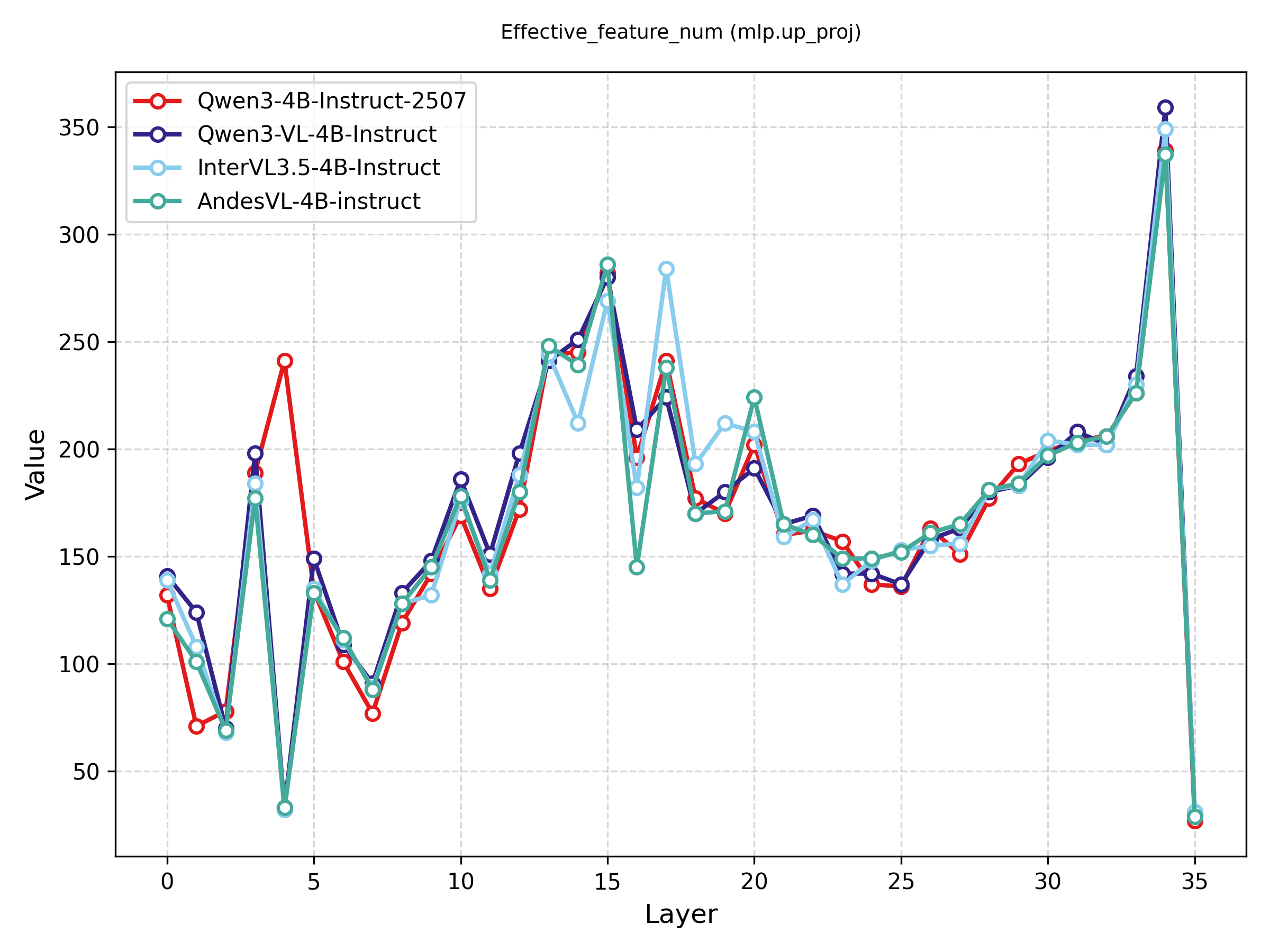}
    \caption{MLP effective feature number}
\end{subfigure}

\caption{Additional MLP-pathway comparisons across instruct models. Both $\alpha$ and effective feature number show near-invariance across model families, supporting the claim in Section~\ref{external_validation} that the MLP pathway is substantially more stable than the attention pathway.}
\label{fig:app_ext_val_mlp_invariance}
\end{figure*}

\begin{figure}[htbp!]
\centering
\small
\includegraphics[width=0.45\linewidth]{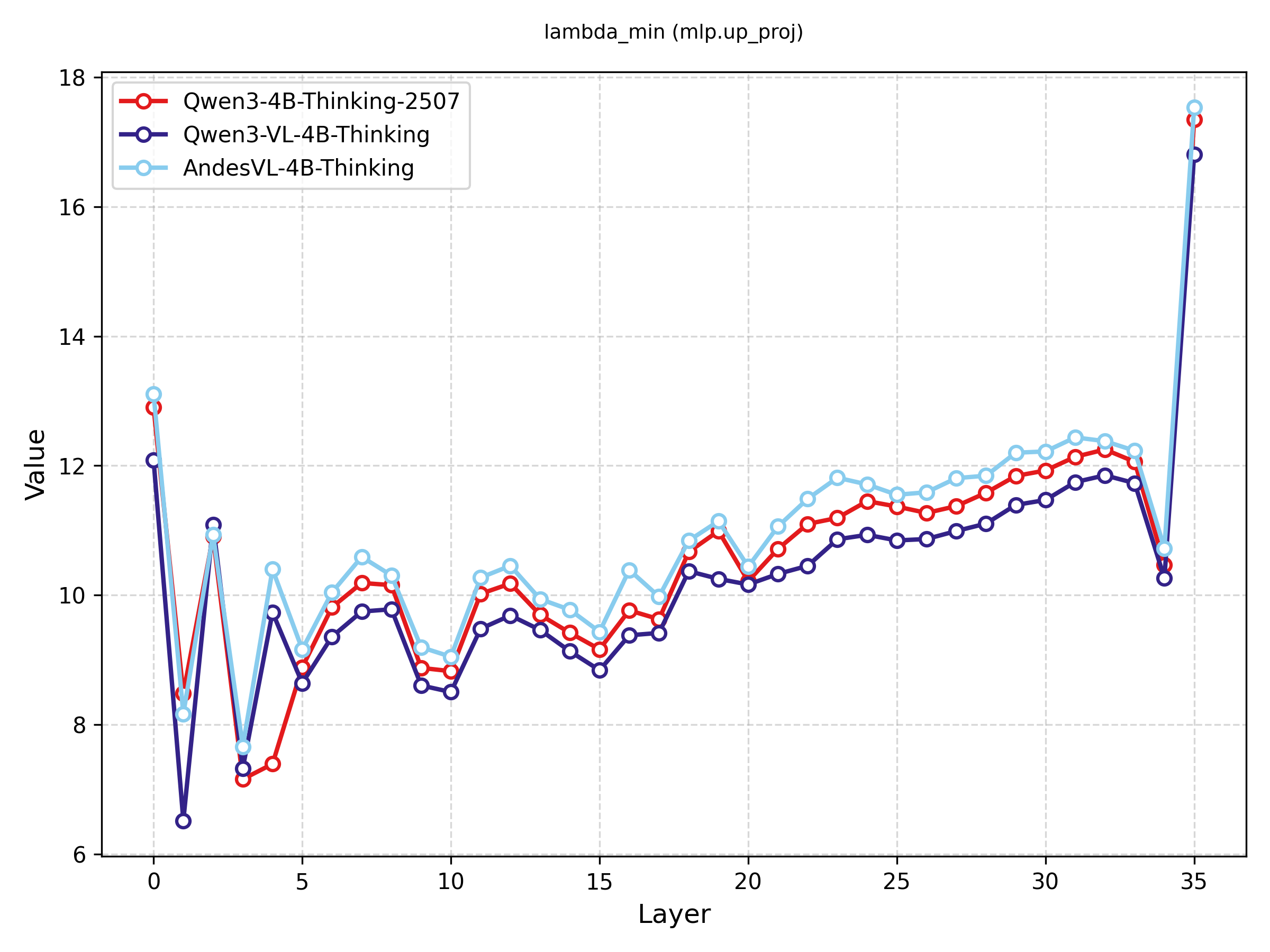}
\caption{Layer-wise $\lambda_{\min}$ in \texttt{mlp.up\_proj} across thinking variants. The rank ordering remains stable, indicating that the MLP-pathway separation identified in the instruct models persists under reasoning alignment.}
\label{fig:app_ext_val_mlp_lammin_think}
\end{figure}

\begin{figure}[htbp!]
\centering
\small
\includegraphics[width=0.45\linewidth]{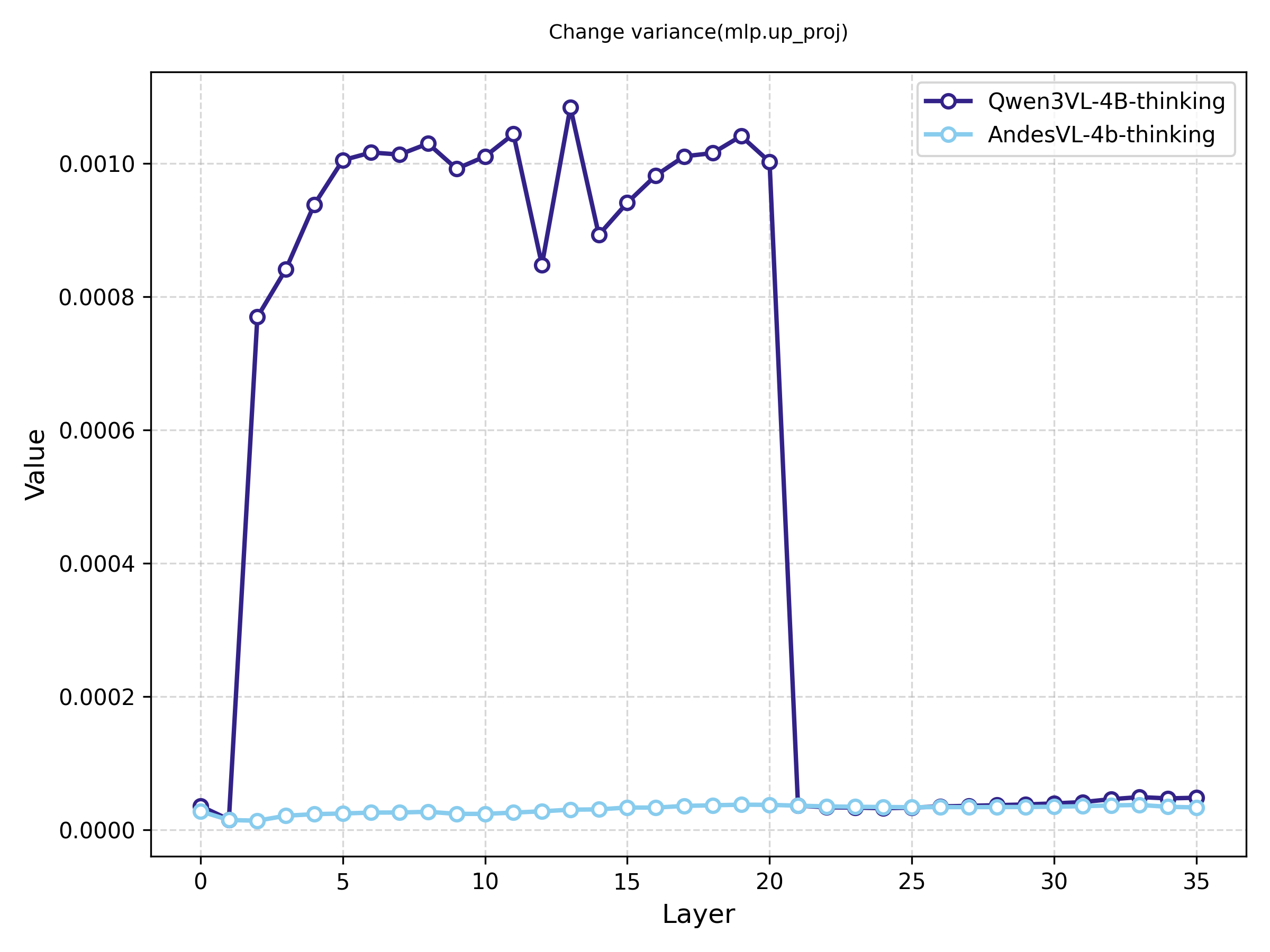}
\caption{Change variance in \texttt{mlp.up\_proj} for thinking-stage adaptation. This figure supports the interpretation that AndesVL undergoes nonzero parameter updates during reasoning alignment even when delta effective rank remains nearly flat, consistent with the spectral inertness discussed in Section~\ref{external_validation}.}
\label{fig:app_ext_val_mlp_chg_var_think}
\end{figure}

\begin{figure}[htbp!]
\centering
\small
\includegraphics[width=0.45\linewidth]{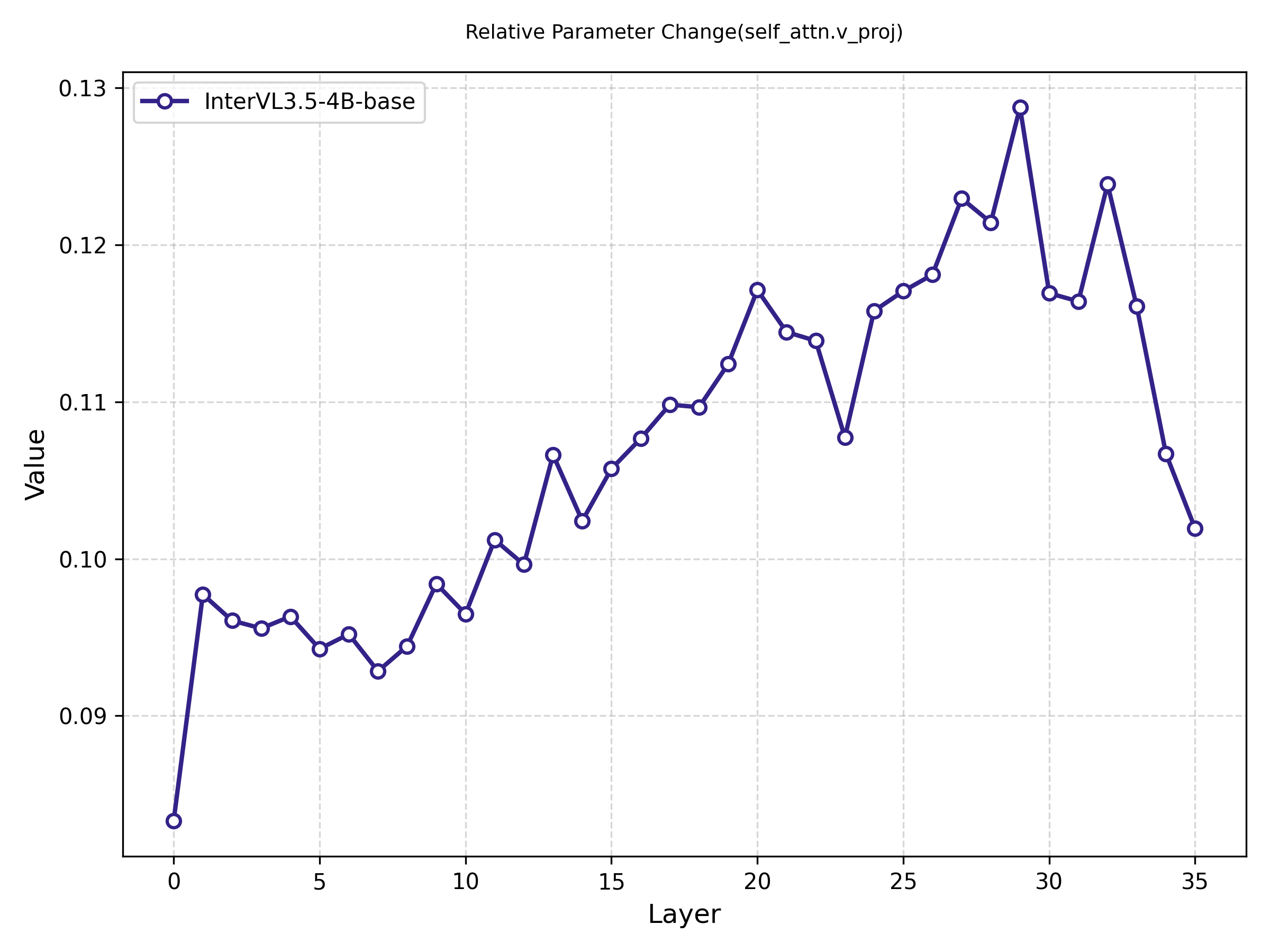}
\caption{Relative parameter change in \texttt{self\_attn.v\_proj} for InternVL at the base stage. This figure provides additional support for the minimal-perturbation regime referenced in Section~\ref{external_validation}.}
\label{fig:app_ext_val_attn_rel_base}
\end{figure}

\begin{figure*}[htbp!]
\centering
\small

\begin{subfigure}[b]{0.45\linewidth}
    \centering
    \includegraphics[width=\linewidth]{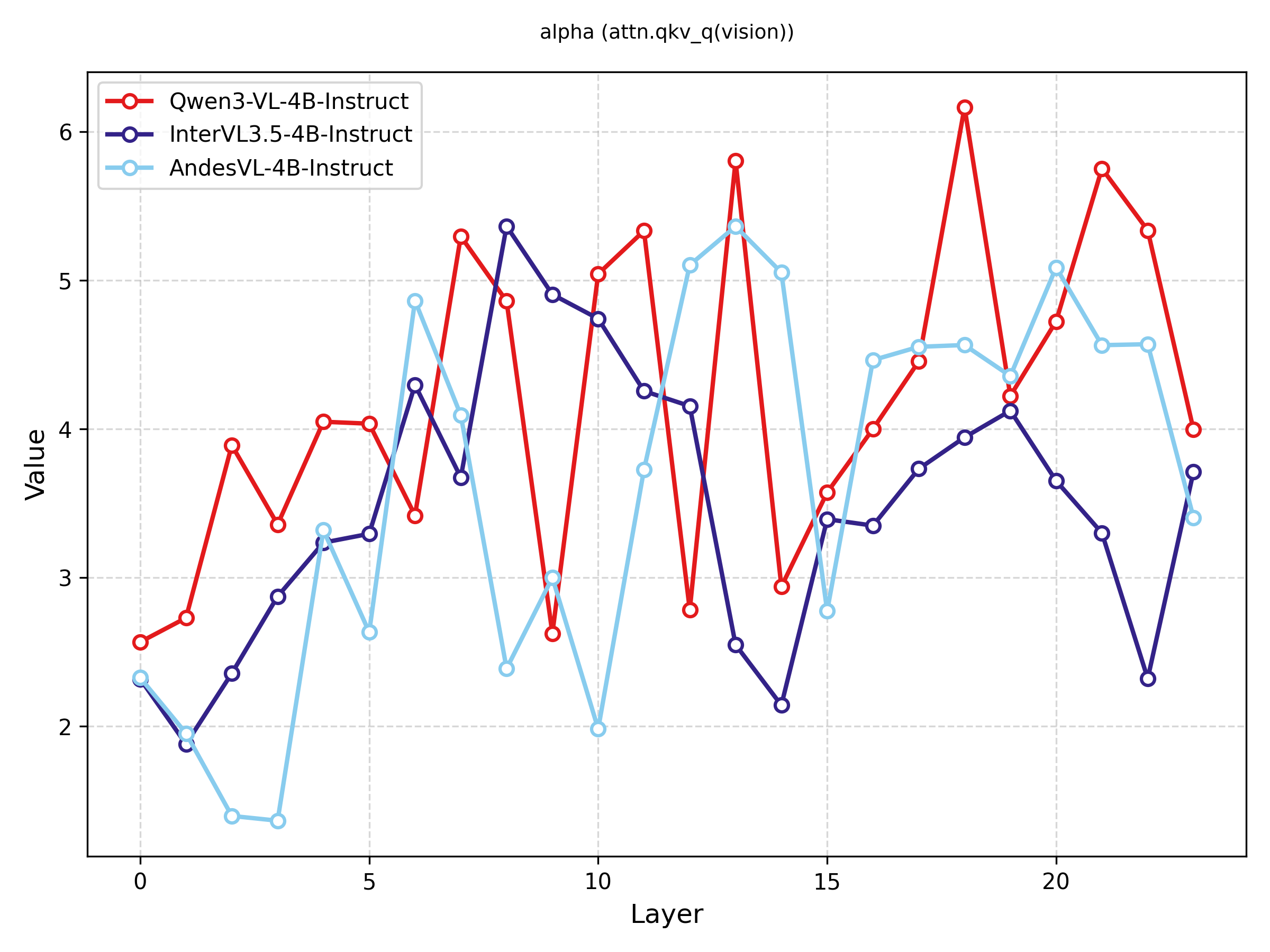}
    \caption{Vision encoder $\alpha$}
\end{subfigure}
\hfill
\begin{subfigure}[b]{0.45\linewidth}
    \centering
    \includegraphics[width=\linewidth]{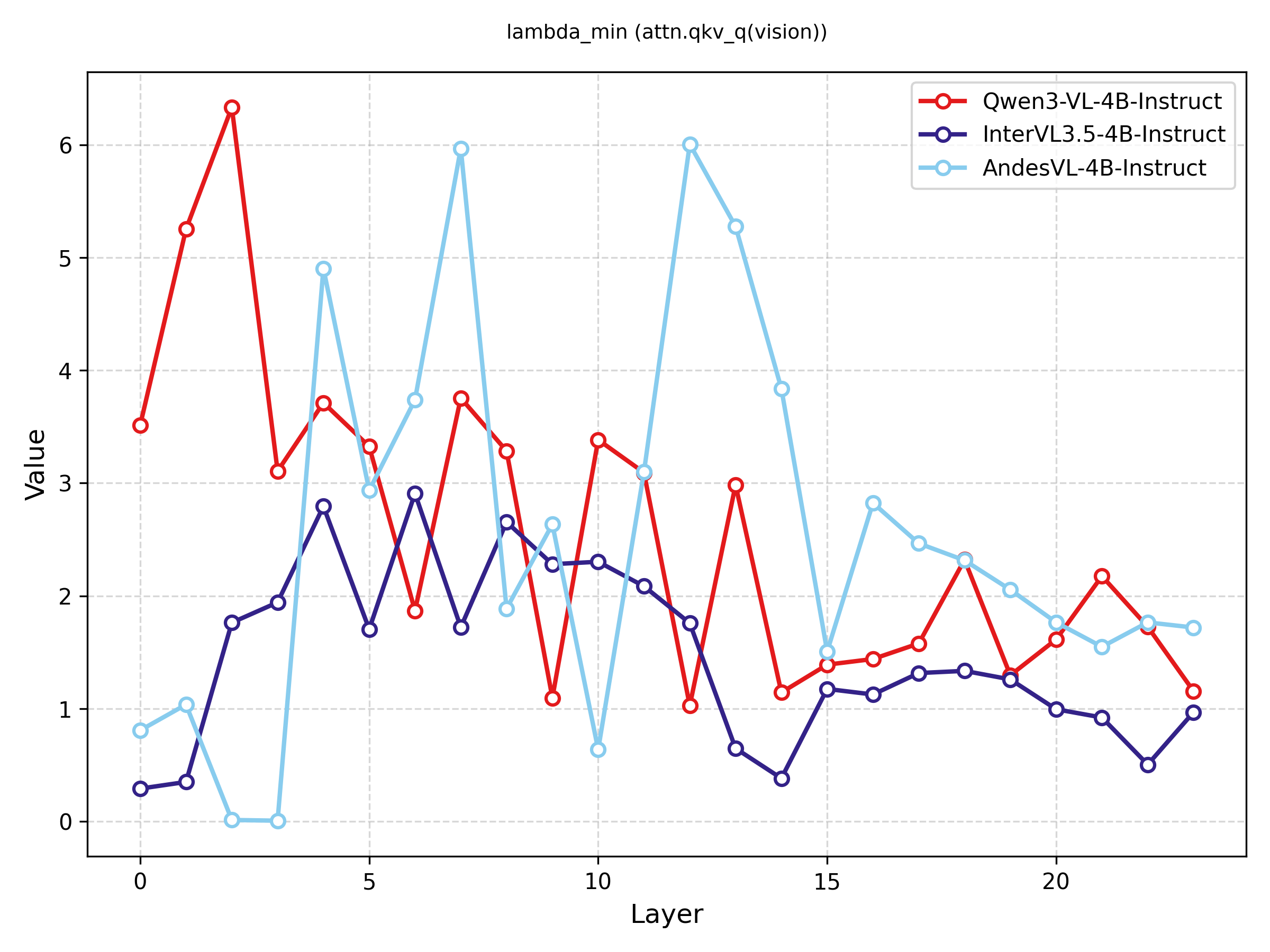}
    \caption{Vision encoder $\lambda_{\min}$}
\end{subfigure}

\caption{Additional vision encoder diagnostics for the attention query projection. AndesVL exhibits a sharp early-layer anomaly with unusually low $\alpha$ and near-zero $\lambda_{\min}$, supporting the interpretation in Section~\ref{external_validation} that localized structural irregularity can appear in modality-specific submodules even when decoder modification remains comparatively conservative.}
\label{fig:app_ext_val_vision_alpha_lammin}
\end{figure*}

\FloatBarrier

\subsection{Prompt De-duplication: Additional Figures}
\label{app:dedup_figures}

\begin{figure}[htbp!]
\centering
\small
\includegraphics[width=0.45\linewidth]{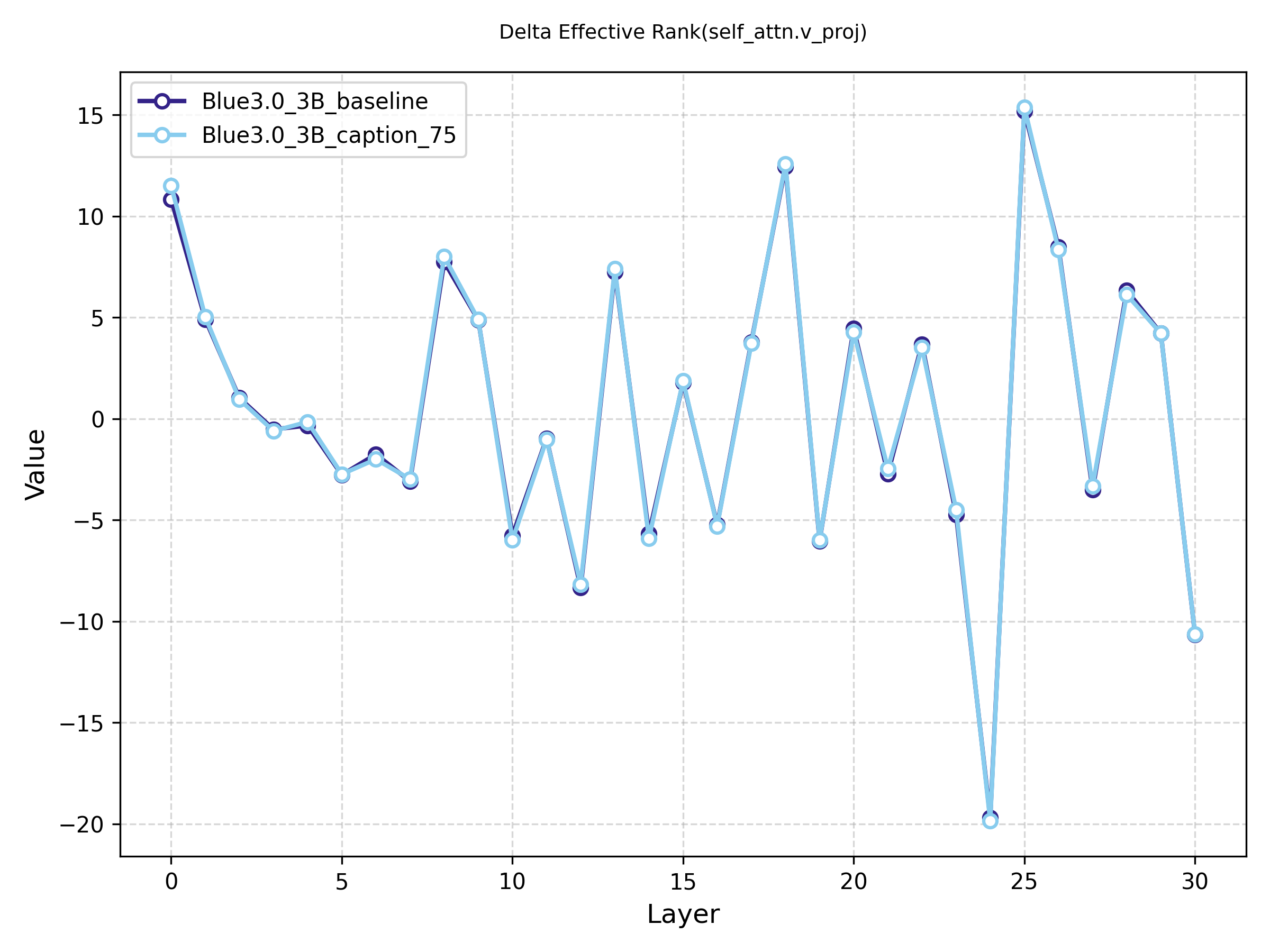}
\caption{Delta effective rank in the attention pathway under prompt de-duplication. The duplicated baseline and de-duplicated condition remain nearly indistinguishable across layers, providing additional evidence that prompt duplication does not induce the type of regime-level restructuring observed under the controlled benchmark-shadow conditions.}
\label{fig:app_dedup_attn_delta_eff_rank}
\end{figure}

\FloatBarrier


\end{document}